%% file: thesis.tex
\documentclass[11pt, a4paper]{book}
\usepackage{svn-multi}
\usepackage{appendix}
\svnid{$Id$}
\usepackage{prelim2e}

\usepackage[hyperindex=true,
			bookmarks=true,
            pdftitle={}, pdfauthor={Xi Yang},
            colorlinks=false,
            pdfborder=0,
            pagebackref=false,
            citecolor=blue,
            plainpages=false,
            pdfpagelabels,
            pagebackref=true,
            hyperfootnotes=false]{hyperref}
\usepackage[all]{hypcap}
\usepackage[palatino]{anuthesis}
\usepackage{afterpage}
\usepackage{graphicx}
\usepackage{thesis}
\usepackage[square]{natbib}
\usepackage[normalem]{ulem}
\usepackage[table]{xcolor}
\usepackage{makeidx}
\usepackage{amsmath}
\usepackage{cleveref}
\usepackage[centerlast]{caption2}
\usepackage{float}
\usepackage[T1]{fontenc}
\usepackage{times}
\usepackage{epsfig}
\usepackage{graphicx}

\usepackage{amssymb}
\usepackage{multirow}
\usepackage{booktabs}
\usepackage{pgfplots}
\usepackage{tikz}
\usepackage{pgfplots}
\usepgfplotslibrary{fillbetween}
\renewcommand*{\backref}[1]{}
\renewcommand*{\backrefalt}[4]{
  \ifcase #1 %
  \or
    (cited on page #2)%
  \else
    (cited on pages #2)%
  \fi
}

\input{macros}

\title{Deep Sequence Learning for Video Anticipation:\\From Discrete and Deterministic to Continuous and Stochastic}
\author{Mohammad Sadegh Aliakbarian}
\date{\today}

\renewcommand{\thepage}{\roman{page}}

\makeindex
\begin{document}
\pagestyle{empty}
\thispagestyle{empty}
\input{titlepage}

\input{frontmatter}

\cleardoublepage
\pagestyle{empty}
\input{dedication}

\cleardoublepage
\pagestyle{empty}
\input{ack}

\cleardoublepage
\pagestyle{headings}
\input{abstract}

\cleardoublepage
\pagestyle{headings}
\markboth{Contents}{Contents}
\tableofcontents
\listoffigures
\listoftables
\input{pub}
\mainmatter

\input{introduction}

\input{background}
\input{chapter1}

\input{chapter2}

\input{chapter3}
\input{chapter4}
\input{conclusion}

\input{appendix}


\backmatter

\bibliographystyle{anuthesis}
\bibliography{thesis}

\printindex

\end{document}

%% file: macros.tex
\usepackage{booktabs}
\usepackage{relsize}
\usepackage{xspace}
\usepackage{subfigure}
\usepackage{listings}
\definecolor{tableheadcolor}{rgb}{0.8,0.8,1.0}
\definecolor{tablealtcolor}{rgb}{0.9,0.9,0.95}

\definecolor{todocolor}{rgb}{0.8,0.8,1.0}
\definecolor{fixcolor}{rgb}{1,0.8,0.8}
\definecolor{commentcolor}{rgb}{0.8,1.0,0.8}

\usepackage[color=todocolor, colorinlistoftodos]{todonotes}

\newcommand{\textjava}[1]{{\lstset{basicstyle=\ttfamily}\lstinline@#1@}}
\newcommand{\textjavafn}[1]{{\lstset{basicstyle=\footnotesize\ttfamily}\lstinline@#1@}}
\usepackage{setspace}
\usepackage{ifthen}

\long\def\sfootnote[#1]#2{\begingroup%
\def\thefootnote{\fnsymbol{footnote}}\footnote[#1]{#2}\endgroup}
\newcommand{\doi}[1]{\href{http://dx.doi.org/#1}{\nolinkurl{doi:#1}}}
\newcommand{\ignore}[1]{}

%% file: titlepage.tex

\begin{titlepage}
  \enlargethispage{2cm}
  \begin{center}
    \makeatletter
    \Huge\textbf{\@title} \\[.4cm]
    \Huge\textbf{\thesisqualifier} \\[2.5cm]
    \huge\textbf{\@author} \\[9cm]
    \makeatother
    \LARGE A thesis submitted for the degree of \\
    Doctor of Philosophy \\
    The Australian National University \\[2cm]
    \thismonth
  \end{center}
\end{titlepage}

%% file: frontmatter.tex
\vspace*{14cm}
\begin{center}
  \makeatletter
  \copyright\ \@author{} 2020
  \makeatother
\end{center}
\noindent
\begin{center}
  \footnotesize{~} 
\end{center}
\noindent

\newpage

\vspace*{7cm}
\begin{center}
  I hereby declare that this thesis is my original work which has been done in
collaboration with other researchers. This document has not been submitted to obtain
a degree or award in any other university or educational institution. Parts of this
thesis have been published in collaboration with other researchers in international
conferences and journals as listed in the Publications section.
\end{center}

\vspace*{4cm}

\hspace{8cm}\makeatletter\@author\makeatother\par
\hspace{8cm}\today

%% file: dedication.tex
\vspace*{7cm}
\begin{center}
To my loving wife and my beloved parents
\end{center}

%% file: ack.tex
\chapter*{Acknowledgments}
\addcontentsline{toc}{chapter}{Acknowledgments}
There are a number of people,  without whom this thesis might not have been written.

Firstly, I would like to express my deepest gratitude to my amazing supervisors, Dr. Lars Petersson, Dr. Mathieu Salzmann, Dr. Basura Fernando, and Prof. Steven Gould for their generous support, motivation, and patience throughout my study. 
I would like to thank Dr. Lars Petersson for his generous and continuous support. 
He has always been friendly and professional, so that the research environment was always enjoyable and fruitful for the entire team.
I remember when I was working on my first paper, he puts a lot of energy and time to make sure that I do not lose my hopes at the beginning of this journey. Near the deadline, he stayed in office until very late at night to help me finish the experiments.
Beyond research, he also cares for the well-being of his students. I never forget the fun moments we had in our all year long Friday teas.
I also would like to thank Dr. Mathieu Salzmann. Although he was not in Australia, he kindly continued supervising me and attended every meetings we had, even very early in the morning back in Switzerland.
I learned a lot about how to approach a problem, how to criticize my solutions to make it perfect, and how to design experiments so that the message is clearly and efficiently conveyed. As a non-English speaker, I also learned a lot how to write clearly and precisely.
I am also very grateful to Dr. Basura Fernando, whom I had the chance to work with during my PhD. While being one of the pioneers in the field, he is so humble. I would like to thank him for his support specially in the practical aspects of my research.
I want to thank Dr. Stephen Gould who gave detailed insightful comments on my recent research projects during my PhD. His comments, guidance, and ideas helped me a lot improving my works. He also toughed me how to prepare better research presentations, which is going to be a skill that I always need. I also would like to thank him and the ACRV for the generous support for the travels.

I would also show my gratitude to my amazing colleagues during my internship at Qualcomm. Thanks to Amir and Babak who made that four-months internship a memorable one. I want to thank my amazing and professional colleagues at FiveAI when I was an intern there. Thanks to my mentor, Nick, and other colleagues, Stuart, Luca, Tommaso, Oscar, and Steinar. Also, thanks to Prof. Philip Torr and Dr. Puneet Dokania for their insightful discussions on my internship project.

I am also grateful to NICTA/Data61 and the Australian National University for providing my Ph.D. scholarship and enriching my academic experience by providing conference travel funding.
I thank Ali C., Ali A., Mehrdad, and Dylan, as well as all other past and the present members of the Data61’s Smart Vision System’s Group especially Lars Andersson and Lachlan Tychsen-Smith for helping me dealing with technical aspects of the work. 

I am also very happy that I found such an amazing friends here in Australia. I want to thank Sarah and Mohammad E., my first and two of the best friends, who did not know me when I just arrived to Australia, but helped me unconditionally to settle in. Throughout these years, they always help me, like siblings, and made sure everything is always alright. 
To Mohammad N., a great officemate and an amazing friend, who I had really great time with. A special thanks to Alireza and Masoumeh, and adorable little Rose who are just our family and made living here much more enjoyable. To Alireza K., Mehdi, Ehsan, Mousa, Salim, Farshad, Hajar, who is no longer with us but remains in our mind and heart, and my all other friends, I won't forget your support and unforgettable weekend parties, game nights, and short trips, which we could relax and forget the stress and the deadlines of academia. 

I would like to express my gratitude to my parents-in-law for their unfailing emotional support. I also would like to thank my sisters-in-law, specially Fahimeh, with whom I feel we have a bigger family here in Australia. 

A very special word of thanks goes for my inspiring and supportive parents. Your love, support and encouragement cannot be expressed by words. This accomplishment would not have been possible without your support. Thanks to my dear sisters Fatemeh and Zahra for their emotional support, endless love and care.

Finally, I would like to thank my special person, my loving wife, who has always been my best friend and my amazing colleague. Fatemeh, thank you so much for your unconditional love, you continued and unfailing support, and your patience and understanding during my study. Words cannot express how I'm grateful for your contributions and suggestions to my research. I am grateful that you never stopped supporting me and believing in me. Every time I fail, you are the one who takes my hand and help me stand again. What can be better than having such a lovely and caring person?

%% file: abstract.tex
\chapter*{Abstract}
\addcontentsline{toc}{chapter}{Abstract}
\vspace{-1em}

Video anticipation is the task of predicting one/multiple future representation(s) given limited, partial observation. This is a challenging task due to the fact that given limited observation, the future representation can be highly ambiguous. 
 Based on the nature of the task, video anticipation can be considered from two viewpoints: the level of details and the level of determinism in the predicted future. In this research, we start from anticipating a coarse representation of a deterministic future and then move towards predicting continuous and fine-grained future representations of a stochastic process. The example of the former is video action anticipation in which we are interested in predicting one action label given a partially observed video and the example of the latter is forecasting multiple diverse continuations of human motion given partially observed one. 

In particular, in this thesis, we make several contributions to the literature of video anticipation. The first two contributions mainly focus on anticipating a coarse representation of a deterministic future while the third and fourth contributions focus on predicting continuous and fine-grained future representations of a stochastic process.
Firstly, we introduce a general action anticipation framework in which,  given very limited observation, the goal is to predict the action label as early as possible. This task is highly critical in scenarios where one needs to react before the action is finalized. This is, for instance, the case of automated driving, where a car needs to, e.g., avoid hitting pedestrians and respect traffic lights. Our work builds on the following observation: a good anticipation model requires (i) a good video representation that is discriminative enough even in presence of partial observation; and (ii) a learning paradigm that not only encourages correct predictions as early as possible, but also accounts for the fact that the future is highly ambiguous. On publicly available action recognition datasets, our proposed method is able to predict markedly accurate action categories given very limited observation, e.g., less than 2\% of the videos of UCF-101, outperforming the state of the art methods by a large margin.
Secondly, we proposed an action anticipation in driving scenarios. Since there was no anticipation-specific dataset covering generic driving scenarios, as part of our second contribution, we introduced a large-scale video anticipation dataset, covering 5 generic driving scenarios, with a total of 25 distinct action classes. It contains videos acquired in various driving conditions, weathers, daytimes and environments, complemented with a common and realistic set of sensor measurements. This dataset is now publicly available\footnote{\tt \url{https://sites.google.com/view/viena2-project/}}.
We then  focus on the continuous future prediction problem on tasks that are stochastic in nature; given one observation, multiple plausible futures are likely. In particular, we target the problem of human motion prediction, i.e., the task of predicting future3D human poses given a sequence of observed ones. To this end, in our third contribution, we propose a novel diverse human motion prediction framework based on variational autoenecoders (VAEs). In this approach, we particularly propose a novel stochastic conditioning scheme that is well-suited for scenarios where we are dealing with a deterministic datasets, with strong conditioning signals, and expressive decoders. Through extensive experiments, we show that our approach performs much better than existing approaches and standard practices in training a conditional VAE. 
Finally, in the fourth contribution, we propose a conditional VAE framework that solves two main issues of a standard conditional VAE: (i) conditioning and sampling the latent variables are two independent processes, and (ii) the prior distribution is set to be unconditional in practice, however, it should be conditioned on the conditioning signal as elaborated in the evidence lower bound of the data likelihood. In our proposed approach, we address both of these issues that leads to substantial improvement in the quality of generated samples.

All the methods introduced in this thesis are evaluated on standard benchmark datasets. The experiments at the end of each chapter provide compelling evidence that all of our approaches are more efficient than the contemporary baselines.

%% file: pub.tex
\chapter*{Publications}
\addcontentsline{toc}{chapter}{Publications}
The following publications have resulted from the work presented in this thesis:
\begin{itemize}
\item Sadegh Aliakbarian, Fatemeh Sadat Saleh, Mathieu Salzmann, Basura Fernando, Lars Petersson, and Lars Andersson. \textit{Encouraging lstms to anticipate actions very early.} In Proceedings of the IEEE International Conference on Computer Vision, pp. 280-289. 2017.
\item Sadegh Aliakbarian, Fatemeh Sadat Saleh, Mathieu Salzmann, Basura Fernando, Lars Petersson, and Lars Andersson. \textit{VIENA$^2$: A Driving Anticipation Dataset.} In Asian Conference on Computer Vision, pp. 449-466. Springer, Cham, 2018.
\item Sadegh Aliakbarian, Fatemeh Sadat Saleh, Mathieu Salzmann, Lars Petersson, and Stephen Gould. \textit{A Stochastic Conditioning Scheme for Diverse Human Motion Prediction.} In Proceedings of the IEEE/CVF Conference on Computer Vision and Pattern Recognition, pp. 5223-5232. 2020.
\item Sadegh Aliakbarian, Fatemeh Sadat Saleh, Mathieu Salzmann, Lars Petersson, and Stephen Gould. \textit{Sampling Good Latent Variables via CPP-VAEs: VAEs with Condition Posterior as Prior.} arXiv preprint arXiv:1912.08521 (2019). To be submitted to CVPR 2021.
\end{itemize}

%% file: introduction.tex
\chapter{Introduction}

As for many other computer vision tasks, a recognition model gets as input the whole observation, e.g., an image or a video, and predicts a label. For instance, in video/action recognition, a model gets as input all the information in the video and predicts a label that represents the most likely category/event happened in that video. Although understanding such information given the full extent of videos is key to the success of wide variety of applications, the full observation is not always available in many real-world cases, such as autonomous navigation, surveillance and sports analysis.  In these application, it is crucial to reliably predict the \textit{future} action of a particular sequence. For instance, an autonomous car should always predict the intention of all nearby pedestrians a few seconds ahead to e.g., avoid hitting them.

Video anticipation is one of the key challenges in computer vision and video analysis. Anticipation is the task of predicting future representation(s) given limited, partial observation. Having a reliable anticipation system is highly critical in scenarios where one needs to react before the action is finalized, such as in pedestrian intention forecasting system in autonomous vehicles.
Sequence anticipation is a challenging task due to the fact that given limited observation, the future representation can be highly ambiguous. Given the task, one may anticipate a coarse, discrete representation per limited observation, called \emph{discrete} anticipation, or a fine-grained, continuous representation, called \emph{continuous} anticipation. There are also cases where the future representation can be considered \emph{deterministic}, i.e., expecting one single likely outcome for a given observation, or can be considered \emph{stochastic}, where multiple plausible future representations are likely and correct.

In this chapter, we first provide a definition of the problem as well as the motivations behind this research. We then summarize our contributions and the thesis outline.

\section{Problem Definition}
\label{sec:definition}

\paragraph{Discrete and Deterministic Anticipation.} In a discrete anticipation problem, the goal is to predict the desired outcome $y_T$ given the partial observation $x_{1:t}$ where $T$ is the length of the sequence and $t<T$ is the length of the partial observation. The desired outcome $y_T$ can be a binary value determining whether what will happen in the future is normal or abnormal or can be a categorical determining the action of interest at time $T$. For discrete anticipation, we specifically study the problem of action anticipation in this thesis. In contrast to the widely studied problem of recognizing an action given a complete sequence, as introduced above, action anticipation aims to identify the action from only partially available videos. As such, it is therefore key to the success of computer vision applications requiring to react as early as possible, such as autonomous navigation. Note that, when addressing discrete anticipation task, we focus on only a deterministic scenario, where we are interested in predicting only one $y_T$ that has the highest likelihood of occurrence.

\paragraph{Continuous and Stochastic Anticipation.} In a continuous anticipation task, the goal is to predict a sequence of desired outcomes $y_{t+1:T}$ given the partial observation $x_{1:t}$ where $T$ is the length of the sequence and $t<T$ is the length of the partial observation. The main difference to discrete anticipation is one need to anticipate the representation at every time-step in the future such that the whole sequence remains natural and plausible. Unlike discrete anticipation where the outcome representation (e.g., action label) differs from the input representation (e.g., consecutive RGB frames), in continuous anticipation these two representation are usually from the same domain, casting video anticipation problem into future representation generation. Examples of continuous anticipation is trajectory prediction, video generation, and human motion prediction, where in this thesis we focus on the latter. Similar to many other real-world tasks in continuous anticipation, human motion prediction is a highly stochastic process: given an observed sequence of poses, multiple future motions are plausible. Thus, it is crucial to take stochasticity of this problem into account. Therefore, in this thesis we focus on stochastic scenario and try to generate multiple likely continuations of an observed human motion.




\section{Thesis Outline and Contributions}
\label{sec:contributions}

This thesis comprises seven chapters. Apart from this chapter that introduces the problem and provides our motivations to address the problem video anticipation, in Chapter 2, we introduce the technical background of the methods we used in this thesis. This background material consists of some concepts and theories that are common to many of the approaches proposed in later chapters, including the convolutional neural network architectures, recurrent neural networks, and variational autoencoders. The next four chapters provide details of our novel techniques for sequence learning for video anticipation. Particularly, in Chapter 3, we introduce our novel techniques for encouraging LSTMs to anticipate actions as early as possible. In Chapter 4, we introduce a new anticipation-specific dataset as well as a novel multi-modal LSTM architecture that generalizes our previous contribution to an arbitrary number of modalities. In Chapter 5, we address the problem of human motion prediction with the focus on the stochastic nature of this problem. Then, in Chapter 6, we propose a novel framework for generating diverse and semantically high quality sequences given partial observation. Finally, in Chapter 7, we summarise the main contributions of the thesis and discuss ongoing and future work stemming from this research.

A brief summary of each contribution is provided below.

\subsection{General Action Anticipation}
We propose a novel action anticipation framework (that can be also seen as an \textit{early recognition} model). In particular, we introduce a novel loss that encourages making correct predictions very early. Our loss models the intuition that some actions, such as running and high jump, are highly ambiguous after seeing only the first few frames, and false positives should therefore not be penalized too strongly in the early stages. By contrast, we would like to predict a high probability for the correct class as early as possible, and thus penalize false negatives from the beginning of the sequence. Our experiments demonstrate that, for a given model, our new loss yields significantly higher accuracy than existing ones on the task of early prediction. We also propose a novel multi-stage Long Short Term Memory (MS-LSTM) architecture for action anticipation. This model effectively extracts and jointly exploits context- and action-aware features. This is in contrast to existing methods that typically extract either global representations for the entire image or video sequence, thus not focusing on the action itself, or localize the feature extraction process to the action itself via dense trajectories, optical flow or actionness, thus failing to exploit contextual information~\citep{sadegh2017encouraging}. 

\subsection{Action Anticipation in Driving Scenarios}
We improve and extend our previous contribution by focusing on driving scenarios, encompassing common the subproblems of anticipating ego car's driver maneuvers, front car's driver maneuver, accidents, violating or respecting traffic rules, and pedestrian intention, with a fixed, sensible set of sensors. To this end, we introduce the VIrtual ENvironment  for Action Analysis  (VIENA$^2$) dataset. Altogether, these subproblems encompass a total of 25 distinct action classes. VIENA$^2$ is acquired using the GTA V video game. It contains more than 15K full HD, 5s long videos, corresponding to more than 600 samples per action class, acquired in various driving conditions, weathers, daytimes, and environments. This amounts to more than 2.25M frames, each annotated with an action label. These videos are complemented by basic vehicle dynamics measurements, reflecting well the type of information that one could have access to in practice\footnote{Our dataset is publicly available at \url{https://sites.google.com/view/viena2-project/}}. We then benchmark state-of-the-art action anticipation algorithms on VIENA2, and as another contribution, introduce a new multi-modal, LSTM-based  architecture that  generalizes  out previous contribution to an arbitrary number of modalities,  together  with  a  new  anticipation  loss,  which  outperforms existing approaches in our driving anticipation scenarios~\citep{aliakbarian2018viena}.

\subsection{A Stochastic Conditioning Scheme for Diverse Sequence Generation}
For continuous, stochastic anticipation task, we address the problem of stochastic human motion prediction. As introduced earlier in this thesis, human motion prediction aims to forecast the sequence of future poses of a person given past observations of such poses. To achieve this, existing methods typically rely on recurrent neural networks (RNNs) that encode the person’s motion. While they predict reasonable motions, RNNs are deterministic models and thus cannot account for the highly stochastic nature of human motion; given the beginning of a sequence, multiple, diverse futures are plausible. To correctly model this, it is therefore critical to develop algorithms that can learn the multiple modes of human motion, even when presented with only deterministic training samples. We introduce an approach to effectively learn the stochasticity in human motion. At the heart of our approach lies the idea of Mix-and-Match perturbations: Instead of combining a noise vector with the conditioning variables in a deterministic manner (as usually done in standard practices), we randomly select and perturb a subset of these variables. By randomly changing this subset at every iteration, our strategy prevents training from identifying the root of variations and forces the model to take it into account in the generation process. This is a highly effective conditioning scheme in scenarios when (1) we are dealing with a deterministic dataset, i.e., one sample per condition, (2) the conditioning signal is very strong and representative, e.g., the sequence of past observations, and (3) the model has an expressive decoder that can generate a plausible sample given only the condition.
We utilize Mix-and-Match by incorporating it into a recurrent encoder-decoder network with a conditional variational autoencoder (CVAE) block that learns to exploit the perturbations. Mix-and-Match then acts as the stochastic conditioning scheme instead of concatenation that usually appears in standard CVAEs~\citep{Aliakbarian_2020_CVPR}. 

\subsection{Variational Autoencoders with Learned Conditional Priors}
In our previous contribution, we identified one limitation of a standard CVAEs when dealing deterministic datasets and strong conditioning signals. In this work, we further investigates this problem from a more theoretical point of view. Specifically, in this contribution we tackle the task of diverse sequence generation in which all the diversely generated sequences carry the same semantic as in the conditioning signal. We observe that in standard CVAE, conditioning and sampling the latent variable are two independent processes, leading to generating samples that are not necessarily carry all the contextual information about the condition. To address this, we propose to explicitly make the latent variables depend on the observations (the conditioning signal). To achieve this, we develop a CVAE architecture that learns a distribution not only of the latent variables, but also of the observations, the latter acting as prior on the former. By doing so, we change the variational family of the posterior distribution of the CVAE, thus, as a side effect, our approach can mitigate posterior collapse to some extent~\citep{aliakbarian2019sampling}. 


%% file: background.tex
\chapter{Technical Background}
In this chapter we introduce the background materials, architectures and models that have been used in this thesis. We use Convolutional Neural Networks (CNNs) widely as the feature/representation learning approach for discrete anticipation task. We employ variants of Recurrent Neural Networks (RNNs) to effectively model sequences. Finally, since we cast the problem of continuous anticipation as generating future representation, we use Variational Autoencoders (VAEs) as our backbone generative model in this thesis. Below, we review the utilized CNN architectures, the variants of RNNs used in this thesis, and VAEs to help the reader better understand the following chapters.

\section{Convolutional Neural Networks}
\label{sec:CNN}

Convolutional neural networks~\citep{karpathy2014large}, or CNNs in short and also known as ConvNets, are powerful tools used in many machine learning and computer vision tasks. With the advances in deep learning and compute, CNNs achieve the state-of-the-art performance on various tasks such as image recognition~\citep{simonyan2014very,karpathy2014large,he2016deep,huang2017densely,krizhevsky2012imagenet}, semantic segmentation~\citep{long2015fully,noh2015learning,saleh2017incorporating,saleh2017bringing,zhao2017pyramid}, action recognition~\citep{wang2016temporal,carreira2017quo,simonyan2014two,feichtenhofer2016convolutional,wu2019long,aliakbarian2016deep}, object detection~\citep{redmon2016you,tychsen2017denet,ren2015faster,girshick2015fast,girshick2014rich}, etc. CNNs comprise stack of layers of different types, including but not limited to convolutional, pooling, and fully-connected (also known as linear) layers. Common CNN architectures stack blocks of these operations to perform feature representation learning which is done by learning trainable parameters of each layer. Since, in most cases, different layers are stacked on top of each other, deeper layers perform operations on the output of shallower layers, thus learning the feature representation in a hierarchical manner. Hence, it is usually the case where the features computed by the deeper layers carry out more semantically meaningful information while the features of shallower layers provide detailed, but low-level information~\citep{saleh2016built,bertasius2015deepedge}. This characteristic of CNNs allows us to use the output features of deeper layers as a good representation of the input image~\citep{wang2016temporal,carreira2017quo,simonyan2014two,feichtenhofer2016convolutional,wu2019long}. In this thesis, we use CNNs as feature extractors for representing the frames of videos such that it becomes feasible to use them as the input to the recurrent models.

In this section, we assume that the reader has a knowledge about CNNs building blocks, layers, and optimization, thus, we only briefly introduce the special CNN architectures we used later in this thesis.

There are many powerful CNN architectures that shown effective in many computer vision tasks. Among those, we use VGG16~\citep{simonyan2014very} and DenseNet~\citep{huang2017densely} architectures in the following chapters. In this section we describe these two architectures briefly.

\begin{figure}
    \centering
    \includegraphics[width=\textwidth]{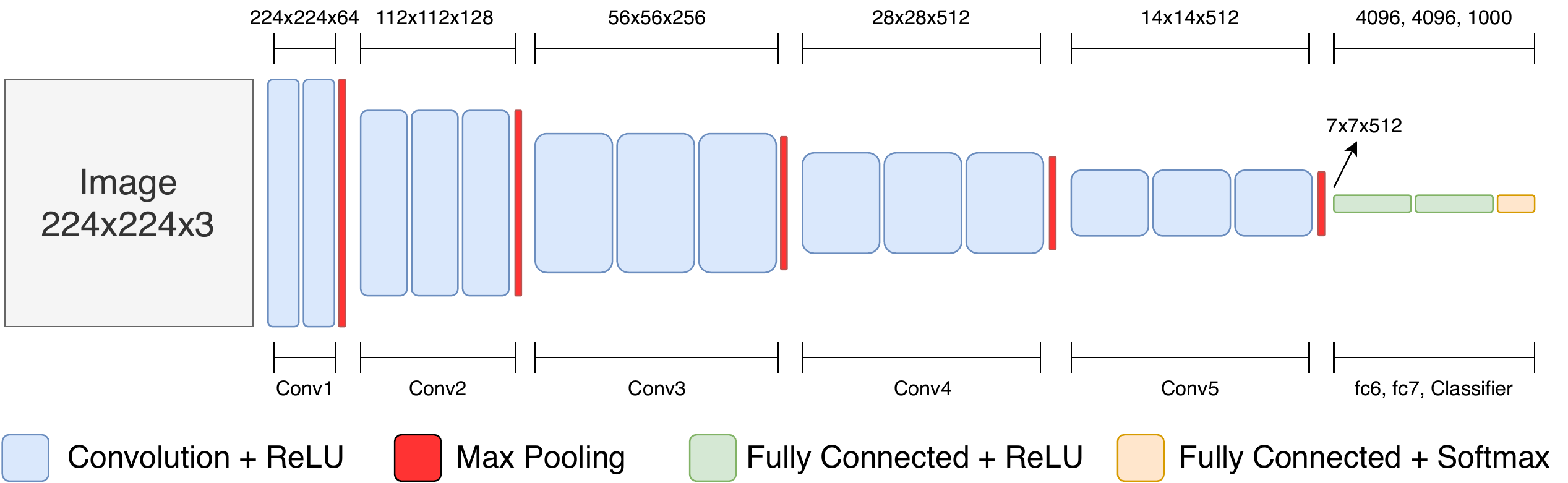}
    \caption{
    VGG16 architecture.
    }
    \label{fig:vgg}
\end{figure}

\paragraph{VGG16.} Introduced in 2014, VGG16~\citep{simonyan2014very} considered to be a very deep network. This network, as shown in Fig.~\ref{fig:vgg} comprises five convolutional blocks followed by three fully-connected layers. Each convolutional block contains convolutional, pooling, and nonlinearity layers. Pre-trained on ImageNet dataset~\citep{krizhevsky2012imagenet} and fine-tuned on the target dataset, this network can extract discriminative features from each image.

{}

\begin{figure}
    \centering
    \includegraphics[width=\textwidth]{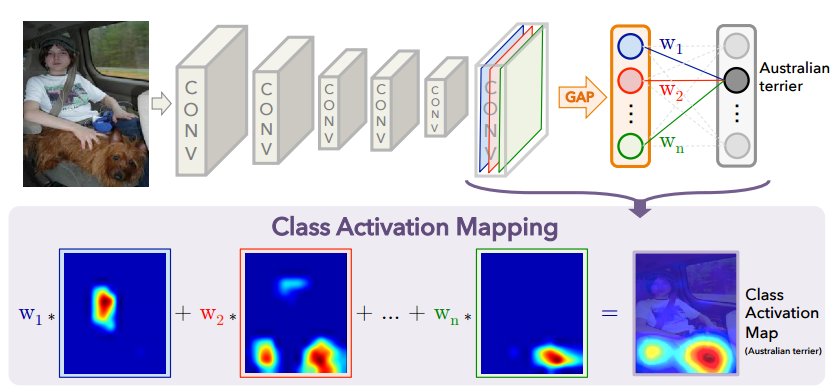}
    \caption{Class Activation Maps}
    \label{fig:cam}
\end{figure}

In this thesis, followed by the idea of Class Activation Map (CAM)~\citep{zhou2016learning}, we modified the VGG16 architecture such that is becomes capable of generating heatmaps that illustrate the most discriminative part of image in determining a particular class. As depicted in Fig.~\ref{fig:cam}, one can add a global average pooling (GAP) layer~\citep{zhou2016learning} after the fifth convolutional block of the VGG16 architecture. The output of GAP is then used as the input features for a fully-connected layer that maps the features to the number of classes. Given such design, one can generate a coarse representation that determines the most discriminative part of the input simply by projecting back the weights of the output layer on to the last convolutional feature maps. More specifically, let $f_l(x,y)$ represent the activation of unit $l$ in the last convolutional layer at the spatial location $(x,y)$. A score $S_k$ for each class $k$ can be obtained by performing GAP to obtain, or each unit $l$, a feature $F^l=\sum_{x,y}f_l(x,y)$, followed by a fully-connected layer with the set of weights $\{w_l^k\}$. That is,

\begin{align}
    S_k=\sum_{k} w^k_lF_l.
\end{align}

A CAM for a class $k$ at location $(x,y)$ can be then computed as

\begin{align}
    M_k(x,y)=\sum_l w_l^k f_l(x,y),
\end{align}
which indicates the importance of the activations at location $(x,y)$ in the final score for class $k$.

{}

\begin{figure}
    \centering
    \includegraphics[width=\textwidth]{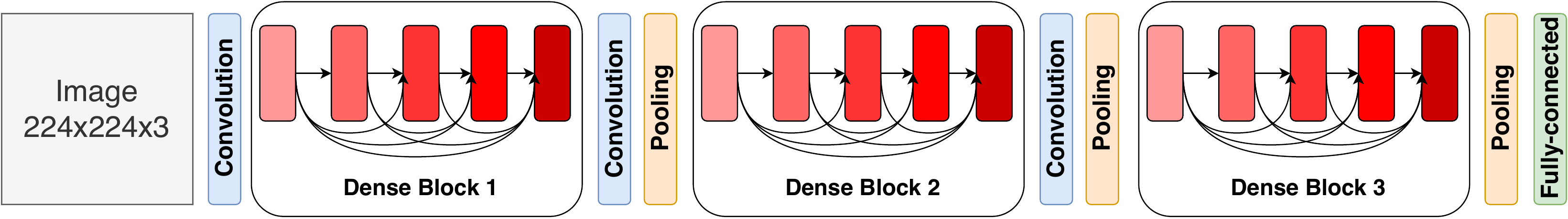}
    \caption{Densely-connected convolutional network architecture.}
    \label{fig:densenet}
\end{figure}

\paragraph{DenseNet.} Densely-connected convolutional neural network~\citep{huang2017densely}, or DenseNet in short, was proposed in 2016 to address the issue of maximum information and gradient flow. To this end, as illustrated in Fig.~\ref{fig:densenet}, in DenseNet, every layer gets as input the output of all previous layers, thus, is connected to every other layers directly. Thus, since each layer has a direct access to the gradient computed by the loss directly, the vanishing gradient problem is remedied to a certain degree.

Another good characteristic of DenseNet is that, unlike simpler architectures such as VGG16, the output of the deepest layer contains features at multiple semantic level, i.e., very rich semantic features as well as low level features representing edges or texture. 

{}

\section{Recurrent Neural Networks}
\label{sec:RNN}

Throughout this thesis, we widely use RNNs to learn the sequences in both cases of discrete and continuous anticipation tasks. In this section, we first describe the main concept of RNNs briefly, then discuss the two variants of RNNs, namely Long-Short Term Memory (LSTM)~\citep{hochreiter1997long} and Gated Recurrent Unit (GRU)~\citep{cho2014learning}, which we later use in the following chapters.

\subsection{Vanilla RNN.}
There are many tasks that are sequential in nature and often cannot be casted as the ones with fixed-length/fixed-size inputs and outputs, e.g., speech recognition~\citep{graves2013speech}, machine translation~\citep{sutskever2014sequence,maruf2019survey,chung2014empirical}, and video analysis~\citep{donahue2015long}. In a sequential problem it is crucial that the decision at each time-step being made based on both the observation at that given time-step and all previous observations since samples at different time-steps are not independent of each other. RNNs perform exactly the same task by taking a representation of previous outputs into account while processing the input of the current time-step by updating and using its \emph{hidden state}. 
At a given time, each RNN cell takes as input the representation of the current time-step and the hidden state that has been updated given the representation of all previous time-steps. Then, it updates the hidden state given the previous hidden state and the current input, and optionally can generate an output. Since RNNs require updating the hidden state at each time-step, they work sequentially.

More specifically, RNNs learn the temporal relations by mapping the input sequence to hidden states, and then mapping the hidden states to the desired output. Given $x_t$, the input at time $t$, the hidden state is updated as

\begin{align}
    h_t=g(W_{hh}h_{t-1}+W_{xh}x_t),
\end{align}
where $h_{t-1}$ is the hidden state at the previous time-step, $g$ is the nonlinearity function such as $tanh$, and $W_{hh}$ and $W_{xh}$ are the set of learnable weights. One can also generate an output, $y_t$, at each time-step by computing

\begin{align}
    y_t = g(W_{hy}h_t),
\end{align}
where $W_{hy}$ is a set of learnable weights that maps the hidden state to the output.

Although shown effective in many sequence learning problems, RNNs suffer from the vanishing/exploding gradient that makes them hard to train for the task requires learning long-range dependencies. To remedy this issue LSTM has been proposed.

\paragraph{LSTM.}
Long-short term memory unit, or LSTM in short, is a variant of RNN which is proposed to address the problem of vanishing/exploding gradient when dealing with long sequences. What makes LSTMs applicable for learning longer-range dependencies is the concept of \emph{memory}, in which it allows the model decide when to forget and when to use and update the previous hidden state given the input at the current time-step.
LSTM comes with a gating of the form

\begin{align}
    c_t = c_{t-1} + i_t \odot g_t,
\end{align}{}
where $c_t$ is a recurrent state, $i_t$ is a gating function, and $g_t$ is a full update. Unlike vanilla RNN, such gating assures that the derivatives of the loss with respect to the recurrent state $c_t$ does not vanish.

The recurrent states comprises a \emph{cell state}, $c_t$, and an \emph{output state}, $h_t$. The gate $f_t$ modulates if the cell state should be \emph{forgotten} and the gate $i_t$ modulates if the new update should be taken into account. The gate $o_t$ modules in the case that output state should be reset.  The structure of these gates and states are

\begin{itemize}
    \item Forget gate: $f_t=\sigma(W_{xf}x_t + W_{hf}h_{t-1}+b_f)$
    \item Input gate: $i_t=\sigma(W_{xi}x_t + W_{hi}h_{t-1}+b_i)$
    \item Full cell state update: $g_t=tanh(W_{xc}x_t + W_{hc}h_{t-1}+b_c)$
    \item Cell state: $c_t=f_t \odot c_{t-1} +i_t \odot g_t$
    \item Output gate: $o_t=\sigma(W_{xo}x_t + W_{ho}h_{t-1}+b_o)$
    \item Output state: $h_t = o_t \odot tanh(c_t)$
\end{itemize}{}
where $\sigma$ is the sigmoid activation function.

In standard practice, the forget gate bias $b_f$ should be initialized with a large value such that $f_t\simeq 1$ and the gating has no effect~\citep{gers1999learning}. Note that a prediction $y_t$ can be made given the hidden state $h_t$, so that is why the $h_t$ is called the output state.

\paragraph{GRU.}
The LSTM can be simplified in the a gated recurrent unit, or GRU in short, with a gating for the recurrent state and a reset gate. Therefore, one can easily have only the following gates and updates

\begin{itemize}
    \item Reset gate: $r_t=\sigma(W_{xr}x_t + W_{hr}h_{t-1}+b_r)$
    \item Forget gate: $z_t=\sigma(W_{xz}x_t + W_{hz}h_{t-1}+b_z)$
    \item Full update: $\bar{h}_t = tanh(W_{xh}x_t + W_{hh}(r_t\odot h_{t-1})+b_h)$
    \item Hidden update: $h_t=z_t \odot h_{t-1}+ (1-z_t)\odot \bar{h}_t$
\end{itemize}{}

\section{Variational Autoencoders}
Consider we have a model $p_\theta(x, z)$, that works with a set of observed variables $x$, which we have access to given a dataset, and a set of unobserved \emph{latent} variables $z$, which typically are not available in the dataset. This model $p_\theta(x, z)$ aims to model the observations, i.e., $x$ through a neural network parameterized with $\theta$. The likelihood of the data can then be computed as the marginal distribution over the observed variables

\begin{align}
    p_\theta(x) = \int_z p_\theta(x, z)dz,
\end{align}{}
which is also called the marginal likelihood or the evidence of a model. $p_\theta(x, z)$ is called a deep latent variable model, or DLVM in short, since its distributions are parameterized by a neural network $\theta$. In a simple case, one can define $p_\theta(x, z)$ as a factorization,

\begin{align}
    p_\theta(x, z) = p_\theta(z)p_\theta(x|z) .
\end{align}{}

In such case, $ p_\theta(z)$ is called the prior distribution over the latent variable $z$. Learning the maximum likelihood in such model is not straightforward since $p_\theta(x)$ is not tractable~\citep{kingma2013auto}. This is related to the intractability of the posterior distribution $p_\theta(z|x)$, which can be computed as 

\begin{align}
    p_\theta(z|x) = \frac{p_\theta(x, z)}{p_\theta(x)} .
\end{align}{}

To solve the maximum likelihood problem, we would like to have $p_\theta(x|z)$ and $p_\theta(z)$. Using \emph{Variational Inference}, we aim to approximate the true posterior $p_\theta(z|x)$ with another distribution $q_\phi(z|x)$ which is computed with another neural network parameterized with $\phi$ (called variational parameters), such that $q_\phi(z|x)\simeq p_\theta(z|x)$. Using such approximation, \emph{Variational Autoencoders}~\citep{kingma2013auto}, or VAEs in short, are able to optimize the marginal likelihood in a tractable way. The optimization objective of the VAEs is variational lower bound, also known as evidence lower bound, or ELBO in short. Recall that variational inference aims to find an approximation of the posterior that represent the true one. One way to do this is to minimize the divergence between the approximate and true posterior using Kullback–Leibler divergence~\citep{kullback1951information}, or KL divergence in short, that is,

\begin{align}
    D_{KL}\Big[q_\phi(z|x) || p_\theta(z|x)\Big] = \sum_{z\sim q_\phi(z|x)}q_\phi(z|x) \log \frac{q_\phi(z|x)}{p_\theta(z|x)} .
\end{align}{}

This can be seen as an expectation, 

\begin{align}
    D_{KL}\big[q_\phi(z|x) || p_\theta(z|x)\big] = \mathbf{E}_{z\sim q_\phi(z|x)} \bigg[\log \frac{q_\phi(z|x)}{p_\theta(z|x)}\bigg]  =  \mathbf{E}_{z\sim q_\phi(z|x)} \Big[ \log q_\phi(z|x) - \log p_\theta(z|x) \Big] .
\end{align}{}

The second term above, i.e., the true posterior, according to the Bayes' theorem, can be written as $p_\theta(z|x) = \frac{p_\theta(x|z)p(z)}{p_\theta(x)}$.
The data distribution $p_\theta(x)$ is independent of the latent variable $z$, thus, can be pulled out of the expectation term, 

\begin{align}
    D_{KL}\big[q_\phi(z|x) || p_\theta(z|x)\big] = \mathbf{E}_{z\sim q_\phi(z|x)} \Big[\log q_\phi(z|x) - \log p_\theta(x|z) - \log p(z) \Big] + \log p_\theta(x) .
\end{align}{}

By putting the $\log p_\theta(x)$ in the right hand side of the above equation, we can write,

\begin{align}
    D_{KL}\big[q_\phi(z|x) || p_\theta(z|x)\big] - \log p_\theta(x)= \mathbf{E}_{z\sim q_\phi(z|x)} \Big[\log q_\phi(z|x) - \log p_\theta(x|z) - \log p(z) \Big] \nonumber \\
    \log p_\theta(x) - D_{KL}\big[q_\phi(z|x) || p_\theta(z|x)\big] = \mathbf{E}_{z\sim q_\phi(z|x)} \Big[\log p_\theta(x|z) - \big(\log q_\phi(z|x) - \log p(z)\big) \Big] \nonumber \\
    = \mathbf{E}_{z\sim q_\phi(z|x)} \Big[\log p_\theta(x|z) \Big] - \mathbf{E}_{z\sim q_\phi(z|x)} \Big[\log q_\phi(z|x) - \log p(z) \Big] .
\end{align}{}

The second expectation term in above equation, according to definitions, is the KL divergence between the approximate posterior $q_\phi(z|x)$ and the prior $\log p(z)$ distributions. Thus, this can be written as 
\begin{align}
    \log p_\theta(x) - D_{KL}\big[q_\phi(z|x) || p_\theta(z|x)\big] = \mathbf{E}_{z\sim q_\phi(z|x)} \big[\log p_\theta(x|z) \big] - D_{KL}\big[q_\phi(z|x) || p(z)\big] .
\end{align}{}

In above equation, $\log p_\theta(x)$ is the log-likelihood of the data which we would like to optimize, $D_{KL}\big[q_\phi(z|x) || p_\theta(z|x)\big]$ is the KL divergence between the approximate and the true posterior distributions, which is not tractable to compute, but, according to definitions, we know that is non-negative, $\mathbf{E}_{z\sim q_\phi(z|x)} \big[\log p_\theta(x|z) \big]$ is the reconstruction loss, and $D_{KL}\big[q_\phi(z|x) || p(z)\big]$ is the KL divergence between the approximate posterior distribution and a prior over the latent variable. The last term can be seen as a regularization term over the latent representation. Therefore, the intractability and non-negativity of $D_{KL}\big[q_\phi(z|x) || p_\theta(z|x)\big]$ only allows us to optimize the lower bound on the log-likelihood of the data, 

\begin{align}
\log p_\theta(x) \geq \mathbf{E}_{z\sim q_\phi(z|x)} \big[\log p_\theta(x|z) \big] - D_{KL}\big[q_\phi(z|x) || p(z)\big],
\end{align}{}
which we call variational or evidence lower bound (ELBO). 

\begin{figure}
    \centering
    \includegraphics[width=0.75\textwidth]{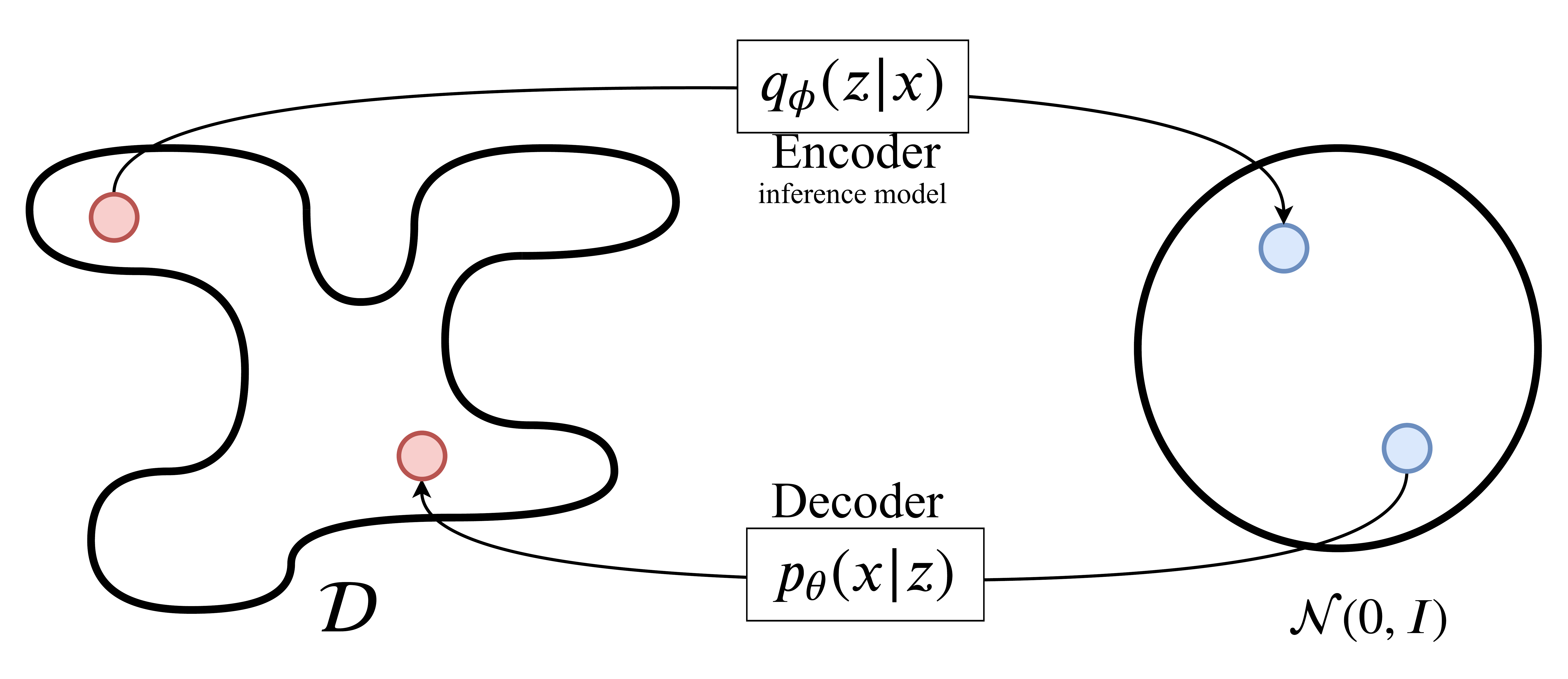}
    \caption{In a standard VAE, the model learns stochastic mappings between the observed dataset whose empirical distribution is $p_{\mathcal{D}}(x)$, typically a complicated one, and a latent space whose distribution is relatively simpler, e.g., $\mathcal{N}(0,I)$.}
    \label{fig:my_label}
\end{figure}{}
In practice, VAEs first learn to generate a latent variable $z$ given the data $x$, i.e., approximate the posterior distribution $q_\phi(z|x)$, the encoder, whose goal is to model the variation of the data. From this latent random variable $z$, VAEs then generate a sample $x$ by learning $p_\theta(x|z)$, the decoder, whose goal is to maximize the log likelihood of the data.

These two networks, i.e., the encoder ($q_\phi(z|x)$) and the decoder ($p_\theta(x|z)$), are trained jointly, using a prior over the latent variable, as mentioned above. This prior is usually the standard Normal distribution, $\mathcal{N}(0,I)$. In practice, the posterior distribution is approximated by a Gaussian $\mathcal{N}(\mu, \sigma^2 I)$, whose parameters are output by the encoder. To facilitate optimization, the reparameterization trick~\citep{kingma2013auto} is used. That is, the latent variable is computed as $z=\mu + \sigma\odot \epsilon$, where $\epsilon$ is a vector sampled from the standard Normal distribution.

As an extension to VAEs, conditional VAEs, or CVAEs~\citep{sohn2015learning} in short, use auxiliary information, i.e., the conditioning variable or observation, to generate the data $x$. In the standard setting, both the encoder and the decoder are conditioned on the conditioning variable $c$. That is, the encoder is denoted as $q_\phi(z|x,c)$ and the decoder as $p_\theta(x|z,c)$. Then, the objective of the model becomes t
\begin{align}
    \log p_\theta(x|c) \geq \mathbf{E}_{z\sim q_\phi(z|x)}\big[\log p_\theta(x|z,c)\big] - D_{KL}\big[q_\phi(z|x, c) || p(z|c)\big]\;.
\end{align}

In practice, conditioning is typically done by concatenation; the input of the encoder is the concatenation of the data $x$ and the condition $c$, i.e., $q_\phi(z| \left[x,c\right])$, and that of the decoder the concatenation of the latent variable $z$ and the condition $c$, i.e., $p_\theta(x|\left[z,c\right])$. Thus, the prior distribution is still $p(z)$, and the latent variable is sampled independently of the conditioning one. It is then left to the decoder to combine the information from the latent and conditioning variables to generate a data sample.

%% file: chapter1.tex
\chapter{General Action Anticipation}
\label{cha:encouraging_lstms}

In this chapter, we focus on action anticipation, the task of predicting one discrete representation, i.e., action label, of a deterministic future.

In contrast to the widely studied problem of recognizing an action given a complete sequence, action anticipation aims to identify the action from only partially available videos. As such, it is therefore key to the success of computer vision applications requiring to react as early as possible, such as autonomous navigation. In this chapter, we propose a new action anticipation method that achieves high prediction accuracy even in the presence of a very small percentage of a video sequence. To this end, we develop a multi-stage LSTM architecture that leverages context-aware and action-aware features, and introduce a novel loss function that encourages the model to predict the correct class as early as possible. Our experiments on standard benchmark datasets evidence the benefits of our approach; We outperform the state-of-the-art action anticipation methods for early prediction by a relative increase in accuracy of 22.0\% on JHMDB-21, 14.0\% on UT-Interaction and 49.9\% on UCF-101.

\section{Introduction}

Understanding actions from videos is key to the success of many real-world applications, such as autonomous navigation and sports analysis. While great progress has been made to recognize actions from complete sequences~\citep{donahue2015long,wang2016temporal,fernando2016discriminative,bilen2016dynamic,fernando2016rank,wang2013action} in the past decade, action anticipation~\citep{ryoo2011human,ma2016learning,soomro2016predicting}, which aims to predict the observed action as early as possible, has become a popular research problem only recently. 
Anticipation is crucial in scenarios where one needs to react before the action is finalized, such as to avoid hitting pedestrians with an autonomous car, or to forecast dangerous situations in surveillance scenarios.

\begin{figure}
\centering
\includegraphics[width=0.7\textwidth]{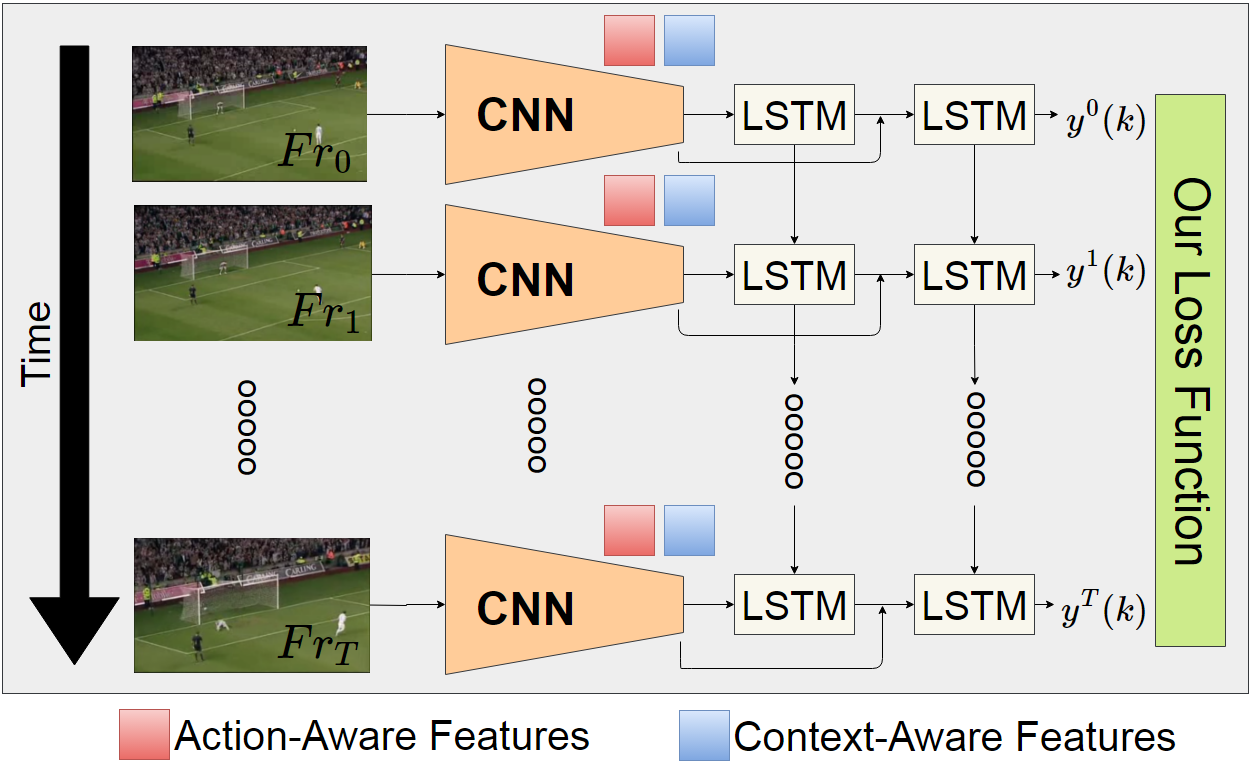}
\caption{
{\bf Overview of our approach.} Given a small portion of sequential data, our approach is able to predict the action category with very high performance. For instance, in UCF-101, our approach anticipates actions with more than 80\% accuracy given only the first 1\% of the video. To achieve this, we design a a model that leverages action- and context-aware features together with a new loss function that encourages the model to make correct predictions as early as possible.
}
\label{FIG:LOGO}
\end{figure}

The key difference between recognition and anticipation lies in the fact that the methods tackling the latter should predict the correct class as early as possible, given only a few frames from the beginning of the video sequence. To address this, several approaches have introduced new training losses encouraging the score~\citep{soomro2016online} or the rank~\citep{ma2016learning} of the correct action to increase with time, or penalizing increasingly strongly the classification mistakes~\citep{jain2016recurrent}. In practice, however, the effectiveness of these losses remains limited for very early prediction, such as from 1\% of the sequence.

In this chapter, we introduce a novel loss that encourages making correct predictions very early. Specifically, our loss models the intuition that some actions, such as running and high jump, are highly ambiguous after seeing only the first few frames, and false positives should therefore not be penalized too strongly in the early stages. By contrast, we would like to predict a high probability for the correct class as early as possible, and thus penalize false negatives from the beginning of the sequence. Our experiments demonstrate that, for a given model, our new loss yields significantly higher accuracy than existing ones on the task of early prediction.

In particular, in this chapter, we also contribute a novel multi-stage Long Short Term Memory (LSTM) architecture for action anticipation. This model effectively extracts and jointly exploits context- and action-aware features (see Fig.~\ref{FIG:LOGO}). This is in contrast to existing methods that typically extract either global representations for the entire image~\citep{diba2016deepcamp,wang2016temporal,donahue2015long} or video sequence~\citep{tran2015learning,karpathy2014large}, thus not focusing on the action itself, or localize the feature extraction process to the action itself via dense trajectories~\citep{TrajectoryPooled,IDT,DiscriminativeRankPooling}, optical flow~\citep{CNN2Stream,TSN,VLAD3} or actionness~\citep{actionness,ActionnessRanking,ActionTubelets,FastActionProposal,OnlineSEEDS,SpatioTemporalProposal}, thus failing to exploit contextual information. To the best of our knowledge, only two-stream networks~\citep{TwoStreamNIPS,CNN2Stream,cheron2015p,SpatioTemporalLSTM} have attempted to jointly leverage both information types by making use of RGB frames in conjunction with optical flow to localize the action. Exploiting optical flow, however, does not allow these methods to explicitly leverage appearance in the localization process.
Furthermore, computing optical flow is typically expensive, thus significantly increasing the runtime of these methods. By not relying on optical flow, our method is significantly more efficient: On a single GPU, our model analyzes a short video (e.g., 50 frames) 14 times faster than~\citep{TwoStreamNIPS} and~\citep{CNN2Stream}.

Our model is depicted in Fig.~\ref{FIG:LSTM}. In a first stage, it focuses on the global, context-aware information by extracting features from the entire RGB image. The second stage then combines these context-aware features with action-aware ones obtained by exploiting class-specific activations, typically corresponding to regions where the action occurs. In short, our model first extracts the contextual information, and then merges it with the localized one. 

As evidenced by our experiments, our approach significantly outperforms the state-of-the-art action anticipation methods on all the standard benchmark datasets that we evaluated on, including UCF-101~\citep{UCF101}, UT-Interaction~\citep{ryoo2010overview}, and JHMDB21~\citep{JHMDB}. We further show that our combination of context- and action-aware features is also beneficial for the more traditional task of action recognition. Moreover, we evaluate the effect of optical flow features for both  action recognition and anticipation.

\section{Related Work}

The focus of this chapter is twofold: Action anticipation, with a new loss that encourages correct prediction as early as possible, and action modeling, with a model that combines context- and action-aware information using multi-stage LSTMs. Below, we discuss the most relevant approaches for these two aspects.

\subsection{Action Anticipation}

The idea of action anticipation was introduced by~\citep{ryoo2009spatio}, which models causal relationships to predict human activities.
This was followed by several attempts to model the dynamics of the observed actions, such as by introducing integral and dynamic bag-of-words~\citep{ryoo2011human}, using spatial-temporal implicit shape models~\citep{yu2012predicting}, extracting human body movements via skeleton information~\citep{zhao2013online}, and accounting for the complete and partial history of observed features~\citep{kong2014discriminative}.

More recently,~\citep{soomro2016predicting} proposed to make use of binary SVMs to classify video snippets into sub-action categories and obtain the final class label in an online manner using dynamic programming. To overcome the need to train one classifier per sub-action,~\citep{soomro2016online} extended this approach to using a structural SVM. Importantly, this work further introduced a new objective function to encourage the score of the correct action to increase as time progresses.

While the above-mentioned work made use of handcrafted features, recent advances have naturally led to the development of deep learning approaches to action anticipation. In this context,~\citep{ma2016learning} proposed to combine a Convolutional Neural Network (CNN) with an LSTM to model both spatial and temporal information. The authors further introduced new ranking losses whose goal is to enforce either the score of the correct class or the margin between the score of the correct class and that of the best score to be non-decreasing over time. Similarly, in~\citep{brain4Cars}, a new loss that penalizes classification mistakes increasingly strongly over time was introduced in an LSTM-based framework that used multiple modalities. While the two above-mentioned methods indeed aim at improving classification accuracy over time, they do not explicitly encourage making correct predictions as early as possible. By contrast, while accounting for ambiguities in early stages, our new loss still aims to prevent false negatives from the beginning of the sequence.

Instead of predicting the future class label, in~\citep{vondrick2016anticipating}, the authors proposed to predict the future visual representation. However, the main motivation for this was to work with unlabeled videos, and the learned representation is therefore not always related to the action itself. 

\begin{figure*}[t!]
\centering
\includegraphics[width=0.9\textwidth]{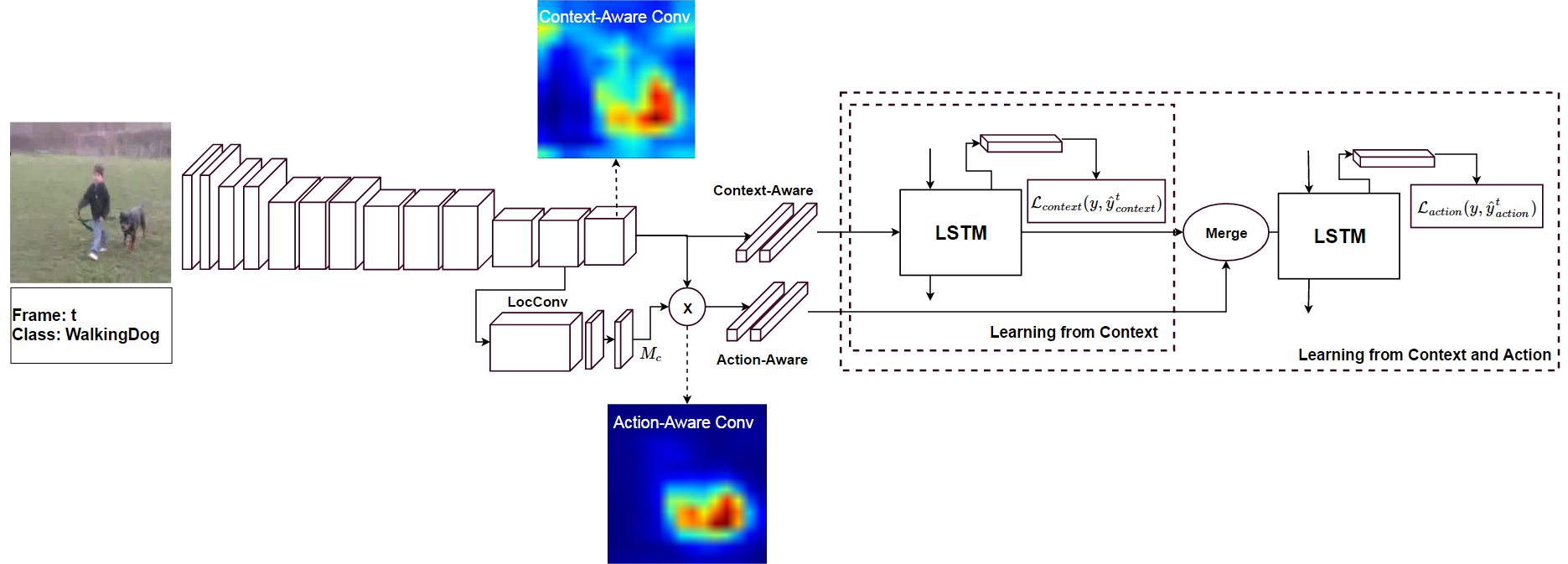}
\caption{{\bf Overview of our approach.} We propose to extract context-aware features, encoding global information about the scene, and combine them with action-aware ones, which focus on the action itself. To this end, we introduce a multi-stage LSTM architecture that leverages the two types of features to predict the action or forecast it. Note that, for the sake of visualization, the color maps were obtained from 3D tensors (\(512\times W\times H\)) via an average pooling operation over the 512 channels.
}
\label{FIG:LSTM}
\end{figure*}

\subsection{Action Modeling}
Most recent action approaches extract global representations for the entire image~\citep{DeepCAMP,ActionTransformation,LRCN} or video sequence~\citep{3DCNN,LargeScaleCNN}. As such, these methods do not truly focus on the actions of interest, but rather compute a \emph{context-aware} representation. Unfortunately, context does not always bring reliable information about the action. For example, one can play guitar in a bedroom, a concert hall or a yard. 
To overcome this, some methods localize the feature extraction process by exploiting dense trajectories~\citep{TrajectoryPooled,IDT,DiscriminativeRankPooling} or optical flow~\citep{VLAD3}. 
Inspired by objectness, the notion of actionness~\citep{actionness,ActionnessRanking,ActionTubelets,FastActionProposal,OnlineSEEDS,SpatioTemporalProposal} has recently also been proposed to localize the regions where a generic action occurs. The resulting methods can then be thought of as extracting \emph{action-aware} representations. In other words, these methods go to the other extreme and completely discard the notion of context which can be useful for some actions, such as playing football on a grass field.

There is nevertheless a third class of methods that aim to leverage these two types of information~\citep{TwoStreamNIPS,CNN2Stream,cheron2015p,SpatioTemporalLSTM,TSN}. By combining RGB frames and optical flow in two-stream architectures, these methods truly exploit context and motion, from which the action can be localized by learning to distinguish relevant motion. This localization, however, does not directly exploit appearance. Here, inspired by the success of these methods, we develop a novel multi-stage network that also leverages context- and action-aware information. However, we introduce a new action-aware representation that exploits the RGB data to localize the action. As such, our approach effectively leverages appearance for action-aware modeling, and, by avoiding the expensive optical flow computation, is much more efficient than the above-mentioned two-stream models.
In particular, our model is about 14 times faster than the state-of-the-art two-stream network of~\citep{CNN2Stream} and has less parameters. The reduction in number of parameters is due to the fact that~\citep{CNN2Stream} and~\citep{TwoStreamNIPS} rely on two VGG-like networks (one for each stream) with a few additional layers (including 3D convolutions for [8]). By contrast, our model has, in essence, a single VGG-like architecture, with some additional LSTM layers, which only have few parameters. Moreover, our work constitutes the first attempt at explicitly leveraging context- and action-aware information for action anticipation. Finally, we introduce a novel multi-stage LSTM fusion strategy to integrate action and context aware features.

\section{Our Approach}
Our goal is to predict the class of an action as early as possible, that is, after having seen only a very small portion of the video sequence. To this end, we first introduce a new loss function that encourages making correct predictions very early. We then develop a multi-stage LSTM model that makes use of this loss function while leveraging both context- and action-aware information.

\subsection{A New Loss for Action Anticipation}
\label{sec:loss}
As argued above, a loss for action anticipation should encourage having a high score for the correct class as early as possible. However, it should also account for the fact that, early in the sequence, there can often be ambiguities between several actions, such as running and high jump. Below, we introduce a new anticipation loss that follows these two intuitions.

Specifically, let $y^t(k)$ encode the true activity label at time $t$, i.e., $y^t(k) = 1$ if the sample belongs to class $k$ and 0 otherwise, and $\hat{y}^t(k)$ denote the corresponding label predicted by a given model. We define our new loss as
\begin{align}
\mathcal{L}(y, \hat{y}) = -\frac{1}{N}\sum^N_{k=1}\sum^T_{t=1}\Bigg[y^t(k) \log(\hat{y}^t(k)) +  \frac{t(1-y^t(k))}{T}\log(1-\hat{y}^t(k))\Bigg]\;,
\label{eq:loss}
\end{align}
where $N$ is the number of action classes and $T$ the length (number of frames) of the input sequence.

This loss function consists of two terms. The first one penalizes false negatives with the same strength at any point in time. By contrast, the second one focuses on false positives, and its strength increases linearly over time, to reach the same weight as that on false negatives.
Therefore, the relative weight of the first term compared to the second one is larger at the beginning of the sequence. Altogether, this encourages predicting a high score for the correct class as early as possible, i.e., preventing false negatives, while accounting for the potential ambiguities at the beginning of the sequence, which give rise to false positives. As we see more frames, however, the ambiguities are removed, and these false positives are encouraged to disappear.


Our new loss matches the desirable properties of an action anticipation loss. In the next section, we introduce a novel multi-stage architecture that makes use of this loss.

\subsection{Multi-stage LSTM Architecture}
To tackle action anticipation, we develop the novel multi-stage recurrent architecture based on LSTMs depicted by Fig.~\ref{FIG:LSTM}. This architecture consists of a stage-wise combination of context- and action-aware information. Below, we first discuss our approach to extracting these two types of information, and then present our complete multi-stage recurrent network.


\begin{figure}
\centering
\includegraphics[width=0.8\textwidth]{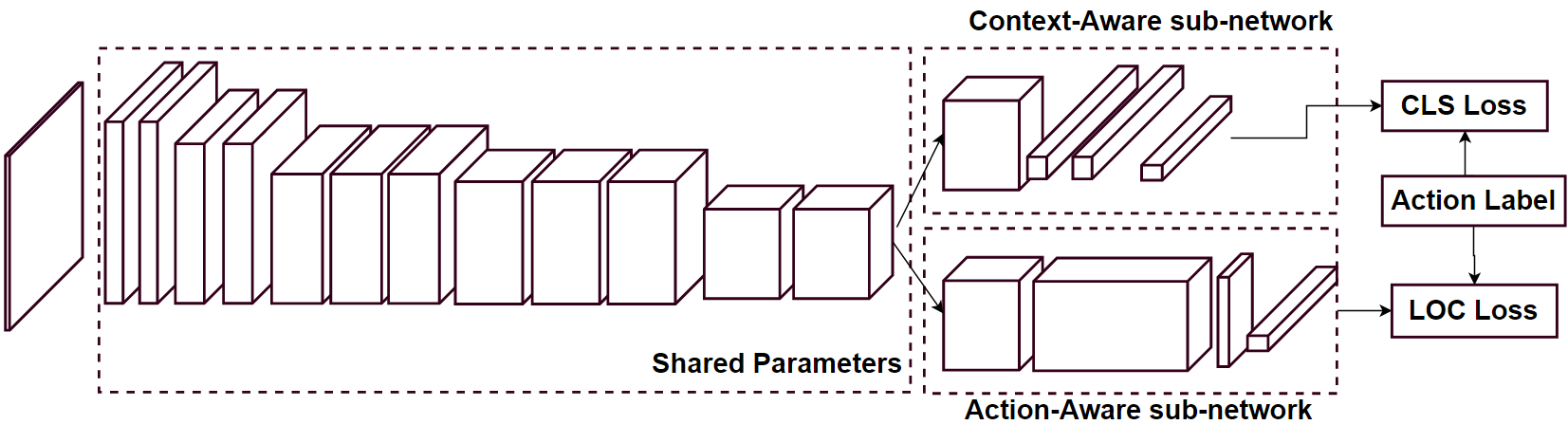}
\caption{{\bf Our feature extraction network.} Our CNN model for feature extraction is based on the VGG-16 structure with some modifications. Up to conv5-2, the network is the same as VGG-16. The output of this layer is connected to two sub-models. The first one extracts context-aware features by providing a global image representation. The second one relies on another network to extract action-aware features.}
\label{FIG:LOC_TRAIN}
\end{figure}

\subsubsection{Context- and Action-aware Modeling}
To model the context- and action-aware information, we introduce the two-stream architecture shown in Fig.~\ref{FIG:LOC_TRAIN}. The first part of this network is shared by both streams and, up to conv5-2, corresponds to the VGG-16 network~\citep{VGG}, pre-trained on ImageNet for object recognition. The output of this layer is connected to two sub-models: One for context-aware features and the other for action-aware ones. We then train these two sub-models for the same task of action recognition from a single image using a cross-entropy loss function defined on the output of each stream. In practice, we found that training the entire model in an end-to-end manner did not yield a significant improvement over training only the two sub-models. In our experiments, we therefore opted for this latter strategy, which is less expensive computationally and memory-wise. Below, we first discuss the context-aware sub-network and then turn to the action-aware one.

\paragraph{Context-Aware Feature Extraction.}
This sub-model is similar to VGG-16 from conv5-3 up to the last fully-connected layer, with the number of units in the last fully-connected layer changed from 1000 (original 1000-way ImageNet classification model) to the number of activities $N$. 

In essence, this sub-model focuses on extracting a deep representation of the whole scene for each activity and thus incorporates context. We then take the output of its fc7 layer as our context-aware features.

\paragraph{Action-Aware Feature Extraction.}
As mentioned before, the context of an action does not always correlate with the action itself. Our second sub-model therefore aims at extracting features that focus on the action itself. To this end, we draw inspiration from the object classification work of~\citep{zhou2015learning}. At the core of this work lies the idea of Class Activation Maps (CAMs). In our context, a CAM indicates the regions in the input image that contribute most to predicting each class label. In other words, it provides information about the location of an action. Importantly, this is achieved without requiring any additional annotations.

More specifically, CAMs are extracted from the activations in the last convolutional layer in the following manner. Let \(f_l(x,y)\) represent the activation of unit \(l\) in the last convolutional layer at spatial location \((x,y)\). A score $S_k$ for each class $k$ can be obtained by performing global average pooling~\citep{NetinNetGAP} to obtain, for each unit $l$, a feature \(F^l = \sum_{x,y}{f_l(x,y)}\), followed by a linear layer with weights $\{w_l^k\}$. That is, \(S_k = \sum_k{w_l^k F_l}\). A CAM for class $k$ at location \((x,y)\) can then be computed as
\begin{equation}
M_k(x,y) = \sum_l{w_l^k f_l(x,y)}\;,
\end{equation}
which indicates the importance of the activations at location \((x,y)\) in the final score for class $k$.

\begin{figure}
\centering
\includegraphics[width=0.8\textwidth]{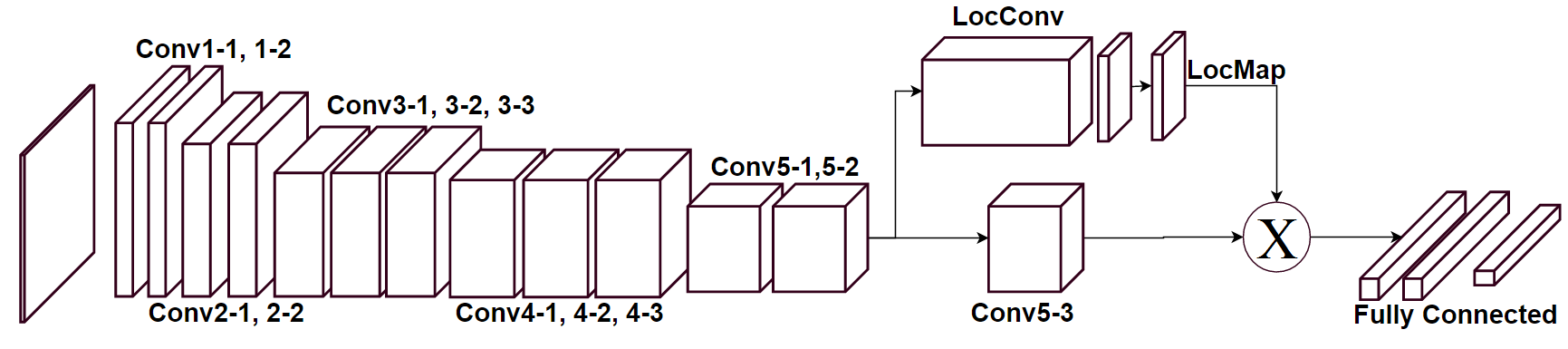}
\caption{{\bf Action-aware feature extraction.} Given the fine-tuned feature extraction network, we introduce a new layer that alters the output of conv5-3. This lets us filter out the conv5-3 features that are irrelevant, to focus on the action itself. Our action-aware features are taken as the output of the last fully-connected layer.}
\label{FIG:LOC_TEST}
\end{figure}

Here, we propose to make use of the CAMs to extract action-aware features. To this end, we use the CAMs in conjunction with the output of the conv5-3 layer of the model. The intuition behind this is that conv5-3 extracts high-level features that provide a very rich representation of the image~\citep{UnderstandingCNN} and typically correspond to the most discriminative parts of the object~\citep{DeepEdge,BuiltinFGBG}, or, in our case, the action. Therefore, we incorporate a new layer to our sub-model, whose output can be expressed as
\begin{equation}
A_k(x,y) = {\rm conv_{5-3}}(x,y) \times {\rm ReLU}(M_k(x,y))\;,
\end{equation}
where ${\rm ReLU}(M_k(x,y)) = {\rm max}(0, M_k(x,y))$. As shown in Fig.~\ref{FIG:LOC_TEST}, this new layer is then followed by fully-connected ones, and we take our action-aware features as the output of the corresponding fc7 layer.

\subsubsection{Sequence Learning for Action Anticipation}
\label{sec:seq_learn}
To effectively combine the information contained in the context-aware and action-aware features described above, we design the novel multi-stage LSTM model depicted by Fig.~\ref{FIG:LSTM}. This model first focuses on the context-aware features, which encode global information about the entire image. It then combines the output of this first stage with our action-aware features to provide a refined class prediction.


To train this model for action anticipation, we make use of our new loss introduced in Section~\ref{sec:loss}. Therefore, ultimately, our network models long-range temporal information and yields increasingly accurate predictions as it processes more frames. 

Specifically, we write the overall loss of our model as
\begin{equation}
\mathcal{L}_o = \frac{1}{V} \sum_{i=1}^V \left(\mathcal{L}_{c,i} + \mathcal{L}_{a,i}\right)\;,
\label{eq.lstm.overall}
\end{equation}
where $V$ is the total number of training sequences. This loss function combines losses corresponding to the context-aware stage and to the action-aware stage, respectively.
Below, we discuss these two stages in more detail.

\paragraph{Learning Context.}
The first stage of our model takes as input our context-aware features, and passes them through a layer of LSTM cells followed by a fully-connected layer that, via a softmax operation, outputs a probability for each action class. Let $\hat{y}_{c,i}$ be the vector of probabilities for all classes and all time steps predicted by the first stage for sample $i$. We then define the loss for a single sample as 
\begin{equation}
\mathcal{L}_{c,i} = \mathcal{L}(y_{i},\hat{y}_{c,i})\;,
\end{equation}
where $\mathcal{L}(\cdot)$ is our new loss defined in Eq.~\ref{eq:loss}, and $y_{i}$ is the ground-truth class label for sample $i$.


\paragraph{Learning Context and Action.}
The second stage of our model aims at combining context-aware and action-aware information. Its structure is the same as that of the first stage, i.e., a layer of LSTM cells followed by a fully-connected layer to output class probabilities via a softmax operation. However, its input merges the output of the first stage with our action-aware features. This is achieved by concatenating the hidden activations of the LSTM layer with our action-aware features. We then make use of the same loss function as before, but defined on the final prediction. For sample $i$, this can be expressed as
\begin{equation}
\mathcal{L}_{a,i} = \mathcal{L}(y_i, \hat{y}_{{a,i}})\;,
\end{equation}
where $\hat{y}_{a}$ is the vector of probabilities for all classes predicted by the second stage.


\paragraph{Inference.}
At inference, the input RGB frames are forward-propagated through our model. We therefore obtain a probability vector for each class at each frame. While one could simply take the probabilities in the current frame $t$ to obtain the class label at time $t$, via $argmax$, we propose to increase robustness by leveraging the predictions of all the frames up to time $t$. To this end, we make use of an average pooling of these predictions over time.

\section{Experiments}
In this section, we first compare our method with state-of-the-art techniques on the task of action anticipation, and then analyze various aspects of our model, such as the influence of the loss function and of the different feature types. In the supplementary material, we provide additional experiments to analyze the effectiveness of different LSTM architectures, and the influence of the number of hidden units and of our temporal average pooling strategy. We also report the performance of our method on the task of action recognition from complete videos with and without optical flow, and action anticipation with optical flow.

\subsection{Datasets}
For our experiments, we made use of the standard UCF-101~\citep{UCF101}, UT-Interaction~\citep{ryoo2010overview}, and JHMDB-21~\citep{JHMDB} benchmarks, which we briefly describe below.

The UCF-101 dataset consists of 13,320 videos (each contains a single action) of 101 action classes including a broad set of activities such as sports, playing musical instruments and human-object interaction, with an average length of 7.2 seconds. UCF-101 is one of the most challenging datasets due to its large diversity in terms of actions and to the presence of large variations in camera motion, cluttered background and illumination conditions. There are three standard training/test splits for this dataset. In our comparisons to the state-of-the-art for both action anticipation and recognition, we report the average accuracy over the three splits. For the detailed analysis of our model, however, we rely on the first split only.

The JHMDB-21 dataset is another challenging dataset of realistic videos from various sources, such as movies and web videos, containing 928 videos and 21 action classes. Similarly to UCF-101, in our comparison to the state-of-the-art, we report the average accuracy over the three standard splits of data. Similar to UCF-101 dataset, each video contains one action starting from the beginning of the video.

The UT-Interaction dataset contains videos of continuous executions of 6 human-human interaction classes: shake-hands, point, hug, push, kick and punch. 
The dataset contains 20 video sequences whose length is about 1 minute each. Each video contains at least one execution of each interaction type, providing us with 8 executions of human activities per video on average. Following the recommended experimental setup, we used 10-fold leave-one-out cross validation for each of the standard two sets of 10 videos. That is, within each set, we leave one sequence for testing and use the remaining 9 for training. 
Following standard practice, we also made use of the annotations provided with the dataset to split each video into sequences containing individual actions.


\subsection{Implementation Details}
\paragraph{CNN and LSTM Configuration.}
The parameters of the CNN were optimized using stochastic gradient descent with a fixed learning rate of 0.001, a momentum of 0.9, a weight decay of 0.0005, and mini-batches of size 32. To train our LSTMs, we similarly used stochastic gradient descent with a fixed learning rate of 0.001, a momentum of 0.9, and mini-batch size of 32. For all LSTMs, we used 2048 hidden units. To implement our method, we used Python and Keras~\citep{keras}. We will make our code publicly available. 

\paragraph{Training Procedure.}
To fine-tune the network on each dataset, we augment the data, so as to reduce the effect of over-fitting. The input images were randomly flipped horizontally and rotated by a random amount in the range -8 to 8 degrees. We then extracted crops according to the following procedure:
\textbf{(1)} Compute the maximum cropping rectangle with a given aspect ratio ($320/240$) that fits inside the input image.
\textbf{(2)} Scale the width and height of the cropping rectangle by a factor randomly selected in the range $0.8$-$1$.
\textbf{(3)} Select a random location for the cropping rectangle within the original input image and extract the corresponding subimage.
\textbf{(4)} Scale the subimage to $224 \times 224$.

After these geometric transformations, we further applied RGB channel shifting~\citep{wu2015deep}, followed by randomly adjusting image brightness, contrast and saturation with a factor $\alpha=0.3$. The operations are: for brightness, $\alpha \times Image$, for contrast, $Image \times \alpha + (1.0 - \alpha)\times mean(grey(Image))$, and for saturation, $Image\times \alpha + (1.0 - \alpha)\times grey(Image)$. 


\subsection{Comparison to the State-of-the-Art}
We compare our approach to the state-of-the-art action anticipation results reported on each of the three datasets discussed above.
We further complement these state-of-the-art results with additional baselines that make use of our context-aware features with the loss of either~\citep{ma2016learning} or~\citep{brain4Cars}. Note that a detailed comparison of different losses within our model is provided in Section~\ref{sec:res_loss}.

Following standard practice, we report the so-called \emph{earliest} and \emph{latest} prediction accuracies. Note, however, that there is no real agreement on the proportion of frames that the \emph{earliest} setting corresponds to. For each dataset, we make use of the proportion that has been employed by the baselines (i.e., either 20\% or 50\%). Note also that our approach relies on at most $T$ frames (with $T=50$ in practice). Therefore, in the \emph{latest} setting, where the baselines rely on the complete sequences, we only exploit the first $T$ frames. We believe that the fact that our method significantly outperforms the state-of-the-art in this setting despite using less information further evidences the effectiveness of our approach.

\paragraph{JHMDB-21.}
The results for the JHMDB-21 dataset are provided in Table~\ref{tab:jhmdb}. In this case, following the baselines, earliest prediction corresponds to observing the first 20\% of the sequence. Note that we clearly outperform all the baselines by a significant margin in both the earliest and latest settings. Remarkably, we also outperform the methods that rely on additional information as input, such as optical flow~\citep{soomro2016online,soomro2016predicting,ma2016learning} and Fisher vector features based on Improved Dense Trajectories~\citep{soomro2016online}. This clearly demonstrates the benefits of our approach for anticipation.

\begin{table}[!h]
\centering
\small
\caption{Comparison with state-of-the-art baselines on the task of action anticipation on the JHMDB-21 dataset. Note that our approach outperforms all baselines significantly in both settings.
}
\label{tab:jhmdb}
\begin{tabular}{l c c c}
\hline
Method & Earliest & Latest\\
\hline
DP-SVM~\citep{soomro2016online} & 5\% & 46\% \\
S-SVM~\citep{soomro2016online} & 5\% & 43\% \\
Where/What~\citep{soomro2016predicting} & 10\% & 43\%\\
Ranking Loss~\citep{ma2016learning} & 29\% & 43\% \\
Context-Aware+Loss of~\citep{brain4Cars} & 28 \% & 43\% \\
Context-Aware+Loss of~\citep{ma2016learning} & 33\% & 39\% \\
\hline
Ours & \textbf{55\%} & \textbf{58\%} \\
\hline
\end{tabular}
\end{table}

\begin{table}[!h]
\centering
\small
\caption{Comparison with state-of-the-art baselines on the task of action anticipation on the UT-Interaction dataset.}
\label{tab:ut}
\begin{tabular}{l c c c}
\hline
Method & Earliest & Latest \\
\hline
D-BoW~\citep{ryoo2011human} & 70.0\% & 85.0\% \\
I-BoW~\citep{ryoo2011human} & 65.0\% & 81.7\% \\
CuboidSVM~\citep{ryoo2010overview} & 31.7\% & 85.0\% \\
BP-SVM~\citep{laviers2009improving} & 65.0\% & 83.3\% \\
CuboidBayes~\citep{ryoo2011human} & 25.0\% & 71.7\% \\
DP-SVM~\citep{soomro2016online} & 13.0\% & 14.6\%\\
S-SVM~\citep{soomro2016online} & 11.0\% & 13.4\% \\
Context-Aware+Loss of~\citep{brain4Cars} & 45.0 \% & 65.0\% \\
Context-Aware+Loss of~\citep{ma2016learning} & 48.0\% & 60.0\% \\
\hline
Ours & \textbf{84.0\%} & \textbf{90.0\%}\\
\hline
\end{tabular}
\end{table}

\begin{table}[!h]
\centering
\small
\caption{Action Anticipation on the UCF-101 dataset. 
}
\label{tab:ucf}
\begin{tabular}{l c c c}
\hline
Method & Earliest & Latest \\
\hline
Context-Aware+Loss of~\citep{brain4Cars} & 30.6 \% & 71.1\% \\
Context-Aware+Loss of~\citep{ma2016learning} & 22.6\% & 73.1\% \\
\hline
Ours & \textbf{80.5\%} & \textbf{83.4\%} \\
\hline
\end{tabular}
\end{table}

\paragraph{UT-Interaction.}
We provide the results for the UT-Interaction dataset in Table~\ref{tab:ut}. Here, following standard practice, 50\% of  the sequence was observed for earliest prediction, and the entire sequence for latest prediction. Recall that our approach uses at most $T=50$ frames for prediction in both settings, while the average length of a complete sequence is around 120 frames. Therefore, as evidenced by the results, our approach yields significantly higher accuracy despite using considerably less data as input.





\paragraph{UCF-101.}
We finally compare our approach with our two baselines on the UCF-101 dataset. While this is not a standard benchmark for action anticipation, this experiment is motivated by the fact that this dataset is relatively large, has many classes, with similarity across different classes, and contains variations in video capture conditions. Altogether, this makes it a challenging dataset to anticipate actions, especially when only a small amount of data is available. The results on this dataset are provided in Table~\ref{tab:ucf}. Here, the earliest setting corresponds to using the first 2 frames of the sequences, which corresponds to around 1\% of the data. Again, we clearly outperform the two baselines consisting of exploiting context-aware features with the loss of either~\citep{ma2016learning} or~\citep{brain4Cars}. We believe that this further evidences the benefits of our approach, which leverages both context- and action-aware features with our new anticipation loss. A detailed evaluation of the influence of the different feature types and losses is provided in the next section.


\subsection{Analysis}
In this section, we provide a more detailed analysis of the influence of our loss function and of the different feature types on anticipation accuracy. Finally, we also provide a visualization of our different feature types, to illustrate their respective contributions.


\begin{figure}
\centering
\includegraphics[width=0.8\textwidth]{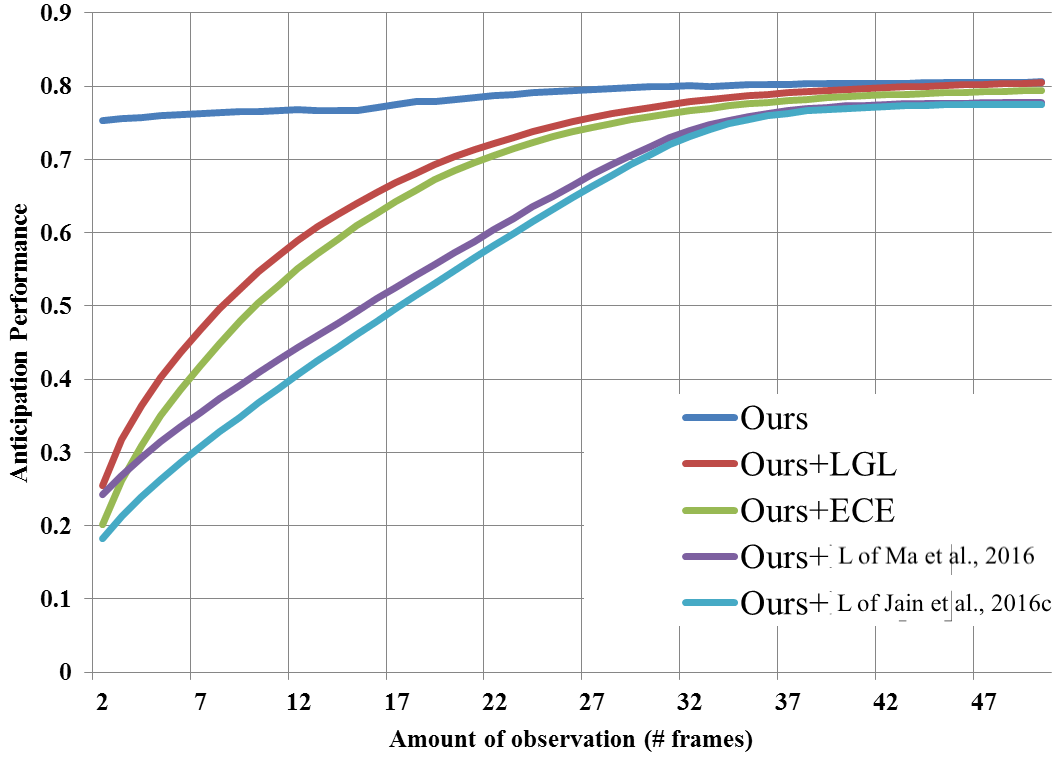}
\caption{{\bf Comparison of different losses for action anticipation on UCF-101.} We evaluate the accuracy of our model trained with different losses as a function of the number of frames observed.
This plot clearly shows the superiority of our loss function.}
\label{fig:anticipation}
\end{figure}

\begin{figure}
\centering
\includegraphics[width=0.8\textwidth]{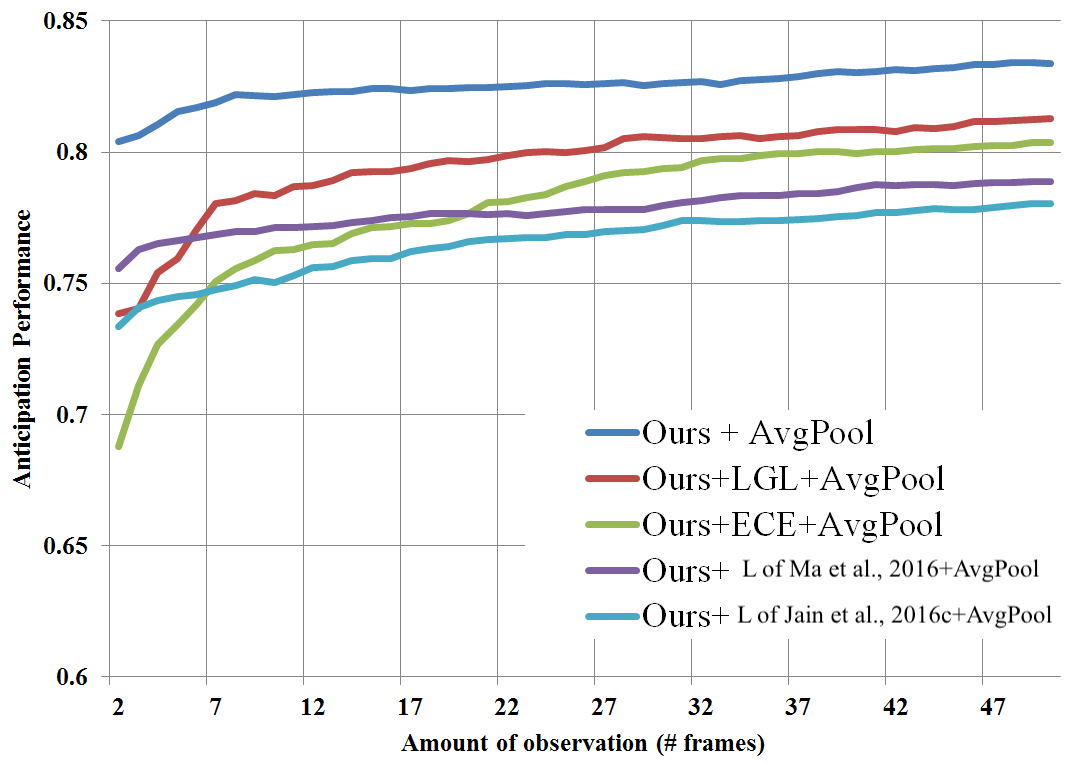}
\caption{{\bf Influence of our average pooling strategy.} Our simple yet effective average pooling leverages the predictions of all the frames up to time $t$. As shown on the UCF-101 dataset, this increases anticipation performance, especially at very early stages. 
}
\label{fig:anticipation_pool}
\end{figure}

\subsubsection{Influence of the Loss Function}
\label{sec:res_loss}
Throughout the chapter, we have argued that our novel loss, introduced in Section~\ref{sec:loss}, is better-suited to action anticipation than existing ones. To evaluate this, we trained several versions of our model with different losses. In particular, as already done in the comparison to the state-of-the-art above, we replaced our loss with the ranking loss of~\citep{ma2016learning} (ranking loss on detection score) 
and the loss of~\citep{brain4Cars}, but this time within our complete multi-stage model, with both context- and action-aware features.

Furthermore, we made use of the standard cross-entropy ({\bf CE}) loss, which only accounts for one activity label for each sequence (at time $T$). This loss can be expressed as
\begin{align}
\mathcal{L}_{CE} = \sum^{N}_{k=1}[y^T(k)\log(\hat{y}^{T}(k))\ \nonumber \\+ (1-y^T(k))\log(1-\hat{y}^{T}(k))]\;.
\end{align}

We then also modified the loss of~\citep{brain4Cars}, which consists of an exponentially weighted softmax, with an exponentially weighted cross-entropy loss ({\bf ECE}),
written as
\begin{align}
\mathcal{L}_{ECE} = \sum^T_{t=1}-e^{-(T-t)}\sum^{N}_{k=1}[y^t(k)\log(\hat{y}^{t}(k)) \nonumber \\+ (1-y^t(k))\log(1-\hat{y}^{t}(k))]\;.
\end{align}

The main drawback of this loss comes from the fact that it does not strongly encourage the model to make correct predictions as early as possible. To address this issue, we also introduce a linearly growing loss  ({\bf LGL}), defined as
\begin{align}
\mathcal{L}_{LGL} = \sum^T_{t=1}{-\frac{t}{T}}\sum^{N}_{k=1}[y^t(k)\log(y^{t}(k)) \nonumber \\ + (1-y^t(k))\log(1-\hat{y}^{t}(k))].
\end{align}
While our new loss, introduced in Section~\ref{sec:loss}, also makes use of a linearly-increasing term, it corresponds to the false positives in our case, as opposed to the false negatives in the LGL. Since some actions are ambiguous in the first few frames, we find it more intuitive not to penalize false positives too strongly at the beginning of the sequence. This intuition is supported by our results below, which show that our loss yields better results than the LGL.

In Fig.~\ref{fig:anticipation}, we report the accuracy of the corresponding models as a function of the number of observed frames on the UCF-101 dataset. Note that our new loss yields much higher accuracies than the other ones, particularly when only a few frames of the sequence are observed; With only 2 frames observed, our loss yields an accuracy similar to the other losses with 30--40 frames. With 30fps, this essentially means that we can predict the action 1 second earlier than other methods. The importance of this result is exemplified by research showing that a large proportion of vehicle accidents are due to mistakes/misinterpretations of the scene in the immediate time leading up to the crash~\citep{brain4Cars,crash}. 


Moreover, in Fig.~\ref{fig:anticipation_pool}, we report the performance of the corresponding models as a function of the number of observed frames when using our average pooling strategy. Note that this strategy can generally be applied to any action anticipation method and, as shown by comparing Figs.~\ref{fig:anticipation} and~\ref{fig:anticipation_pool}, increases accuracy, especially at very early stages, which clearly demonstrates its effectiveness. Note that using it in conjunction with our loss still yields the best results by a significant margin.



\subsubsection{Influence of the Features}
We then evaluate the importance of the different feature types, context-aware and action-aware, on anticipation accuracy. To this end, we compare models trained using each feature type individually with our model that uses them jointly. For the models using a single feature type, we made use of a single LSTM to model temporal information. By contrast, our approach relies on a multi-stage LSTM, which we denote by \emph{MS-LSTM}. Note that all models were trained using our new anticipation loss. The results of this experiment on the UCF-101 dataset are provided in Table~\ref{tab:features}. These results clearly evidence the importance of using both feature types, which consistently outperforms individual ones.

\begin{table}[!h]
\renewcommand{\arraystretch}{1.2}
\small
\centering
\caption{Importance of the different feature types.}
\label{tab:features}
\begin{tabular}{l l c c c}
\hline
Feature & Model  & Earliest \scriptsize{K=1}& Latest \scriptsize{K=50}\\
\hline
Context-Aware 	& LSTM& 62.80\%& 72.71\%\\
Action-Aware  	& LSTM & 69.33\%& 77.86\% \\
Context+Action 	& MS-LSTM	& 80.5\%	& 83.37\%	\\
\hline
\end{tabular}
\end{table}

Since we extract action-aware features, and not motion-aware ones, our approach will not be affected by irrelevant motion. The CNN that extracts these features learns to focus on the discriminative parts of the images, thus discarding the irrelevant information. To confirm this, we conducted an experiment on some classes of UCF-101 that contain irrelevant motions/multiple actors, such as Baseball pitch, Basketball, Cricket Shot and Ice dancing. The results of our action-aware and context-aware frameworks for these classes are: 66.1\% vs. 58.2\% for Baseball pitch, 83\% vs. 76.4\% for Basketball, 65.1\% vs. 58\% for Cricket Shot, and 92.3\% vs. 91.7\% for Ice dancing. This shows that our action-aware features can effectively discard irrelevant motion/actors to focus on the relevant one(s).

\subsubsection{Visualization}
\begin{figure}[!h]
\centering
\includegraphics[width=\textwidth]{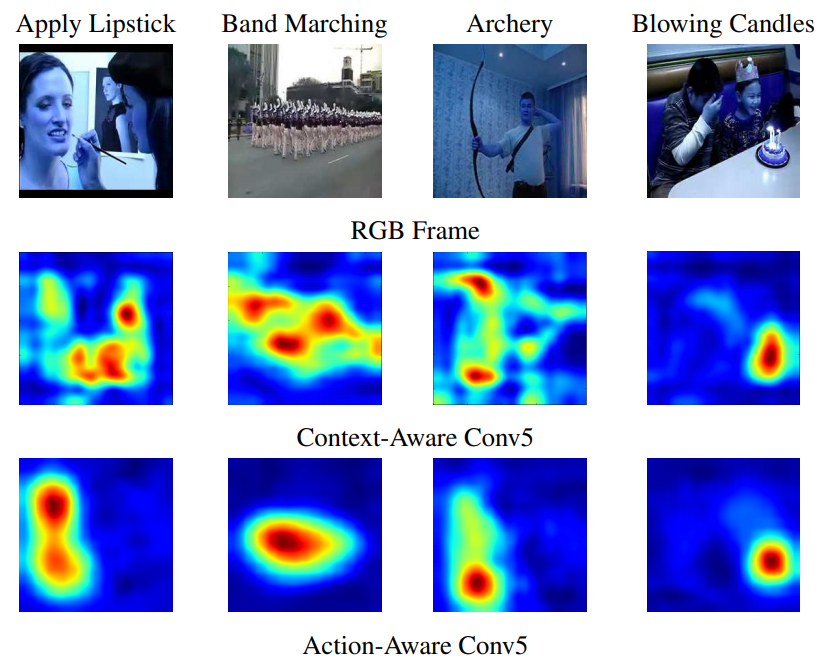}
\caption{Visualization of action-aware and context-aware features on UCF-101. These samples are representative of the data.}
\label{fig:visulaization}
\end{figure}

Finally, we provide a better intuition of the kind of information each of our feature types encode (see Fig.~\ref{fig:visulaization}). 
This visualization was computed by average pooling over the 512 channels of Conv5-3 (of both the context-aware and action-aware sub-networks). As can be observed in the figure, our context-aware features have high activations on regions corresponding to any relevant object in the scene (context). By contrast, in our action-aware features, high activations correctly correspond to the focus of the action. Therefore, they can reasonably localize the parts of the frame that most strongly participate in the action happening in the video and reduce the noise coming from context.

\section{Comparison to State-of-the-Art Action Recognition Methods}
We first compare the results of our approach to state-of-the-art methods on UCF-101, JHMDB-21 and UT-Interaction in terms of average accuracy over the standard training and testing partitions. In Table~\ref{tab:soa_recognition_ucf}, we provide the results on the UCF-101 dataset. Here, for the comparison to be fair, we only report the results of the baselines that do not use any other information than the RGB image and the activity label (we refer the readers to the baselines' papers and the survey~\citep{survey} for more detail). In other words, while it has been shown that additional, handcrafted features, such as dense trajectories and optical flow, can help improve accuracy~\citep{IDT,TrajectoryPooled,VLAD3,TwoStreamNIPS,DynamicNetwork}, our goal here is to truly evaluate the benefits of our method, not of these features. Note, however, that, as discussed in the next section of this supplementary material, our approach can still benefit from such features. As can be seen from the table, our approach outperforms all these RGB-based baselines. In Tables~\ref{tab:soa_recognition_jhmdb} and~\ref{tab:soa_recognition_uti}, we provide the results for JHMDB-21 and UT-Interaction. Again, we outperform all the baselines, even though, in this case, some of them rely on additional information such as optical flow~\citep{FindingActionTubes,actionness,soomro2016online,soomro2016predicting,ma2016learning} or IDT Fisher vector features~\citep{soomro2016online}.
We believe that these experiments show the effectiveness of our approach at tackling the action recognition problem.

\begin{table}[!h]
\centering
\caption{Comparison with state-of-the-art methods on UCF-101 (average accuracy over all training/testing splits). For the comparison to be fair, we focus on the baselines that, as us, only use the RGB frames as input.}
\label{tab:soa_recognition_ucf}
\begin{tabular}{l c}
\hline
Method & Accuracy\\
\hline
Dynamic Image Network~\citep{DynamicNetwork} & 70.0\% \\
Dynamic Image Network + Static RGB~\citep{DynamicNetwork} & 76.9\%\\
Rank Pooling~\citep{DiscriminativeRankPooling} & 72.2\%\\
DHR~\citep{DiscriminativeRankPooling} & 78.8\%\\
Zhang et al.~\citep{RealTimeAction} & 74.4\% \\
LSTM~\citep{LSTMAction} & 74.5\%\\
LRCN~\citep{LRCN} & 68.8\% \\
C3D~\citep{3DCNN} & 82.3\% \\
Spatial Stream Net~\citep{TwoStreamNIPS} & 73.0\% \\
Deep Network~\citep{LargeScaleCNN} & 65.4\% \\
ConvPool (Single frame)~\citep{BeyondAction} & 73.3\%  \\
ConvPool (30 frames)~\citep{BeyondAction} & 80.8\%\\
ConvPool (120 frames)~\citep{BeyondAction} & 82.6\% \\
\hline
Ours & {\bf 83.3\%}\\
\hline
Diff. to State-of-the-Art & +0.7\% \\
\hline
\end{tabular}
\end{table}

\begin{table}[!h]
\centering
\caption{Comparison with state-of-the-art methods on JHMDB-21 (average accuracy over all training/testing splits). Note that while the methods of~\citep{FindingActionTubes,actionness,soomro2016online,soomro2016predicting} use motion/optical flow information and~\citep{soomro2016online} uses IDT Fisher vector features, our method yields better performance.}
\label{tab:soa_recognition_jhmdb}
\begin{tabular}{l c}
\hline
Method & Accuracy \\
\hline
Where and What~\citep{soomro2016predicting} & 43.8\%\\
DP-SVM~\citep{soomro2016online} & 44.2\% \\
S-SVM~\citep{soomro2016online} & 47.3\%\\
Spatial-CNN~\citep{FindingActionTubes} & 37.9\% \\
Motion-CNN~\citep{FindingActionTubes} & 45.7\% \\
Full Method~\citep{FindingActionTubes} & 53.3\%\\
Actionness-Spatial~\citep{actionness} & 42.6\%\\
Actionness-Temporal~\citep{actionness} & 54.8\% \\
Actionness-Full Method~\citep{actionness} & 56.4\%\\
\hline
Ours& \bf{58.3\%} \\
\hline
Diff. to State-of-the-Art & +1.9\%\\
\hline
\end{tabular}
\end{table}

\begin{table}[!h]
\centering
\caption{Comparison with state-of-the-art methods on UT-Interaction (average accuracy over all training/testing splits). Note that while the methods of~\citep{soomro2016online} uses motion/optical flow information and IDT Fisher vector features, our method yields better performance.}
\label{tab:soa_recognition_uti}
\begin{tabular}{l c}
\hline
Method & Accuracy \\
\hline
D-BoW~\citep{ryoo2011human} & 85.0\%\\
I-BoW~\citep{ryoo2011human} & 81.7\%\\
Cuboid SVM~\citep{ryoo2010overview} & 85.0\%\\
BP-SVM~\citep{laviers2009improving} & 83.3\%\\
Cuboid/Bayesian~\citep{ryoo2011human} & 71.7\%\\
DP-SVM~\citep{soomro2016online} & 14.6\%\\
Yu et al.~\citep{yu2010real} & 83.3\%\\
Yuan et al.~\citep{yuan2010middle} & 78.2\%\\
Waltisberg et al.~\citep{waltisberg2010variations} & 88.0\%\\
\hline
Ours & \bf{90.0\%} \\
\hline
Diff. to State-of-the-Art & +2.0\%   \\
\hline
\end{tabular}
\end{table}

\section{Exploiting Optical Flow}
\label{sec:flow}
Note that our approach can also be extended into a two-stream architecture to benefit from optical flow information, as state-of-the-art action recognition methods do. In particular, to extract optical flow features, we made use of the pre-trained temporal network of~\citep{TwoStreamNIPS}. We then computed the CNN features from a stack of 20 optical flow frames (10 frames in the $x$-direction and 10 frames in the $y$-direction), from $t-10$ to $t$ at each time $t$. As these features are potentially loosely related to the action (by focusing on motion), we merge them with the input to the second stage of our multi-stage LSTM. In Table~\ref{tab:opticalFlow}, we compare the results of our modified approach with state-of-the-art methods that also exploit optical flow. Note that our two-stream approach yields accuracy comparable to the state-of-the-art.


\begin{table}[!h]
\renewcommand{\arraystretch}{1.2}
\centering
\small
\caption{Comparison with the state-of-the-art approaches that use optical flow. For the comparison to be fair, we focus on the baselines that, as us, use RGB frames+optical flow as input.}
\label{tab:opticalFlow}
\begin{tabular}{l c }
\hline
Method & Accuracy \\
\hline
Spatio-temporal ConvNet~\citep{LargeScaleCNN}			& 65.4\% \\
LRCN + Optical Flow~\citep{LRCN}  										& 82.9\% \\
LSTM + Optical Flow~\citep{LSTMAction}									& 84.3\% \\
Two-Stream Fusion~\citep{CNN2Stream} & 92.5\% \\
CNN features + Optical Flow~\citep{TwoStreamNIPS} 		& 73.9\% \\
ConvPool (30 frames) + OpticalFlow~\citep{BeyondAction} 	& 87.6\%\\
ConvPool (120 frames) + OpticalFlow~\citep{BeyondAction} & 88.2\%\\
VLAD3 + Optical Flow~\citep{VLAD3} 						& 84.1\% \\
Two-Stream ConvNet~\citep{TwoStreamNIPS} 					& 88.0\% \\
Two-Stream Conv.Pooling~\citep{BeyondAction}				& 88.2\% \\
Two-Stream TSN~\citep{TSN} 											& 91.5\% \\

\hline
Ours + Optical Flow 	& 91.8\% \\
\hline
\end{tabular}
\end{table}

We also conducted an experiment to evaluate the effectiveness of incorporating optical flow in our framework for action anticipation. To handle the case where less than 10 frames are used, we padded the frame stack with gray images (with values 127.5). Our flow-based approach achieved 86.8\% for earliest and 91.8\% for latest prediction on UCF-101, thus showing that, if runtime is not a concern, optical flow can indeed help increase the accuracy of our approach.

We further compare our approach with the two-stream network~\citep{TwoStreamNIPS}, designed for action recognition, applied to the task of action anticipation. On UCF-101, this model achieved 83.2\% for earliest and 88.6\% for latest prediction, which our approach with optical flow clearly outperforms.

\section{Effect of Different Feature Types}
Here, we evaluate the importance of the different feature types, context-aware and action-aware, on recognition accuracy. To this end, we compare models trained using each feature type individually with our model that uses them jointly. For all models, we made use of LSTMs with 2048 units. Recall that our approach relies on a multi-stage LSTM, which we denote by \emph{MS-LSTM}. The results of this experiment for different losses are reported in Table~\ref{tab:features}. These results clearly evidence the importance of using both feature types, which consistently outperforms using individual ones in all settings.

\begin{table}[!h]
\renewcommand{\arraystretch}{1.2}
\centering
\caption{Importance of the different feature types using different losses. Note that combining both types of features consistently outperforms using a single one. Note also that, for a given model, our new  loss yields higher accuracies than the other ones.}
\label{tab:features}
\begin{tabular}{l l c}
\hline
Feature & Sequence Learning & Accuracy \\
\hline
Context-Aware	& LSTM (CE)	& 72.38\%\\
Action-Aware 	& LSTM (CE) & 74.24\% \\
Context+Action 	& MS-LSTM (CE)	& 78.93\%\\
\hline
Context-Aware	& LSTM (ECE)& 72.41\% \\
Action-Aware 	& LSTM (ECE) & 77.20\% \\
Context+Action 	& MS-LSTM (ECE)	& 80.38\%\\
\hline
Context-Aware	& LSTM (LGL) & 72.58\%\\
Action-Aware 	& LSTM (LGL) & 77.63\% \\
Context+Action 	& MS-LSTM (LGL) & 81.27\%\\
\hline
Context-Aware	& LSTM (Ours)& 72.71\%\\
Action-Aware 	& LSTM (Ours) & 77.86\% \\
Context+Action 	& MS-LSTM (Ours)	& 83.37\%	\\
\hline
\end{tabular}
\end{table}

\section{Robustness to the Number of Hidden Units}

Based on our experiments, we found that for large datasets such as UCF-101, the 512 hidden units that some baselines use (e.g.~\citep{LRCN,LSTMAction}) do not suffice to capture the complexity of the data. Therefore, to study the influence of the number of units in the LSTM, we evaluated different versions of our model with 1024 and 2048 hidden units (since 512 yields poor results and higher numbers, e.g., 4096, would require too much memory) and trained the model with 80\% training data and validated on the remaining 20\%. 
For a single LSTM, we found that using 2048 hidden units performs best. For our multi-stage LSTM, using 2048 hidden units also yields the best results. We also evaluated the importance of relying on average pooling in the LSTM. The results of these different versions of our MS-LSTM framework are provided in Table~\ref{tab:AvgPool}. This shows that, typically, more hidden units and average pooling can improve accuracy slightly.

\begin{table}[!h]
\renewcommand{\arraystretch}{1.2}
\centering
\small
\caption{Influence of the number of hidden LSTM units and of our average pooling strategy in our multi-stage LSTM model. These experiments were conducted on the first splits of UCF-101 and JHMDB-21.}
\label{tab:AvgPool}
\begin{tabular}{l  c c c c}
\hline
  & Average  & Hidden & & \\
Setup  & Pooling & Units & UCF-101 & JHMDB-21\\
\hline
Ours (CE)& wo/ & 1024 & 77.26\%	& 52.80\% \\
Ours (CE)& wo/ & 2048 & 78.09\%	& 53.43\% \\
Ours (CE)& w/ & 2048 &	78.93\% & 54.30\%\\
\\
Ours (ECE)& wo/ & 1024 & 79.10\%	& 55.33\% \\
Ours (ECE)& wo/ & 2048 & 79.41\%	& 56.12\% \\
Ours (ECE)& w/ & 2048 & 80.38\%	& 57.05\%\\
\\
Ours (LGL)& wo/ & 1024 & 79.76\%	& 55.70\% \\
Ours (LGL)& wo/ & 2048 & 80.10\%	& 56.83\% \\
Ours (LGL)& w/ & 2048 & 81.27\%	& 57.70\%\\
\\
Ours & wo/ & 1024 & 81.94\%	& 56.24\% \\
Ours & wo/ & 2048 & 82.16\%	& 57.92\%\\
Ours & w/ & 2048 & 83.37\%	& 58.41\%\\
\hline
\end{tabular}
\end{table}

\section*{Effect of the LSTM Architecture}
Finally, we study the effectiveness of our multi-stage LSTM architecture at merging our two feature types. To this end, we compare the results of our MS-LSTM with the following baselines: A single-stage LSTM that takes as input the concatenation of our context-aware and action-aware features (Concatenation); The use of two parallel LSTMs whose outputs are merged by concatenation and then fed to a fully-connected layer (Parallel). A multi-stage LSTM where the two different feature-types are processed in the reverse order (Swapped), that is, the model processes the action-aware features first and, in a second stage, combines them with the context-aware ones;
The results of this comparison are provided in Table~\ref{tab:LSTMArch}. Note that both multi-stage LSTMs outperform the single-stage one and the two parallel LSTMs, thus indicating the importance of treating the two types of features sequentially. Interestingly, processing context-aware features first, as we propose, yields higher accuracy than considering the action-aware ones at the beginning. This matches our intuition that context-aware features carry global information about the image and will thus yield noisy results, which can then be refined by exploiting the action-aware features.

\begin{table}[!h]
\renewcommand{\arraystretch}{1.2}
\centering
\caption{Comparison of our multi-stage LSTM model with diverse fusion strategies. We report the results of simple concatenation of the context-aware and action-aware features, their use in two parallel LSTMs with late fusion, and swapping their order in our multi-stage LSTM, i.e., action-aware first, followed by context-aware. Note that multi-stage architectures yield better results, with the best ones achieved by using context first, followed by action, as proposed in this chapter.}
\label{tab:LSTMArch}
\begin{tabular}{l l c}
\hline
Feature & Sequence & \\
Order & Learning & Accuracy \\
\hline
Concatenation 	& LSTM  				& 77.16\% \\
Parallel 		& 2 Parallel LSTMs  	& 78.63\% \\
Swapped 		& MS-LSTM (Ours)  				& 78.80\% \\
Ours 			& MS-LSTM (Ours) 			& 83.37\% \\
\hline
\end{tabular}
\end{table}

Furthermore, we evaluate a CNN-only version of our approach, where we removed the LSTM, but kept our average pooling strategy to show the effect of our MS-LSTM architecture on top of the CNN. On UCF-101, this achieved 69.53\% for earliest and 73.80\% for latest prediction. This shows that, while this CNN-only framework yields reasonable predictions, our complete approach with our multistage LSTM benefits from explicitly being trained on multiple frames, thus achieving significantly higher accuracy (80.5\% and 83.4\%, respectively). While the LSTM could in principle learn to perform average pooling, we believe that the lack of data prevents this from happening.

\section{Conclusion}
In this chapter, we have introduced a novel loss function to address very early action anticipation. Our loss encourages the model to make correct predictions as early as possible in the input sequence, thus making it particularly well-suited to action anticipation. Furthermore, we have introduced a new multi-stage LSTM model that effectively combines context-aware and action-aware features. Our experiments have evidenced the benefits of our new loss function over existing ones. Furthermore, they have shown the importance of exploiting both context- and action-aware information. Altogether, our approach significantly outperforms the state-of-the-art in action anticipation on all the datasets we applied it to. However, all of these datasets are mainly designed for action recognition task, predicting the action label given the full extent of the video. We argue that this is not very well-suited for the task of anticipation since most actions start from the beginning of the videos. A proper dataset for action anticipation should contain videos that actions start after a reasonable amount of observation, e.g., after 25\% of the video length. In the next chapter, we focus on creating such dataset for the very challenging task of action anticipation in driving scenarios. We also extend the multi-stage LSTM from processing two modalities (action-aware and context-aware) to arbitrary number of modalities.

%% file: chapter2.tex
\chapter{Action Anticipation in Driving Scenarios}
\label{cha:intro}

Following previous chapter, we continue focusing on a discrete and determinsitic anticipation task. Motivated by the success of our approach discussed in previous chapter, we move towards a more challenging and crucial application, action anticipation in driving scenarios. In such scenarios, anticipation becomes critical since one, the driver of ego car, other car's driver, pedestrian, or cyclist, needs to react before the action is finalized, for instance, where a car needs to, e.g., avoid hitting pedestrians and respect traffic lights. While solutions have been proposed to tackle subsets of the driving anticipation tasks, by making use of diverse, task-specific sensors, there is no single dataset or framework that addresses them all in a consistent manner. In this chapter, we therefore introduce a new, large-scale dataset, called VIENA$^2$, covering 5 generic driving scenarios, with a total of 25 distinct action classes. It contains more than 15K full HD, 5s long videos acquired in various driving conditions, weathers, daytimes and environments, complemented with a common and realistic set of sensor measurements. This amounts to more than 2.25M frames, each annotated with an action label, corresponding to 600 samples per action class. We discuss our data acquisition strategy and the statistics of our dataset, and benchmark state-of-the-art action anticipation techniques, including a new multi-modal LSTM architecture with an effective loss function for action anticipation in driving scenarios.

\section{Introduction}
\label{sec:introduction}
Understanding actions/events from videos is key to the success of many real-world applications, such as autonomous navigation, surveillance and sports analysis. While great progress has been made to recognize actions from complete sequences~\citep{feichtenhofer2016convolutional,LRCN,TSN,DynamicNetwork},
action anticipation, which aims to predict the observed action as early as possible, has only reached a much lesser degree of maturity
~\citep{sadegh2017encouraging,vondrick2016anticipating,soomro2016predicting}. 
Nevertheless, anticipation is a crucial component in scenarios where a system needs to react quickly, such as in robotics~\citep{koppula2016anticipating}, and automated driving
~\citep{jain2016brain4cars,liebner2013generic,li2017unified}. 
Its benefits have also been demonstrated in surveillance settings~\citep{ramanathan2016detecting,wang2017hierarchical}.

\begin{figure}[t]
\centering
\includegraphics[width=\textwidth]{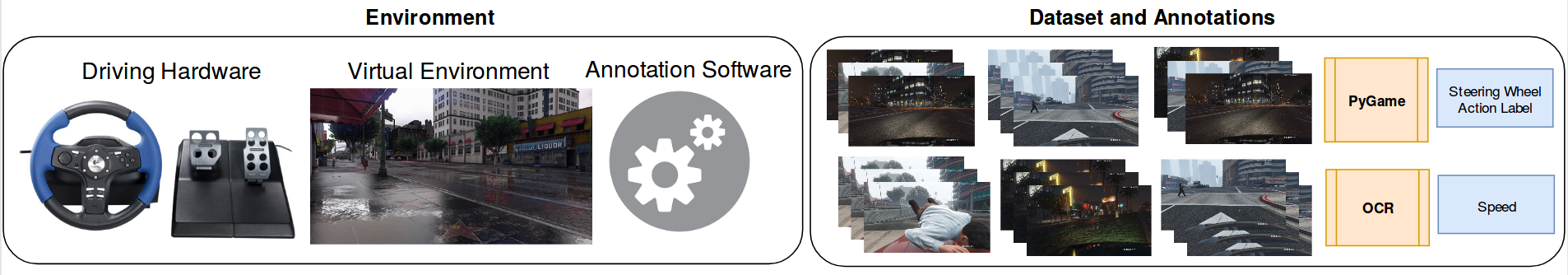}
\caption{{\bf Overview of our data collection.} Using the GTA V environment and driving equipment depicted in the left box, we captured a new dataset covering 5 generic scenarios, illustrated in the right box, each containing multiple action classes.}
\label{fig:Viena2Overal}
\end{figure}

In this chapter, we focus on the driving scenario. In this context, when consulting the main actors in the field, may they be from the computer vision community, the intelligent vehicle one or the automotive industry, the consensus is that predicting the intentions of a car's own driver, for Advanced Driver Assistance Systems (ADAS), remains a challenging task for a computer, despite being relatively easy for a human~\citep{dong2017intention,olabiyi2017driver,jain2016recurrent,jain2016brain4cars,rasouli2017agreeing}.
Anticipation then becomes even more complex when one considers the maneuvers of other vehicles and pedestrians~\citep{klingelschmitt2016probabilistic,zyner2017long,dong2017intention}. However, it is key to avoiding dangerous situations, and thus to the success of autonomous driving.

Over the years, the researchers in the field of anticipation for driving scenarios have focused on specific subproblems of this challenging task, such as lane change detection~\citep{morris2011lane,tawari2014looking}, a car's own driver's intention~\citep{ohn2014head} or maneuver~\citep{jain2015car,jain2016recurrent,jain2016brain4cars,olabiyi2017driver} recognition and pedestrian intention prediction~\citep{rasouli2017agreeing,pool2017using,li2017unified,schulz2015controlled}. 
Furthermore, these different subproblems are typically addressed by making use of different kinds of sensors, without considering the fact that, in practice, the automotive industry might not be able/willing to incorporate all these different sensors to address all these different tasks.

In this chapter, we study the general problem of anticipation in driving scenarios, 
encompassing all the subproblems discussed above, and others, such as other drivers' intention prediction, with a fixed, sensible set of sensors. To this end, we introduce the \textbf{VI}rtual \textbf{EN}vironment for \textbf{A}ction \textbf{A}nalysis (VIENA$^2$) dataset, covering the five different subproblems of predicting driver maneuvers, pedestrian intentions, front car intentions, traffic rule violations, and accidents.
Altogether, these subproblems encompass a total of 25 distinct action classes. 
VIENA$^2$ was acquired using the GTA V video game~\citep{GTAgame}. It contains more than 15K full HD, 5s long videos, corresponding to more than 600 samples per action class, acquired in various driving conditions, weathers, daytimes, and environments. This amounts to more than 2.25M frames, each annotated with an action label. These videos are complemented by basic vehicle dynamics measurements, and therefore reflect well the type of information that one could have access to in practice.

Below, we describe how VIENA$^2$ was collected and compare its statistics and properties to existing datasets. We then benchmark state-of-the-art action anticipation algorithms on VIENA$^2$, and introduce a new multi-modal, LSTM-based architecture, together with a new anticipation loss, which outperforms existing approaches in our driving anticipation scenarios. Finally, we investigate the benefits of our synthetic data to address anticipation from real images.
In short, our contributions are: {\bf (i)} a large-scale action anticipation dataset for general driving scenarios; {\bf(ii)} a multi-modal action anticipation architecture. 

VIENA$^2$ is meant as an extensible dataset that will grow over time to include not only more data but also additional scenarios. Note that, for benchmarking purposes, however, we will clearly define training/test partitions. A similar strategy was followed by other datasets such as CityScapes, which contains a standard benchmark set but also a large amount of additional data. VIENA$^2$ is publicly available, together with our benchmark evaluation, our new architecture and our multi-domain training strategy.

\begin{figure}[t]
\centering
\includegraphics[width=\textwidth]{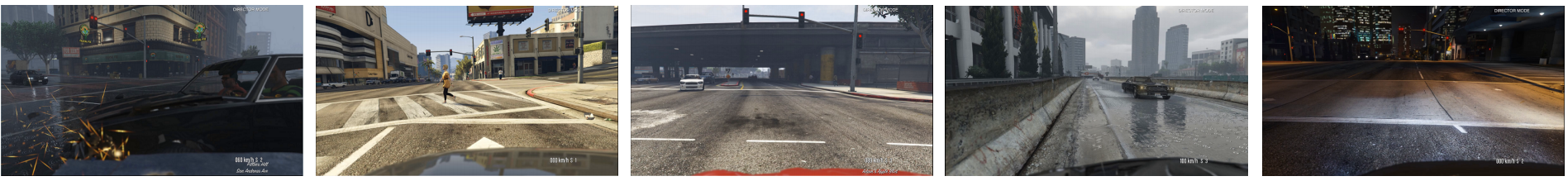} 

\caption{Samples from different scenarios of VIENA$^2$.} 
\label{fig:dataset}
\end{figure}

\section{VIENA$^2$}
\label{sec:viena2}
VIENA$^2$ is a large-scale dataset for action anticipation, and more generally action analysis, in driving scenarios.
While it is generally acknowledged that anticipation is key to the success of automated driving, 
to the best of our knowledge, there is currently no dataset that covers a wide range of scenarios with a common, yet sensible set of sensors. Existing datasets focus on specific subproblems, such as driver maneuvers
and pedestrian intentions~\citep{rasouli2017agreeing,pool2017using,kooij2014context}, and make use of different kinds of sensors. Furthermore, with the exception of~\citep{jain2016brain4cars}, none of these datasets provide videos whose first few frames do not already show the action itself or the preparation of the action. 

To create VIENA$^2$, we made use of the GTA V video game, whose publisher allows, under some conditions, for the non-commercial use of the footage~\citep{GTA_file}. Beyond the fact that, as shown in~\citep{richterplaying} via psychophysics experiments, GTA V provides realistic images that can be captured in varying weather and daytime conditions, it has the additional benefit of allowing us to cover crucial anticipation scenarios, such as accidents, for which real-world data would be virtually impossible to collect. 

In this section, we first introduce the different scenarios covered by VIENA$^2$ and discuss the data collection process. We then study the statistics of VIENA$^2$ and compare it against existing datasets.

\subsection{Scenarios and Data Collection}
\label{sec:scenarios}

As illustrated in Fig.~\ref{fig:dataset}, VIENA$^2$ covers five generic driving scenarios.
These scenarios are all human-centric, i.e., consider the intentions of humans, but three of them focus on the car's own driver, while the other two relate to the environment (i.e., pedestrians and other cars). These scenarios are:
\begin{enumerate}
\item \textbf{Driver Maneuvers (DM).} This scenario covers the 6 most common maneuvers a driver performs while driving: Moving forward (FF), stopping (SS), turning (left (LL) and right (RR)) and changing lane (left (CL) and right (CR)). Anticipation of such maneuvers as early as possible is critical in an ADAS context to avoid dangerous situations.

\item \textbf{Traffic Rules (TR).} This scenario contains sequences depicting the car's own driver either violating or respecting traffic rules, e.g., stopping at (SR) and passing (PR) a red light, driving in the (in)correct direction (WD,CD), and driving off-road (DO). Forecasting these actions is also crucial for ADAS.

\item \textbf{Accidents (AC).} In this scenario, we capture the most common real-world accident cases~\citep{volvoaccident}: Accidents with other cars (AC), with pedestrians (AP), and with assets (AA), such as buildings, traffic signs, light poles and benches, as well as no accident (NA).
Acquiring such data in the real world is virtually infeasible. Nevertheless, these actions are crucial to anticipate for ADAS and autonomous driving.

\item \textbf{Pedestrian Intentions (PI).} This scenario addresses the question of whether a pedestrian is going to cross the road (CR), or has stopped (SS) but does not want to cross, or is walking along the road (AS) (on the sidewalk). We also consider the case where no pedestrian is in the scene (NP). As acknowledged in the literature~\citep{pool2017using,schulz2015controlled,rasouli2017agreeing}, early understanding of pedestrians' intentions is critical for automated driving.

\item \textbf{Front Car Intentions (FCI).} The last generic scenario of VIENA$^2$ aims at anticipating the maneuvers of the front car. This knowledge has a strong influence on the behavior to adopt to guarantee safety. The classes are same as the ones in Driver Maneuver scenario, but for the driver of the front car.
\end{enumerate}

We also consider an additional scenario consisting of the same driver maneuvers as above but for heavy vehicles, i.e., trucks and buses. In all these scenarios, for the data to resemble a real driving experience, we made use of the equipment depicted in Fig.~\ref{fig:Viena2Overal}, consisting of a steering wheel with a set of buttons and a gear stick, as well as of a set of pedals. 
We then captured images at 30 fps with a single virtual camera mounted on the vehicle and facing the road forward. Since the speed of the vehicle is displayed at a specific location in these images, we also extracted it using an OCR module~\citep{smith2007overview} (see supplementary material for more detail on data collection). 
Furthermore, we developed an application that records measurements from the steering wheel. In particular, it gives us access to the steering angle every 1 microsecond, which allowed us to obtain a value of the angle synchronized with each image. Our application also lets us obtain the ground-truth label of each video sequence by recording the driver input from the steering wheel buttons. This greatly facilitated our labeling task, compared to~\citep{richterplaying,richter2016playing}, which had to use a middleware to access the rendering commands from which the ground-truth labels could be extracted. Ultimately, VIENA$^2$ consists of video sequences with synchronized measurements of steering angles and speed, and corresponding action labels.

\begin{figure*}[t]
\centering
\includegraphics[width=\textwidth]{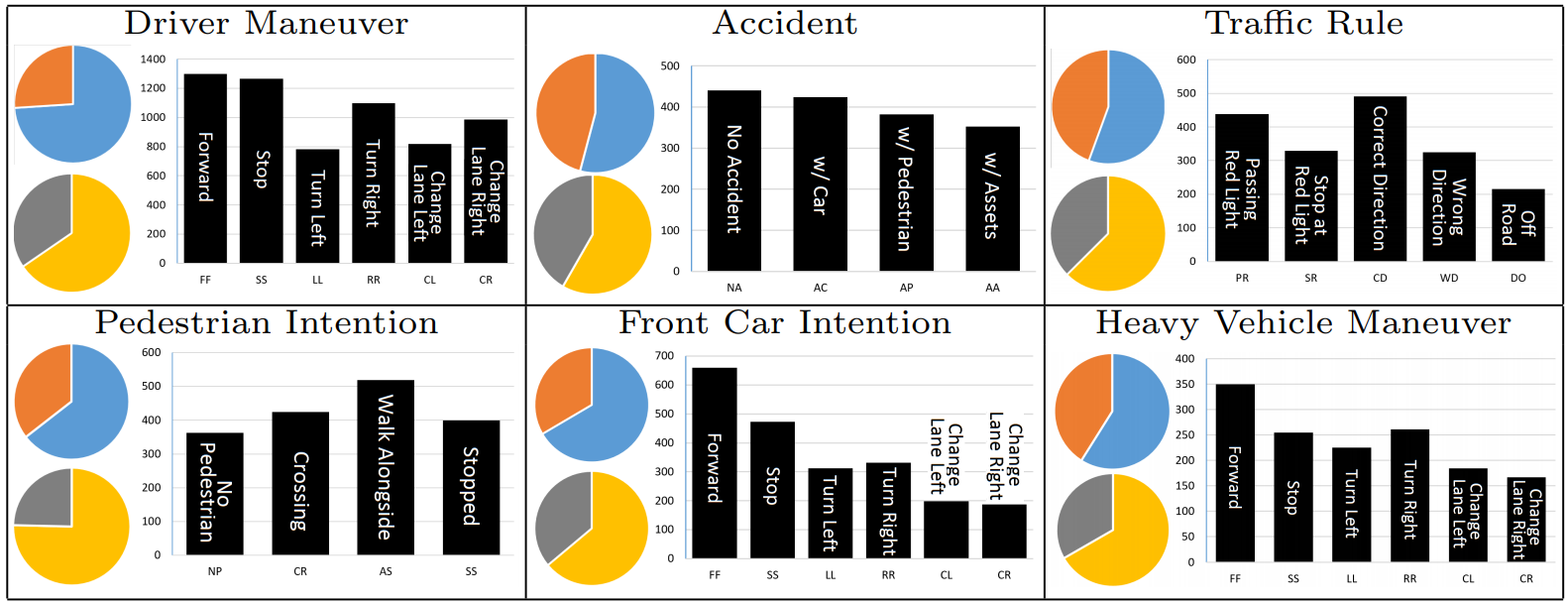}
\caption{{\bf Statistics for each scenario of VIENA$^2$.} We plot the number of videos per class, and proportions of different weather conditions (clear in yellow vs rainy/snowy in gray) and different daytime (day in orange vs night in blue). Best seen in color.}
\label{fig:plot_stat}
\end{figure*}

Altogether, VIENA$^2$ contains more than 15K full HD videos (with frame size of $1920\times 1280$), corresponding to a total of more than 2.25M annotated frames. The detailed number of videos for each class and the proportions of different weather and daytime conditions of VIENA$^2$ are provided in Fig.~\ref{fig:plot_stat}. Each video contains 150 frames captured at 30 frames-per-second depicting a single action from one scenario. The action occurs in the second half of the video (mostly around the $4$ second mark), which makes VIENA$^2$ well-suited to research on action anticipation, where one typically needs to see what happens before the action starts. 

Our goal is for VIENA$^2$ to be an extensible dataset. Therefore, by making our source code and toolbox for data collection and annotation publicly available, we aim to encourage the community to participate and grow VIENA$^2$.
Furthermore, while VIENA$^2$ was mainly collected for the task of action anticipation in driving scenarios, as it contains full length videos, i.e., videos of a single drive of 30 minutes on average depicting multiple actions, it can also be used for the tasks of action recognition and temporal action localization. 

\subsection{Comparison to Other Datasets}
\label{sec:compare_datasets}

The different scenarios and action classes of VIENA$^2$ make it compatible with existing datasets, thus potentially allowing one to use our synthetic data in conjunction with real images.
For instance, the action labels in the Driver Maneuver scenario correspond to the ones in Brain4Cars~\citep{jain2016brain4cars} and in the Toyota Action Dataset~\citep{olabiyi2017driver}. Similarly, our last two scenarios dealing with heavy vehicles contain the same labels as in Brain4Cars~\citep{jain2016brain4cars}. Moreover, the actions in the Pedestrian Intention scenario corresponds to those in~\citep{ped_benchmark}.
Note, however, that, to the best of our knowledge, there is no other dataset covering our Traffic Rules and Front Car Intention scenarios, or containing data involving heavy vehicles. Similarly, there is no dataset that covers accidents involving a driver's own car. In this respect, the most closely related dataset is DashCam~\citep{chan2016anticipating}, which depicts accidents of other cars. Furthermore, VIENA$^2$ covers a much larger diversity of environmental conditions, such as daytime variations (morning, noon, afternoon, night, midnight), weather variations (clear, sunny, cloudy, foggy, hazy, rainy, snowy), and location variations (city, suburbs, highways, industrial, woods), than existing public datasets. In the supplementary material, we provide examples of each of these different environmental conditions.

In addition to covering more scenarios and conditions than other driving anticipation datasets, VIENA$^2$ also contains more samples per class than existing action analysis datasets, both for recognition and anticipation. As shown in Table~\ref{tbl:dataset_comparison}, with 600 samples per class, VIENA$^2$ outsizes (at least class-wise) the datasets that are considered \emph{large} by the community. This is also the case for other synthetic datasets, such as VIPER~\citep{richterplaying}, GTA5~\citep{richter2016playing}, Virtual KITTI~\citep{gaidon2016virtual}, and SYNTHIA~\citep{ros2017semantic}, which, by targeting different problems, such as semantic segmentation for which annotations are more costly to obtain, remain limited in size. We acknowledge, however, that, since we target driving scenarios, our dataset cannot match in absolute size more general recognition datasets, such as Kinetics.


\begin{table}[t]
\caption{Statistics comparison with action recognition and anticipation datasets. A * indicates a dataset specialized to one scenario, e.g., driving, as opposed to generic.}
\label{tbl:dataset_comparison}
    \centering
    \begin{tabular}{l c c c}
    \toprule
    Recognition Dataset & Samples/Class & classes & videos \\
    \midrule
    UCF-101~\citep{soomro2012ucf101} & 150 & 101 & 13.3K \\
    HMDB/JHMDB~\citep{Kuehne11} & 120 & 51/21 & 5.1K/928 \\
    UCF-Sport*~\citep{ucfsport} & 30 & 10 & 150 \\
    Charades~\citep{sigurdsson2016hollywood} & 100 & 157 & 9.8K \\
    ActivityNe~\citep{caba2015activitynet} & 144 & 200 & 15K \\
    Kinetics~\citep{kay2017kinetics} & 400 & 400 & 306K \\

    \\
    \toprule
    Anticipation Dataset  & Samples/Class & classes & videos \\
    \midrule
    UT-Interaction~\citep{ryoo2009spatio} & 20 & 6 & 60\\
    Brain4Cars*~\citep{jain2016brain4cars} & 140 & 6 & 700 \\
    JAAD*~\citep{rasouli2017agreeing}  & 86 & 4 & 346\\
    \midrule
    VIENA$^2$* & 600 & 25 & 15K \\
    \bottomrule
    \end{tabular}
\end{table}

\section{Benchmark Algorithms}
\label{sec:benchmark}
In this section, we first discuss the state-of-the-art action analysis and anticipation methods that we used to benchmark our dataset. We then introduce a new multi-modal LSTM-based approach to action anticipation, and finally discuss how we model actions from our images and additional sensors.

\subsection{Baseline Methods}
The idea of anticipation was introduced in the computer vision community almost a decade ago by~\citep{ryoo2009spatio}. While the early methods~\citep{ryoo2011human,soomro2016predicting,soomro2016online}
relied on handcrafted-features, they have now been superseded by end-to-end learning methods~\citep{ma2016learning,jain2016brain4cars,sadegh2017encouraging}, focusing on designing new losses better-suited to anticipation. In particular, the loss of our approach in Chapter~\ref{cha:encouraging_lstms} has proven highly effective, achieving state-of-the-art results on several standard benchmarks. 

Despite the growing interest of the community in anticipation, action recognition still remains more thoroughly investigated. Since recognition algorithms can be converted to performing anticipation by making them predict a class label at every frame, we include the state-of-the-art recognition methods in our benchmark.
Specifically, we evaluate the following baselines:
\paragraph{Baseline 1: CNN+LSTMs.} The high performance of CNNs in image classification makes them a natural choice for video analysis, via some modifications. This was achieved in~\citep{LRCN} by feeding the frame-wise features of a CNN to an LSTM model, and taking the output of the last time-step LSTM cell as  prediction. For anticipation, we can then simply consider the prediction at each frame. We then use the temporal average pooling strategy introduced in Chapter~\ref{cha:encouraging_lstms}, which has proven effective to increase the robustness of the predictor for action anticipation.
\paragraph{Baseline 2: Two-Stream Networks.} 
Baseline 1 only relies on appearance, ignoring motion inherent to video (by motion, we mean explicit motion information as input, such as optical flow). Two-stream architectures, such as the one of~\citep{feichtenhofer2016convolutional}, have achieved state-of-the-art performance by explicitly accounting for motion. In particular, this is achieved by  taking a stack of 10 externally computed optical flow frames as input to the second stream. A prediction for each frame can be obtained by considering the 10 previous frames in the sequence for optical flow. We also make use of temporal average pooling of the predictions.
\paragraph{Baseline 3: Multi-Stage LSTMs.} The Multi-Stage LSTM (MS-LSTM) discussed in Chapter 3 constitutes the state of the art in action anticipation. This model jointly exploits context- and action-aware features that are used in two successive LSTM stages.
As mentioned above, the key to the success of MS-LSTM is its training loss function. This loss function can be expressed as

\begin{align}
\mathcal{L}(y, \hat{y}) = -\frac{1}{N}\sum^N_{k=1}\sum^T_{t=1}\Bigg[y^t(k) \log(\hat{y}^t(k)) + w(t)(1-y^t(k))\log(1-\hat{y}^t(k))\Bigg]\;,
\label{eq:loss}
\end{align}

where $y^t(k)$ is the ground-truth label of sample $k$ at frame $t$, $\hat{y}^t(k)$ the corresponding prediction, and $w(t) = \frac{t}{T}$. The first term encourages the model to predict the correct action at any time, while the second term accounts for ambiguities between different classes in the earlier part of the video.



\subsection{A New Multi-Modal LSTM}
While effective, MS-LSTM suffers from the fact that it was specifically designed to take two modalities as input, the order of which needs to be manually defined. 
As such, it does not naturally apply to our more general scenario, and must be actively modified, in what might be a sub-optimal manner, to evaluate it with our action descriptors. To overcome this, we therefore introduce a new multi-modal LSTM (MM-LSTM) architecture that generalizes the multi-stage architecture, introduced in Chapter 3 of this thesis, to an arbitrary number of modalities. Furthermore, our MM-LSTM also aims to learn the importance of each modality for the prediction.

\begin{figure}[t]
\centering
\begin{tabular} {cc}
\includegraphics[width=.6\textwidth]{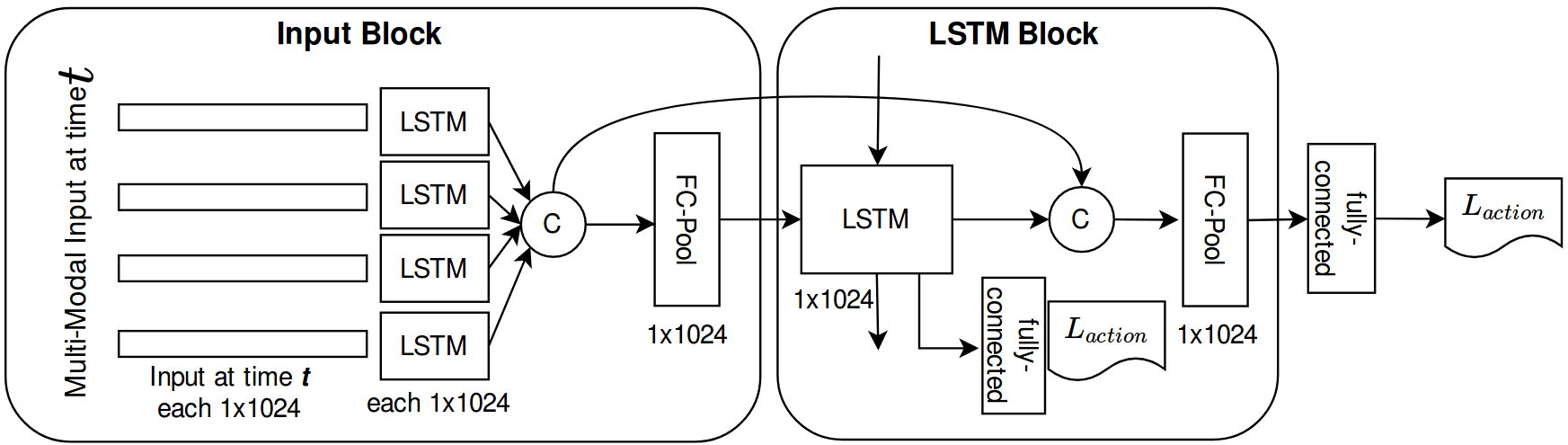} & 
\includegraphics[width=.3\textwidth]{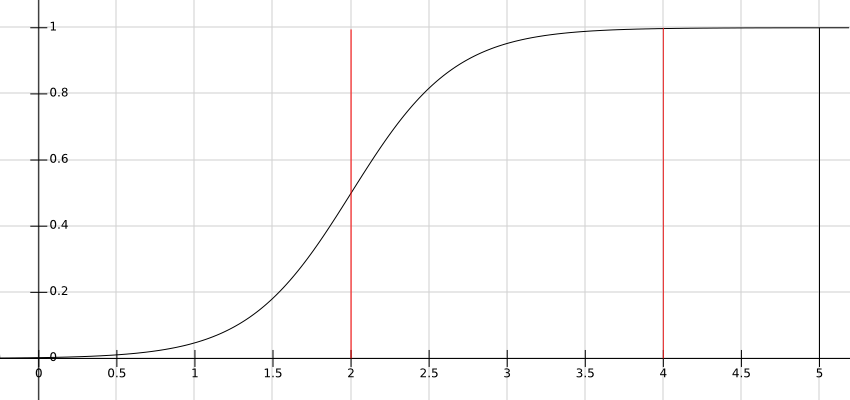} \\
Our MM-LSTM architecture & $w(t) = \frac{e^{(\alpha t-\beta)}}{1 + e^{(\alpha t-\beta)}}$\\
\end{tabular}
\caption{(Left) Our Multi-Stage LSTM architecture. (Right) Visualization of our weighting function for the anticipation loss of Eq.~\ref{eq:loss}.}
\label{fig:MM_LSTM}
\end{figure}

Specifically, as illustrated in Fig.~\ref{fig:MM_LSTM} for $M=4$ modalities, at each time $t$, the representations of the $M$ input modalities are first passed individually into an LSTM with a single hidden layer. The activations of these $M$ hidden layers are then concatenated into an $M \times 1024$ matrix $D^t$, which acts as input to a time-distributed fully-connected layer (FC-Pool). This layer then combines the $M$ modalities to form a single vector $O^t \in \mathbb{R}^{1024}$. This representation is then passed through another LSTM whose output is concatenated with the original $D^t$ via a skip connection. The resulting $(M+1) \times 1024$ matrix is then compacted into a 1024D vector via another FC-Pool layer. The output of this FC-Pool layer constitutes the final representation and acts as input to the classification layer.

The reasoning behind this architecture is the following. The first FC-Pool layer can learn the importance of each modality. While its parameters are shared across time, the individual, modality-specific LSTMs can produce time-varying outputs, thus, together with the FC-Pool layer, providing the model with the flexibility to change the importance of each modality over time. In essence, this allows the model to learn the importance of the modalities dynamically. The second LSTM layer then models the temporal variations of the combined modalities. The skip connection and the second FC-Pool layer produce a final representation that can leverage both the individual, modality-specific representations and the learned combination of these features.

\paragraph{Learning.} To train our model, we make use of the loss of Eq.~\ref{eq:loss}. However, we modify the weights as $w(t) = \frac{e^{(\alpha t-\beta)}}{1 + e^{(\alpha t-\beta)}}$, allowing the influence of the second term to vary nonlinearly. In practice, we set $\alpha=3$ and $\beta=6$, yielding the weight function of Fig.~\ref{fig:MM_LSTM}. These values were motivated by the study of~\citep{pentland1999modeling}, which shows that driving actions typically undergo the following progression: In a first stage, the driver is not aware of an action or decides to take an action. In the next stage, the driver becomes aware of an action or decides to take one. This portion of the video contains crucial information for anticipating the upcoming action. In the last portion of the video, the action has started. In this portion of the video, we do not want to make a wrong prediction, thus penalizing false positives strongly. Generally speaking, our sigmoid-based strategy to define the weight reflects the fact that, in practice and in contrast with many academic datasets, such as UCF-101~\citep{soomro2012ucf101} and JHMDB-21~\citep{JhuangICCV2013}, actions do not start right at the beginning of a video sequence, but at any point in time, the goal being detecting them as early as possible.

During training, we rely on stage-wise supervision, by introducing an additional classification layer after the second LSTM block, as illustrated in Fig.~\ref{fig:MM_LSTM}. At test time, however, we remove this intermediate classifier to only keep the final one. We then make use of the temporal average pooling strategy introduced in Chapter~\ref{cha:encouraging_lstms} to accumulate the predictions over time.

\subsection{Action Modeling}
\label{sec:action_modeling}

Our MM-LSTM can take as input multiple modalities that provide diverse and complementary information about the observed data. Here, we briefly describe the different descriptors that we use in practice.

\begin{itemize}
\item {\bf Appearance-based Descriptors.} Given a frame at time $t$, the most natural source of information to predict the action is the appearance depicted in the image. To encode this information, 
we make use of a slightly modified DenseNet~\citep{huang2016densely}, pre-trained on ImageNet. See Section~\ref{sec:implementation} for more detail.
Note that we also use this DenseNet as appearance-based CNN for Baselines 1 and 2.
\item {\bf Motion-based Descriptors.} Motion has proven a useful cue for action recognition~\citep{feichtenhofer2017spatiotemporal,feichtenhofer2016convolutional}. To encode this, we make use of a similar architecture as for our appearance-based descriptors, but modify it to take as input a stack of optical flows. Specifically, we extract optical flow between $L$ consecutive pairs of frames, in the range $[t-L, t]$, and form a $2L$ flow stack encoding horizontal and vertical flows. We fine-tune the model pre-trained on ImageNet for the task of action recognition, and take the output of the additional fully-connected layer as our motion-aware descriptor. Note that we also use this DenseNet for the motion-based stream of Baseline 2. 
\item {\bf Vehicle Dynamics.} In our driving context, we have access to additional vehicle dynamics measurements. For each such measurement, at each time $t$, we compute a vector from its value $s_t$, its velocity $(s_t - s_{t-\delta})$ and its acceleration $(s_t - 2s_{t-\delta} + s_{t-2\delta})$. To map these vectors to a descriptor of size comparable to the appearance- and motion-based ones, inspired by~\citep{fernando2017going}, we train an LSTM with a single hidden layer modeling the correspondence between vehicle dynamics and action label. In our dataset, we have two types of dynamics measurements, steering angle and speed, which results in two additional descriptors. 

\end{itemize}

When evaluating the baselines, we report results of both their standard version, relying on the descriptors used in the respective papers, and of modified versions that incorporate the four descriptor types discussed above. Specifically, for CNN-LSTM, we simply concatenate the vehicle dynamics descriptors and the motion-based descriptors to the appearance-based ones. For the Two-Stream baseline, we add a second two-stream sub-network for the vehicle dynamics and merge it with the appearance and motion streams by adding a fully-connected layer that takes as input the concatenation of the representation from the original two-stream sub-network and from vehicle dynamics two-stream sub-network. Finally, for MS-LSTM, we add a third stage that takes as input the concatenation of the second-stage representation with the vehicle dynamics descriptors. 

\subsection{Implementation Details}
\label{sec:implementation}
We make use of the DenseNet-121~\citep{huang2016densely}, pre-trained on ImageNet, to extract our appearance- and motion-based descriptors. Specifically, we replace the classifier with a fully-connected layer with 1024 neurons followed by a classifier with $N$ outputs, where $N$ is the number of classes. We fine-tune the resulting model using stochastic gradient descent for $10$ epochs with a fixed learning rate of $0.001$ and mini-batches of size $16$. Recall that, for the motion-based descriptors, the corresponding DenseNet relies on $2L$ flow stacks as input, which requires us to also replace the first layer of the network. To initialize the parameters of this layer, we average the weights over the three channels corresponding to the original RGB channels, and replicate these average weights $2L$ times~\citep{TSN}. We found this scheme to perform better than random initialization. 
\section{Benchmark Evaluation and Analysis}
We now report and analyze the results of our benchmarking experiments. For these experiments to be as extensive as possible given the available time, we performed them on a representative subset of VIENA$^2$ containing about 6.5K videos acquired in a large variety of environmental conditions and covering all 25 classes. This subset contains 277 samples per class, and thus still outsizes most action analysis datasets, as can be verified from Table~\ref{tbl:dataset_comparison}. The detailed statistics of this subset are provided in the supplementary material.

To evaluate the behavior of the algorithms in different conditions, we defined three different partitions of the data. The first one, which we refer to as \texttt{Random} in our experiments, consists of randomly assigning 70\% of the samples to the training set and the remaining 30\% to the test set. The second partition considers the daytime of the sequences, and is therefore referred to as \texttt{Daytime}. In this case, the training set is formed by the day images and the test set by the night ones. The last partition, \texttt{Weather}, follows the same strategy but based on the information about weather conditions, i.e., a training set of clear weather and a test set of rainy/snowy/... weather.

Below, we first present the results of our benchmarking on the \texttt{Random} partition, and then analyze the challenges related to our new dataset. We finally evaluate the benefits of our synthetic data for anticipation from real images, and analyze the bias of VIENA$^2$. Note that additional results including benchmarking on the other partitions and ablation studies of our MM-LSTM model are provided in the supplementary material. Note also that the scenarios and classes acronyms are defined in Section~\ref{sec:scenarios}.

\subsection{Action Anticipation on VIENA$^2$}
We report the results of our benchmark evaluation on the different scenarios of VIENA$^2$ in Table~\ref{tbl:baseline_original} for the original versions of the baselines, relying on the descriptors used in their respective paper, and in Table~\ref{tbl:baselines_full} for their modified versions that incorporate all descriptor types. Specifically, we report the recognition accuracies for all scenarios after every second of the sequences. Note that, in general, incorporating all descriptor types improves the results. Furthermore, while the action recognition baselines perform quite well in some scenarios, such as Accidents and Traffic Rules for the two-stream model, they are clearly outperformed by the anticipation methods in the other cases. Altogether, our new MM-LSTM consistently outperforms the baselines, thus showing the benefits of learning the dynamic importance of the modalities.

\begin{table}[t]
\centering
\small
\caption{Results on the \texttt{Random} split of VIENA$^2$ for the original versions our three baselines:  CNN+LSTM~\citep{LRCN} with only appearance, Two-Stream~\citep{feichtenhofer2016convolutional} with appearance and motion, and MS-LSTM (introduced in Chapter~\ref{cha:encouraging_lstms}) with action-aware and context-aware features.}
\label{tbl:baseline_original}
\scalebox{1}
{
\begin{tabular}{l| c@{ }@{ }c@{ }@{ }c@{ }@{ }c@{ }@{ }c| c@{ }@{ }c@{ }@{ }c@{ }@{ }c@{ }@{ }c| c@{ }@{ }c@{ }@{ }c@{ }@{ }c@{ }@{ }c}

&\multicolumn{5}{c}{\small CNN+LSTM}  & \multicolumn{5}{c}{\small Two-Stream} &\multicolumn{5}{c}{\small MS-LSTM}\\
 \hline
 & 1" & 2" & 3" & 4" & 5" & 1" & 2" & 3" & 4" & 5" & 1" & 2" & 3" & 4" & 5"\\
\hline

DM
&   22.8     &   24.2    &   26.5    &   27.9    &   28.0     
&   23.3     &   24.8    &   30.6    &   37.5    &   41.5    
&   22.4     &   28.1    &   37.5    &   42.6   &    44.0 \\

AC
&   53.6     &   53.6    &   55.0    &   56.3    &   57.0     
&   68.5     &   70.0    &   74.5    &   76.3    &   78.0    
&   50.3     &   55.6    &   60.4    &   68.3   &    72.5 \\

TR
&    26.6    &   28.3    &   29.5    &   30.1    &   32.1     
&    28.3    &   35.6    &   44.5    &   51.5    &   53.1    
&    30.7    &   33.4    &   41.0    &   49.8   &    52.3 \\

PI
&   38.4     &   40.4    &   41.8    &   41.8    &   42.1    
&   36.8     &   37.5    &   40.0    &   40.0    &   41.2    
&   50.6     &   52.4    &   55.6    &   56.8   &    58.3 \\

FCI
&   33.0     &   36.3    &   39.5    &    39.5   &    39.6    
&   37.1     &   38.0    &   35.5    &    39.3   &    39.3   
&   44.0     &   45.3    &   51.3    &    60.2  &     63.1 \\

\end{tabular}
}
\end{table}

\begin{table}[t]
\centering
\small
\caption{Results on the \texttt{Random} split of VIENA$^2$ for our three baselines with our action descriptors and for our approach.}
\label{tbl:baselines_full}
\scalebox{0.81}
{
\begin{tabular}{l| c@{ }@{ }c@{ }@{ }c@{ }@{ }c@{ }@{ }c| c@{ }@{ }c@{ }@{ }c@{ }@{ }c@{ }@{ }c| c@{ }@{ }c@{ }@{ }c@{ }@{ }c@{ }@{ }c|   c@{ }@{ }c@{ }@{ }c@{ }@{ }c@{ }@{ }c}

&\multicolumn{5}{c}{\small CNN+LSTM}  & \multicolumn{5}{c}{\small Two-Stream} &\multicolumn{5}{c}{\small MS-LSTM} &\multicolumn{5}{c}{\small Ours MM-LSTM}\\
 \hline
 & 1" & 2" & 3" & 4" & 5" & 1" & 2" & 3" & 4" & 5" & 1" & 2" & 3" & 4" & 5" & 1" & 2" & 3" & 4" & 5"\\
\hline

DM 
& 24.6 &   25.6 &   28.0 &   30.0 &  30.3  
& 26.8 & 30.5 &  40.4 &  53.4 &   62.6 
& 28.5 & 35.8 & 57.8 & 68.1 & 78.7 
& 32.0 &  38.5 &   60.5 &   71.5 &   83.6
\\

AC 
& 56.7 &  58.3 &   59.0 &  61.6 &  61.7  
& 70.0 &  72.0 &  74.0 &  77.1 &   79.7 
&  69.6 &   75.3 &   80.6 &   83.3 &   83.6 
& 76.3 &  79.0 &   81.7&   86.3 &   86.7\\

TR 
& 28.0 &  28.7 &  30.6 &  32.2 &  32.8  
& 30.6 &  38.7 &  48.0 & 49.6 &  54.1 
&  33.3 & 39.4 & 48.3 & 57.1 & 61.0 
& 39.8 &   49.8 &   58.8 &   63.7 &   68.8\\

PI
& 39.6 & 39.6 & 40.4 & 42.0 & 42.4
& 42.0 &  42.8 &  44.4 &  46.0 &  48.0 
&  55.8 & 57.6 & 62.6 &69.0 & 70.8 
& 57.3 &   59.7 &   68.9 &   72.5 &   73.3\\

FCI 
& 37.2 & 38.8 & 39.3 & 40.6 & 40.6
& 37.7 &  39.1 &  39.3 &  40.7 &   43.0
& 41.7 & 49.1 & 58.3 & 70.0 & 75.5
& 49.9 &   51.7 &  60.4 &   71.5 &   77.8
\end{tabular}
}
\end{table}

A comparison of the baselines with our approach on the \texttt{Daytime} and \texttt{Weather} partitions of VIENA$^2$ is provided in the supplementary material. In essence, the conclusions of these experiments are the same as those drawn above.
\subsection{Challenges of VIENA$^2$}
Based on the results above, we now study what challenges our dataset brings, such as which classes are the most difficult to predict and which classes cause the most confusion. We base this analysis on the per-class accuracies of our MM-LSTM model, which achieved the best performance in our benchmark. This, we believe, can suggest new directions to investigate in the future.

\begin{figure*}[t]
\centering
\includegraphics[width=\textwidth]{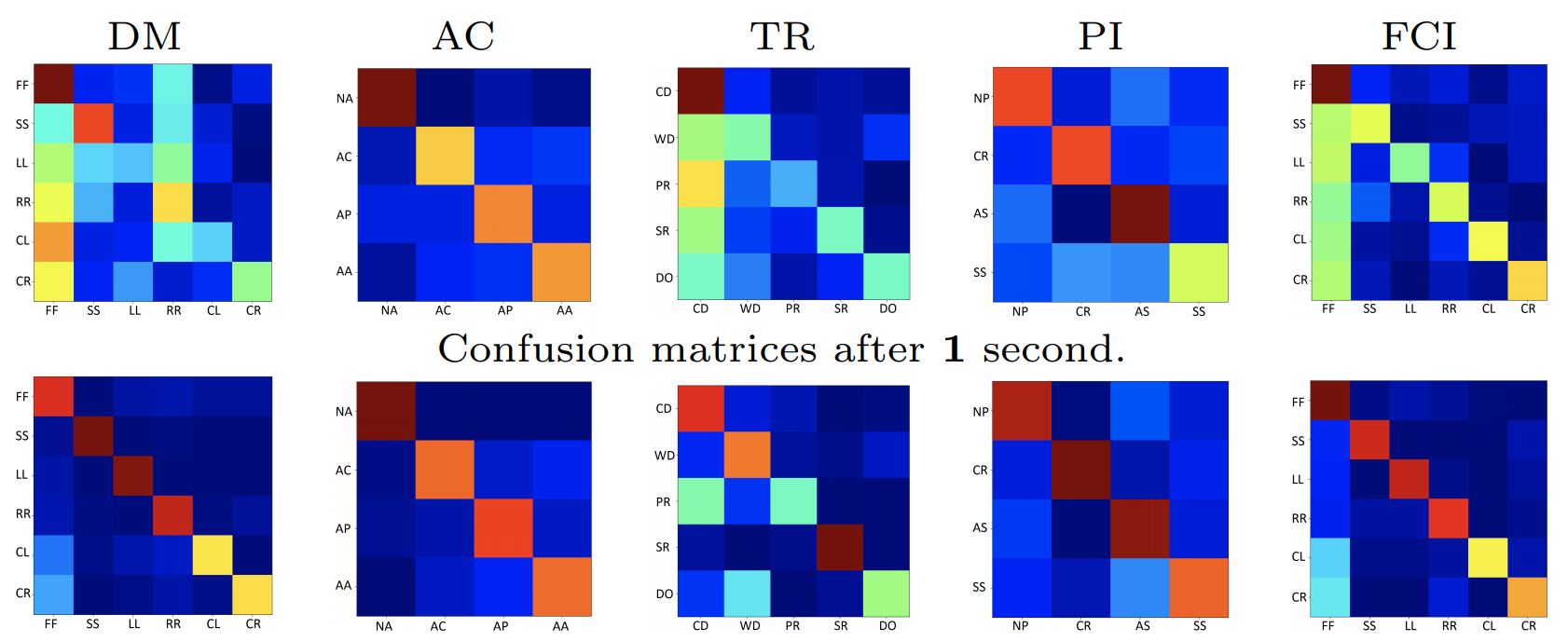}
\caption{\textbf{Confusion Matrices.} Confusion matrices of all five scenarios after observing 1 second (top) and 5 seconds (bottom) of each video sample.}
\label{fig:confusion}
\end{figure*}

Our MM-LSTM per-class accuracies are provided in Table~\ref{tbl:per_class}, and the corresponding confusion matrices at the earliest (after seeing 1 second) and latest (after seeing 5 seconds) predictions in Fig.~\ref{fig:confusion}. Below, we discuss the challenges of the various scenarios. 

\begin{table}[t]
\renewcommand{\arraystretch}{1.2}
\centering
\caption{Per-class accuracy of our approach on all scenarios of VIENA$^2$ (\texttt{Random}).}
\label{tbl:per_class}
\tiny
\scalebox{1}
{
\begin{tabular}{l @{ }@{ }|@{ }@{ } c@{ }@{ } c@{ }@{ } c@{ }@{ } c@{ }@{ } c@{ }@{ } c@{ }@{ } |@{ }@{ } c@{ }@{ } c@{ }@{ } c@{ }@{ } c@{ }@{ } |@{ }@{ } c@{ }@{ } c@{ }@{ } c@{ }@{ } c@{ }@{ } c@{ }@{ } |@{ }@{ } c@{ }@{ } c@{ }@{ } c@{ }@{ } c@{ }@{ } |@{ }@{ } c@{ }@{ } c@{ }@{ } c@{ }@{ } c@{ }@{ } c@{ }@{ } c}

&\multicolumn{6}{c}{\small DM} & \multicolumn{4}{c}{\small AC} & \multicolumn{5}{c}{\small TR} & \multicolumn{4}{c}{\small PI} & \multicolumn{6}{c}{\small FCI} \\
\hline
&{\scriptsize FF} & {\scriptsize SS} & {\scriptsize LL} & {\scriptsize RR} & {\scriptsize CL} & {\scriptsize CR}  
 & {\scriptsize NA} & {\scriptsize AP} & {\scriptsize AC} & {\scriptsize AA}
 & {\scriptsize CD} & {\scriptsize WD} & {\scriptsize PR} & {\scriptsize SR} & {\scriptsize DO}
 & {\scriptsize NP} & {\scriptsize CR} & {\scriptsize SS} & {\scriptsize AS}
 & {\scriptsize FF} & {\scriptsize SS} & {\scriptsize LL} & {\scriptsize RR} & {\scriptsize CL} & {\scriptsize CR}    \\
 \hline
1" &  50.7 &   43.8 &   17.8 &   35.0 &  18.7  &  26.1  
& 94.9 &   65.7 &   73.2 &   71.3
& 75.5 &   35.0 &   23.7 &   32.8 &  32.2 
& 59.3 &   59.1 &   68.8 &   42.2
& 74.5 &   46.3 &   35.6 &   44.6 &  47.8  & 50.9  
\\ 
2" &   60.1 &   46.8 &   26.3 &   38.7 &  27.1  &  32.1 
& 98.7 &   70.7 &   71.6 &   75.0
& 79.6 &   49.3 &   29.7 &  52.3  &  37.9 
& 63.0 &   51.2 &  71.4 &  53.4
& 76.9 &   48.6 &   37.1 &   45.9 &  49.6  &  52.0  
\\ 
3" &   81.3 &   75.6 &   54.4 &   63.4 &  42.9  &  45.4  
& 100 &   75.4 &   76.1 &   75.2
& 83.7 &   60.0 &  35.1 &  69.5 &  45.8 
& 70.4 &   67.6 &  79.2 &   58.6
& 85.7 &   63.9 &   54.1 &   50.7 &  52.7  &  57.3  
\\ 
4" &   81.2 &   87.3 &   72.9 &   77.3 &  55.4  &  55.0  
& 100 &   81.6 &   79.4 &   84.3
& 86.7 &  65.3 &   37.9 &  78.7 &  50.0 
& 72.2 &   75.9 &  80.1 &  61.35
& 89.1 &  77.8 &   74.4 &  69.5 & 56.1 & 62.2  
\\ 
5" &   88.0 &   97.2 &   95.8 &   90.4 &  64.9  &  65.4  
& 100 &   80.5 &   86.1 &   80.2
& 85.7 &   75.0 &  40.0 &  95.1 &  48.6 
& 74.1 &   78.2 &   76.6 &  63.6
& 91.2 &  83.5 &  84.6 &  81.4 & 59.4  &  66.8  
\\ 
\end{tabular}
}
\end{table}

\begin{enumerate}
\item {\bf Driver maneuver:} After 1s, most actions are mistaken for \emph{Moving Forward}, which is not surprising since the action has not started yet. After 5s, most of the confusion has disappeared, except for \emph{Changing Lane} (left and right), for which the appearance, motion and vehicle dynamics are subject to small changes only, thus making this action look similar to \emph{Moving Forward}.
\item {\bf Accident:} Our model is able to distinguish \emph{No Accident} from the different accident types early in the sequence. Some confusion between the different types of accident remains until after 5s, but this would have less impact in practice, as long as an accident is predicted.
\item {\bf Traffic rule:} As in the maneuver case, there is initially a high confusion with \emph{Correct Direction}, due to the fact that the action has not started yet. The confusion is then much reduced as we see more information, but \emph{Passing a Red Light} remains relatively poorly predicted.
\item {\bf Pedestrian intention:} The most challenging class for early prediction in this scenario is \emph{Pedestrian Walking along the Road}. The prediction is nevertheless much improved after 5s.
\item {\bf Front car intention:} Once again, at the beginning of the sequence, there is much confusion with the \emph{Forward} class. After 5s, the confusion is significantly reduced, with, as in the maneuver case, some confusion remaining between the \emph{Change lane} classes and the \emph{Forward} class, illustrating the subtle differences between these actions.
\end{enumerate}
\subsection{Benefits of VIENA$^2$ for Anticipation from Real Images}
\label{sec:real_images}
To evaluate the benefits of our synthetic dataset for anticipation on real videos, we make use of the JAAD dataset~\citep{rasouli2017agreeing} for pedestrian intention recognition, which is better suited to deep networks than other datasets, such as~\citep{ped_benchmark}, because of its larger size (58 videos vs. 346). This dataset is, however, not annotated with the same classes as we have in VIENA$^2$, as its purpose is to study pedestrian and driver behaviors at pedestrian crossings. To make JAAD suitable for our task, we re-annotated its videos according to the four classes of our \emph{Pedestrian Intention} scenario, and prepared a corresponding train/test split. JAAD is also heavily dominated by the \emph{Crossing} label, requiring augmentation of both training and test sets to have a more balanced number of samples per class. 

To demonstrate the benefits of  VIENA$^2$ in real-world applications, we conduct two sets of experiments: 1) Training on JAAD from scratch, and 2) Pre-training on VIENA$^2$ followed by fine-tuning on JAAD.
For all experiments, we use appearance-based and motion-based features, which can easily be obtained for JAAD. The results are shown in Table~\ref{tbl:jaad}. This experiment clearly demonstrates the effectiveness of using our synthetic dataset that contains photo-realistic samples simulating real-world scenarios.

\begin{table}[t]
\centering
\caption{\textbf{Anticipating actions on real data.} Pre-training our MM-LSTM with our VIENA$^2$ dataset yields higher accuracy than training from scratch on real data.
}
\label{tbl:jaad}
\begin{tabular}{l| @{ }@{ }@{ } c @{ }@{ }@{ } c @{ }@{ }@{ } c @{ }@{ }@{ } c @{ }@{ }@{ } c}
Model / Training & After 1" & After 2" & After 3" & After 4" & After 5" \\
\hline
MM-LSTM / From Scratch & 41.01\% & 45.84\% & 51.38\% & 54.94\% & 56.12\% \\
MM-LSTM / Fine-Tuned & 45.06\% & 54.15\% & 58.10\% & 65.61\% & 66.0\% \\
\end{tabular}
\end{table}

\begin{figure}[t]
\centering
\includegraphics[width=\textwidth]{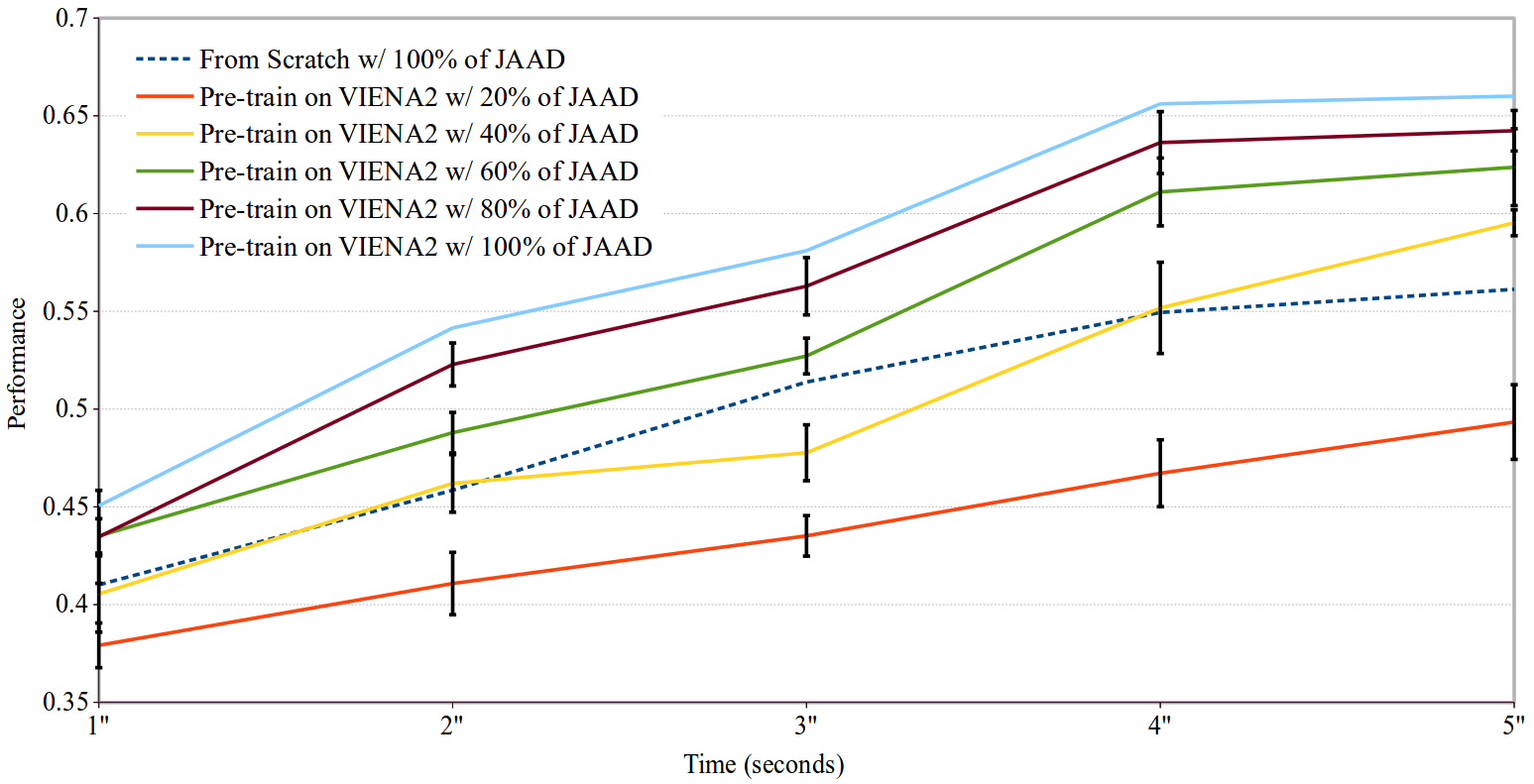}
\caption{{\bf Effect of the amount of real training data for fine-tuning MM-LSTM.} MM-LSTM was pre-trained on VIENA$^2$ in all cases, except for \emph{From Scratch w/100\% of JAAD} (dashed line). Each experiment was conducted with 10 random subsets of JAAD. We report the mean accuracy and standard deviation (error bars) over 10 runs.}
\label{fig:jaad_perc}
\end{figure}

Another potential benefit of using synthetic data is that it can reduce the amount of real data required to train a model. To evaluate this, we fine-tuned an MM-LSTM trained on VIENA$^2$ using a random subset of JAAD ranging from 20\% to 100\% of the entire dataset. The accuracies at every second of the sequence and for different percentages of JAAD data are shown in Fig.~\ref{fig:jaad_perc}. Note that with 60\% of real data, our MM-LSTM pre-trained on VIENA$^2$ already outperforms a model trained from scratch on 100\% of the JAAD data. This shows that our synthetic data can save a considerable amount of labeling effort on real images. 

\subsection{Bias Analysis}

For a dataset to be unbiased, it needs to be representative of the entire application domain it covers, thus being helpful in the presence of other data from the same application domain. This is what we aimed to achieve when capturing data in a large diversity of environmental conditions. Nevertheless, every dataset is subject to some bias. For example, since our data is synthetic, its appearance differs to some degree from real images, and the environments we cover are limited by those of the GTA V video game. However, below, we show empirically that the bias in VIENA$^2$ remains manageable, making it useful beyond evaluation on VIENA$^2$ itself.
In fact, the experiments of Section~\ref{sec:real_images} on real data already showed that performance on other datasets, such as JAAD, can be improved by making use of VIENA$^2$. To further evaluate the bias of the visual appearance of our dataset, we relied on the idea of domain adversarial training introduced in~\citep{ganin2015unsupervised}. In short, given data from two different domains, synthetic and real in our case, domain adversarial training aims to learn a feature extractor, such as a DenseNet, so as to fool a classifier whose goal is to determine from which domain a sample comes. If the visual appearance of both domains is similar, such a classifier should perform poorly. We therefore trained a DenseNet to perform action classification from a single image using both VIENA$^2$ and JAAD data, while learning a domain classifier to discriminate real samples from synthetic ones. The performance of the domain classifier quickly dropped down to chance, i.e., 50\%. To make sure that this was not simply due to failure to effectively train the domain classifier, we then froze the parameters of the DenseNet while continuing to train the domain classifier. Its accuracy remained close to chance, thus showing that the features extracted from both domains were virtually indistinguishable. Note that the accuracy of action classification improved from 18\% to 43\% during the training, thus showing that, while the features are indistinguishable to the discriminator, they are useful for action classification.

\begin{table}[t]
\centering
\small
\caption{Accuracy of MM-LSTM on the Driver Maneuver scenario when the training and test sets were captured by different users.}
\label{tab:new_driver}
\begin{tabular}{l@{ }@{ }@{ } l@{ }@{ }  | @{ }c @{ }@{ }@{ }c @{ }@{ }@{ }c @{ }@{ }@{ }c @{ }@{ }@{ }c}
Train Set & Test Set& & & & &\\
Captured by & Captured by & After 1" & After 2" & After 3" & After 4" & After 5" \\
\hline
User 1 & User 1 & 32.0\% &  38.5\% &   60.5\% &   71.5\% &   83.6\% \\
User 1 & User 2 & 32.8\% & 37.3\% & 60.7\% & 70.9\% & 82.8\% \\
\end{tabular}
\end{table}

In our context of synthetic data, another source of bias could arise from the specific users who captured the data. To analyze this, we trained an MM-LSTM model from the data acquired by a single user, covering all classes and all environmental conditions, and tested it on the data acquired by another user. In Table~\ref{tab:new_driver}, we compare the average accuracies of this experiment to those obtained when training and testing on data from the same user. Note that there is no significant differences, showing that our data generalizes well to other users.

\section{Benchmark Evaluation on All Splits of  VIENA$^2$}
We now report our results on the different scenarios of VIENA$^2$. In Table~\ref{tbl:weather} and Table~\ref{tbl:daytime}, we report the recognition accuracies for all scenarios after every second of the sequences. 
Note that some scenarios are easier to anticipate than others, e.g., accidents vs. driver maneuvers. 
Altogether, the trend remains the same as in the \texttt{Random} split.

\begin{table}[t]
\centering
\caption{Results on the \texttt{Weather} split of VIENA$^2$  for all scenarios and for our three baselines and our approach.}
\label{tbl:weather}
\scalebox{0.82}{
\begin{tabular}{l| c@{ }c@{ }c@{ }c@{ }c|  c@{ }c@{ }c@{ }c@{ }c|  c@{ }c@{ }c@{ }c@{ }c|  c@{ }c@{ }c@{ }c@{ }c}

&\multicolumn{5}{c}{CNN+LSTM}  & \multicolumn{5}{c}{Two-Stream} &\multicolumn{5}{c}{MS-LSTM} &\multicolumn{5}{c}{Ours MM-LSTM}\\
 \hline
 & 1" & 2" & 3" & 4" & 5" & 1" & 2" & 3" & 4" & 5" & 1" & 2" & 3" & 4" & 5" & 1" & 2" & 3" & 4" & 5"\\
\hline

DM 
& 22.7 &   22.8 &   25.1 &   28.7 &  30.3  
& 26.3 &  26.8 &  33.5 &  35.2 &   36.1 
& 26.4 &  32.5 &  53.2 &  59.8 &   72.3 
& 33.2 &  38.0 &   67.8 &   70.9 &   80.5
\\

AC 
& 51.5 &   53.7 &   53.9 &   55.2 &  55.8 
& 44.3 & 49.2 & 55.2 & 57.0 & 59.3
& 46.6 &  50.3 &  54.8 &  60.2 &   62.7 
& 50.1 &  53.6 &   62.2&   64.0 &   65.3\\

TR 
& 26.4 &   27.5 &   27.5 &   29.8 &  30.3
& 30.0  & 31.7 & 34.1 & 38.9 & 40.2
& 34.5 &  39.5 &  44.8 &  51.2 &   57.0 
& 39.1 &   45.7 &   54.9 &   59.6 &   65.6\\

PI
& 36.3 &   37.5 &   37.9 &   40.0 &  40.4  
& 38.2 & 38.9 & 41.5 & 47.9 & 51.4
& 42.5 &  48.3 &  52.9 &  55.8 &   62.4 
& 48.0 &   59.5 &   65.6 &   69.9 &   72.0\\

FCI 
& 34.0 &   36.9 &   38.4 &  40.1 &  41.0
& 35.4 &  36.3 &  45.0 &  48.2 &   54.2 
& 39.8 &  44.3 &  53.4 &  62.8 &   70.2 
& 42.9 &   49.1 &  60.7 &   70.2 &   76.7\\

\end{tabular}
}
\end{table}

\begin{table}[t]
\centering
\caption{Results on the \texttt{Daytime} split of VIENA$^2$ for all scenarios and for our three baselines and our approach.}
\label{tbl:daytime}
\scalebox{0.82}{
\begin{tabular}{l| c@{ }c@{ }c@{ }c@{ }c|  c@{ }c@{ }c@{ }c@{ }c|  c@{ }c@{ }c@{ }c@{ }c|  c@{ }c@{ }c@{ }c@{ }c}

&\multicolumn{5}{c}{CNN+LSTM}  & \multicolumn{5}{c}{Two-Stream} &\multicolumn{5}{c}{MS-LSTM} &\multicolumn{5}{c}{Ours MM-LSTM}\\
 \hline
 & 1" & 2" & 3" & 4" & 5" & 1" & 2" & 3" & 4" & 5" & 1" & 2" & 3" & 4" & 5" & 1" & 2" & 3" & 4" & 5"\\
\hline

DM 
& 23.4 &   24.1 &  27.2 &   27.8 &  30.0  
& 26.3 &  28.4 &  33.5 &  40.2 &   42.4 
& 27.0 &  34.2 &  55.3 &  62.1 &   75.9
& 33.3 &  40.4 &  57.2 &   77.6 &   84.3\\

AC 
& 48.4 &   52.3 &   52.8 &   55.9 &  57.4  
& 47.2 & 49.4 & 51.0 & 55.9 & 60.0 
& 48.0 &  50.4 &  53.2 &  57.8 &   61.6
& 50.7 &  54.5 &   59.7 &   64.7 &   66.2\\

TR 
& 26.3 &  26.9 &   27.3 &   29.8 &  31.1 
& 29.1 & 34.6 & 36.8 & 38.0 & 44.4
& 33.1 &  36.8 &  42.5 &  49.3 &   54.7 
& 35.0 &   39.5 &   49.6 &   61.1 &   68.2\\

PI
& 38.0 &   38.2 &   39.3 &   40.1 &  41.3  
& 39.3 & 40.9 & 44.1 & 50.2 & 55.5
& 46.5 &  53.3 &  60.0 &  66.1 &   68.7
& 49.0 &  61.8 &   69.1 &   70.1 &   70.1\\

FCI 
& 33.5 &   36.0 &   37.5 &   38.8 & 39.0
& 34.5 &  35.2 &  38.8 &  44.3 &   49.6 
& 41.1 &  45.5 &  53.7 &  56.7 &   62.2 
& 41.8 &  45.1 &  57.1 &   69.4 &   77.1\\

\end{tabular}
}
\end{table}

\section{Ablation Study for our MM-LSTM}
In this section, we evaluate different aspects of our new MM-LSTM model.

\paragraph{Effect of Loss Function}
We first evaluate the influence of our new loss function on our results. To this end, we replaced it with the standard cross-entropy loss, and with the loss of~\citep{sadegh2017encouraging}, which has proven effective at action anticipation and corresponds to a linear weight instead of our nonlinear one. In Table~\ref{effect_loss}, we compare the results of our MM-LSTM model trained with these losses and with ours on the \emph{Accidents} scenario of VIENA$^2$. Note that our loss yields more accurate predictions, particularly in the early stages of the videos. This confirms the effectiveness of our false positive weighting strategy, well-suited for action anticipation.

\begin{table}[!h]
\centering

\caption{{\bf Influence of the loss function.} We compare the use of the standard cross-entropy, the loss introduced in Chapter~\ref{cha:encouraging_lstms} and our loss within our MM-LSTM model using the \emph{Accidents} of VIENA$^2$. Note that our loss consistently yields higher accuracy, particularly in the early stages of the videos.}
\label{effect_loss}
\begin{tabular}{l@{ }@{ } @{ }@{ }  c @{ }@{ } @{ }@{ } c @{ }@{ } @{ }@{ } c @{ }@{ } @{ }@{ } c @{ }@{ } @{ }@{ } c}
 & After & After & After & After & After \\ 
Loss Function & 1 sec & 2 sec & 3 sec & 4 sec & 5 sec \\
\hline
\multicolumn{6}{c}{wo/ AvgPool}\\
\hline
Cross Entropy  & 48.3\% & 60.3\% & 72.0\% & 76.8\% & 76.9\%\\
Linear weight (as in Chapter~\ref{cha:encouraging_lstms})  & 45.2\% & 58.0\% & 70.2\% & 77.2\% & 77.7\%\\
Ours & \textbf{56.8\%} & \textbf{67.0\%} & \textbf{76.7\%} & \textbf{80.4\%} & \textbf{80.3\%}\\
\hline
\multicolumn{6}{c}{w/ AvgPool}\\
\hline
Cross Entropy  & 69.3\% & 74.3\% & 78.0\% & 82.0\% & 83.3\%\\
Linear weight (as in Chapter~\ref{cha:encouraging_lstms})  & 73.3\% & 75.7\% & 79.0\% & 82.7\% & 83.3\%\\
Ours  & \textbf{76.3\%} & \textbf{79.0\%} & \textbf{81.7\%} & \textbf{86.3\%} & \textbf{86.7\%}\\
\end{tabular}
\end{table}

\begin{table}[!h]
\centering
\caption{{\bf Influence of different features.} We evaluate the accuracy obtained by using our different features individually on the \emph{Accidents} and \emph{Front car intention} scenarios for the \emph{Random} partition. See the discussion in the text for more detail.}
\label{effect_feature}
\begin{tabular}{l @{ }@{ } @{ }@{ } c @{ }@{ } @{ }@{ } c @{ }@{ } @{ }@{ }c @{ }@{ } @{ }@{ } c @{ }@{ } @{ }@{ } c}
 & After & After & After & After & After \\ 
Features & 1 sec & 2 sec & 3 sec & 4 sec & 5 sec \\
\hline
\multicolumn{6}{c}{Accident} \\
\hline
Steering  & 26.0\% & 27.5\% & 27.7\% &  28.2\% & 28.7\%\\
Speed  & 46.3\% & 46.3\% & 48.8\% & 49.3\% & 50.7\%\\
Appearance & 48.1\% & 48.7\% & 51.0\% & 51.3\% & 53.7\%\\
Motion & 21.7\% & 23.6\%& 24.3\%& 24.3\%& 25.5\%\\
All & \textbf{76.3\%} & \textbf{79.0\%} & \textbf{81.7\%} & \textbf{86.3\%} & \textbf{86.7\%}\\
\hline
\multicolumn{6}{c}{Front car intention}\\
\hline
Steering & 34.2\% &   33.3\% &   34.7\% &   34.7\% &   35.0\% \\
Speed  & 17.3\% &   15.0\% &   16.2\% &   16.5\% &  17.3\% \\
Appearance & 29.8\% &   30.3\% &   29.9\% &   29.6\% &   30.3\% \\
Motion & 34.7\% &   36.9\% &   42.0\% &   47.0\% &   55.3\% \\
All & \textbf{49.9\%} &    \textbf{51.7\%} &    \textbf{60.4\%} &    \textbf{71.5\%} &    \textbf{77.8\%} \\
\end{tabular}
\end{table}

\paragraph{Effect of Different Features}
We then analyze the impact of the different features on our results. To this end, we trained separate MM-LSTM models with a single modality as input. We then evaluate these models on the \emph{Accidents} and \emph{Front car intention} scenarios of VIENA$^2$. The results are provided in Table~\ref{effect_feature}. Note that, in the former scenario, speed and appearance play an important role, whereas in the latter one, the steering angle and motion are the dominant features, closely followed by appearance. This shows that different scenarios need to rely more strongly on different features. Nevertheless, using all descriptors jointly always improves over using them individually, which evidences that our MM-LSTM model can effectively discover the importance of the different modalities.

\paragraph{Effect of the Number of Hidden Units}
Furthermore, we study the effect of varying the number of hidden units in the LSTM blocks of our MM-LSTM model. To this end, we simultaneously varied this number between 256 and 4096 across all LSTM units in our model. As can be seen in Table~\ref{tbl:hidden}, where we provide the detailed numbers for this analysis for the Random partition, the results are stable and only marginally improving with more hidden units. However, the number of parameters of the MM-LSTM model increases drastically, in this case from $7.6M$ to $436.3M$. In our experiments in this chapter, we therefore used 1024 hidden units for all LSTMs, which is a good balance between performance and number of parameters ($46.2M$). 

\begin{table}[!h]
\centering
\caption{{\bf Effect of number of LSTM hidden units on anticipation performance.} Using 1024 units represents a good trade-off between memory consumption and accuracy.}
\label{tbl:hidden}
\begin{tabular}{l @{ }@{ } @{ }@{ } | l @{ }@{ } @{ }@{ } | c @{ }@{ } @{ }@{ } c @{ }@{ } @{ }@{ } c @{ }@{ } @{ }@{ } c @{ }@{ } @{ }@{ } c}
& & After & After & After & After & After\\
Units & Params & 1 sec  & 2 sec  & 3 sec  & 4 sec  & 5 sec \\
\hline
256 & 7.6M & 68.77\% & 71.33\% & 77.50\% & 82.11\% & 82.32\% \\
512 & 17.8M & 75.01\% & 77.78\% & 79.90\% & 84.63\% & 86.10\% \\
1024 & 46.2M & 76.31\% & 79.00\% & 81.73\% & 86.29\% & 86.68\% \\
2048 & 134.3M & 76.45\% & 79.22\% & 82.01\% & 86.43\% & 87.01\% \\
4096 & 436.3M & 76.66\% & 78.12\% & 83.33\% & 86.37\% & 86.60\% \\
\end{tabular}
\end{table}

\paragraph{Effect of Input Block}
As mentioned earlier, the FC-Pool layer of our MM-LSTM model encodes the importance of the modalities and is shared across time. While, on its own, this cannot change the importance of the modalities over time, the modality-specific LSTMs can adapt their outputs over time so as to serve this purpose. To further evidence the importance of these two modules in learning a good combination of the modalities, we performed an experiment where we deactivated them in turn, and trained models on all 24 permutations of the 4 input modalities. The results in Table~\ref{tbl:input_block}, where we provide the average accuracy and standard deviation over the 24 permutations, show that using both modules not only yields higher accuracy, but also makes our network virtually unaffected by the order of the modalities, i.e., very small standard deviation. 

\begin{table*}[!h]
\centering
\small
\caption{Importance of the \emph{Input Block} to learn a good combination of the modalities. We report the mean accuracy and standard deviation over the 24 permutations of the 4 input modalities for the Accident scenario. Note that using both modality-specific LSTMs (L) and the FC-Pool (P) layer yields not only the highest accuracies, but also very low std, thus indicating that the modality order has virtually no impact on our results.}
\label{tbl:input_block}
\begin{tabular}{c c | c c c c c}
LSTM & FC-Pool & After 1" & After 2" & After 3" & After 4" & After 5" \\
\hline
w/ & w/ & \textbf{76.38$\pm$0.16} & \textbf{79.08$\pm$0.18} & \textbf{81.52$\pm$0.16} & \textbf{86.30$\pm$0.25} & \textbf{86.56$\pm$0.15} \\

w/ & wo/ & 70.04$\pm$2.37 & 73.31$\pm$3.11 & 76.41$\pm$3.05 & 80.78$\pm$2.37 & 82.25$\pm$2.78 \\

wo/ & w/ & 50.58$\pm$4.02 & 51.48$\pm$4.01 & 53.21$\pm$3.63 & 54.47$\pm$3.10 & 56.03$\pm$3.24 \\

wo/ & wo/ & 48.48$\pm$4.71 & 49.27$\pm$4.36 & 50.76$\pm$3.73 & 52.58$\pm$3.06 & 53.57$\pm$2.25 \\

\end{tabular}
\end{table*}

\paragraph{Comparison to the State of the Art}

To evaluate the effectiveness of our model on another dataset, and compare it to the state-of-the-art anticipation techniques, we make use of the standard JHMDB-21 benchmark. For this experiment, we followed the experimental setup of~\citep{sadegh2017encouraging} as we did in Chapter~\ref{cha:encouraging_lstms}, and used their context-aware and action-aware features instead of our features, to make the comparison more fair. The results of several state-of-the-art methods on this dataset are shown in Table~\ref{tbl:architecture}. Note that we outperform all these methods. Importantly, we outperform the multi-stage LSTM introduced in Chapter~\ref{cha:encouraging_lstms}, which manually defines the order in which the two feature types should be considered. This, we believe, truly evidences the benefits of our approach, which can automatically 
learn the dynamic importance of the input modalities for the final prediction. Moreover, we evaluated our model when trained with the loss function introduced in Chapter~\ref{cha:encouraging_lstms}. Note that, in the JHMDB-21 dataset, all actions start from the beginning of the videos and continue to the end. This, however, is not the case in VIENA$^2$, where the actions begin to occur only during the second half of the video. Therefore, the loss function introduced in Chapter~\ref{cha:encouraging_lstms}, which linearly grows over time, performs better on JHMDB-21, as it is better suited for such cases, but performs worse on VIENA$^2$, as shown in Table~\ref{effect_loss}. In other words, the design of the loss function should consider the nature of the target dataset. JHMDB depicts non-driving activities, where the actions occur over the entire duration of the video. This is in fact a case where 
linearly growing weights better model the way the action occurs, i.e., linearly over the entire sequence, and thus perform better. Note, however, that the best accuracy in Table~\ref{tbl:architecture} is still achieved by our approach, but by using linear weights, and that our approach with nonlinear weights follows very closely, thus showing good robustness to the precise definition of the weights.

\begin{table}[!h]
\centering
\caption{{\bf Comparison to the state-of-the-art anticipation techniques on JHMDB-21.} Note that our MM-LSTM architecture outperforms all the baselines, thus achieving state-of-the-art performance on this dataset.}
\label{tbl:architecture}
\begin{tabular}{l@{ }@{ } @{ }@{ }  c@{ }@{ } @{ }@{ }  c}
 & Earliest & Latest \\
Architecture / Method &  (20\%) &  (100\%) \\
\hline
DP-SVM~\citep{soomro2016online} & 5.0\% & 46.0\% \\
S-SVM~\citep{soomro2016online} & 5.0\% & 43.0\% \\
Where/What~\citep{soomro2016predicting} & 10.0\% & 43.0\%\\
Ranking Loss~\citep{ma2016learning} & 29.0\% & 43.0\% \\
MS-LSTM (Our approach in Chapter~\ref{cha:encouraging_lstms}) & 55.0\% & 58.0\% \\
\hline
MM-LSTM w/ Loss of~\citep{sadegh2017encouraging} & \textbf{59.3\%} & 61.0\% \\
MM-LSTM w/ our Loss & 58.1\% & \textbf{62.7\%} \\
\hline
\end{tabular}
\end{table}

\section{VIENA$^2$ Visualization: Environmental Diversity}
\begin{figure}[!h]
\centering
\includegraphics[width=\textwidth]{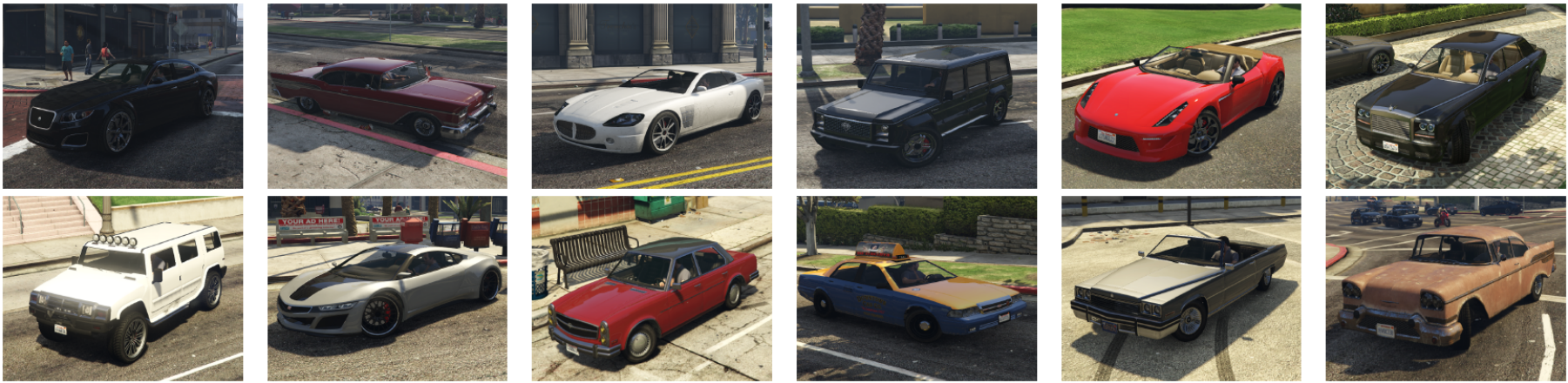}
\caption{Examples of cars used to capture our data.}
\label{fig:cars}
\end{figure}

\begin{figure}[!h]
\centering
\includegraphics[width=\textwidth]{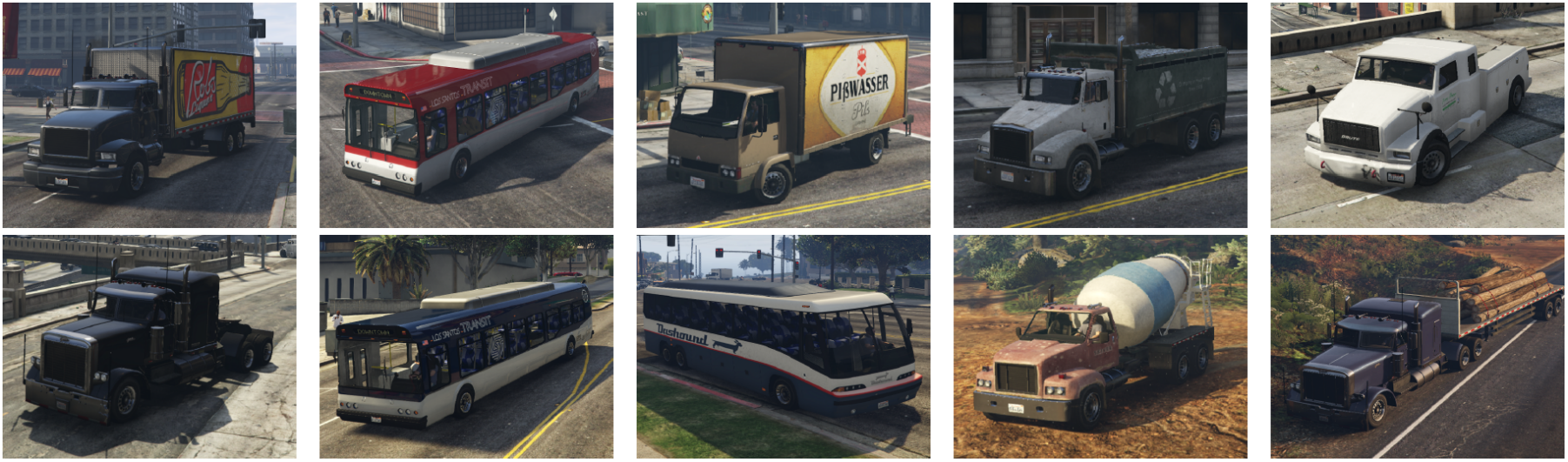}
\caption{Examples of heavy vehicles (trucks and buses) used to capture data for heavy vehicles' driver maneuvers.}
\label{fig:heavy}
\end{figure}

To collect data for the different scenarios of VIENA$^2$, we made use of various vehicles, including heavy vehicles for Driver Maneuver of heavy vehicles' driver. To provide diversity in terms of vehicle conditions, type, etc., we made use of different kinds of cars, such as old cars that are harder to control during turns or at high speed, and sport cars that are better-suited when driving at high speed (see Fig.~\ref{fig:cars}). For heavy vehicles, we collected data with two types of vehicles, trucks and buses. However, again for the sake of diversity, we made use of heavy vehicles with different specifications, top speed, and length (see Fig.~\ref{fig:heavy}).

\begin{figure}[!h]
\centering
\includegraphics[width=\textwidth]{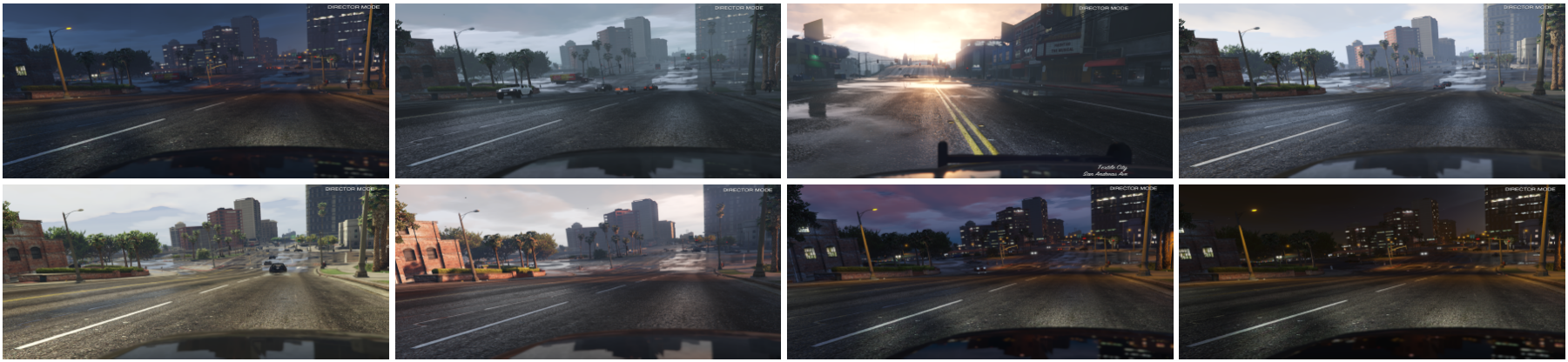}
\caption{Examples of different daytime conditions. From top-left to bottom-right: Pre-dawn, dawn, sunrise, morning, midday, sunset, dusk, and midnight.}
\label{fig:daytime}
\end{figure}

\begin{figure}[!h]
\centering
\includegraphics[width=\textwidth]{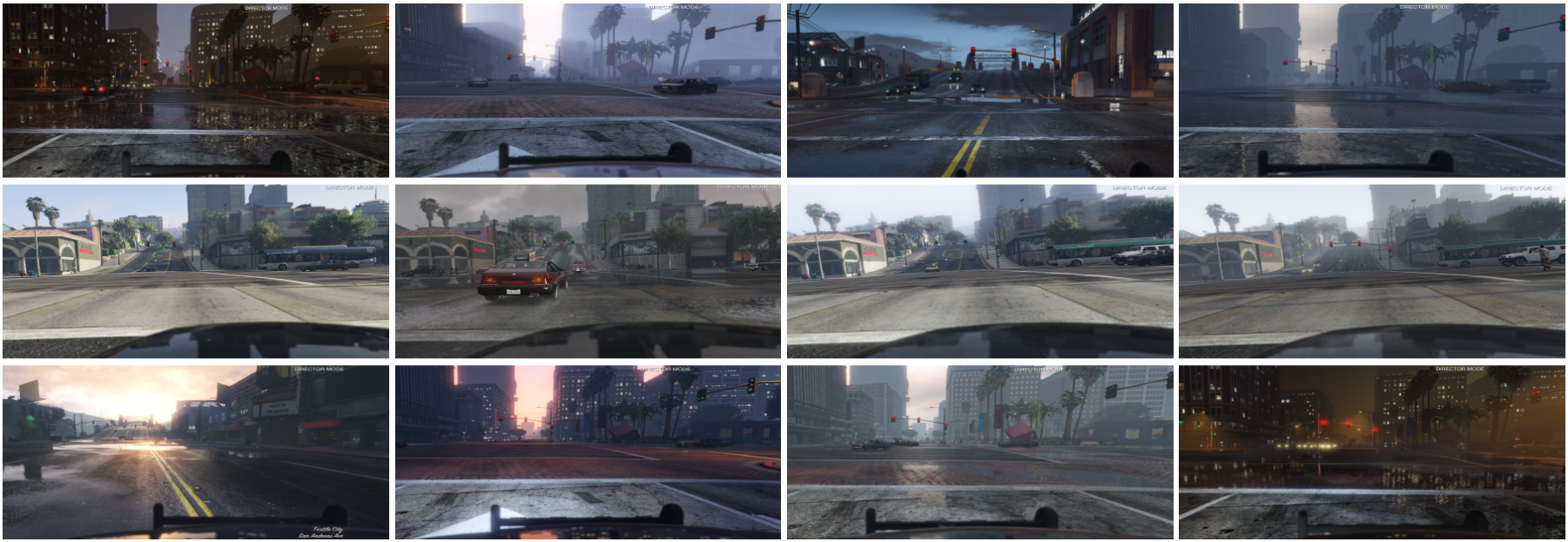}
\caption{Examples of different weather conditions (from a single view/location), including rainy, snowy, foggy, hazy, and clear.}
\label{fig:weather}
\end{figure}

\begin{figure}[!h]
\centering
\includegraphics[width=\textwidth]{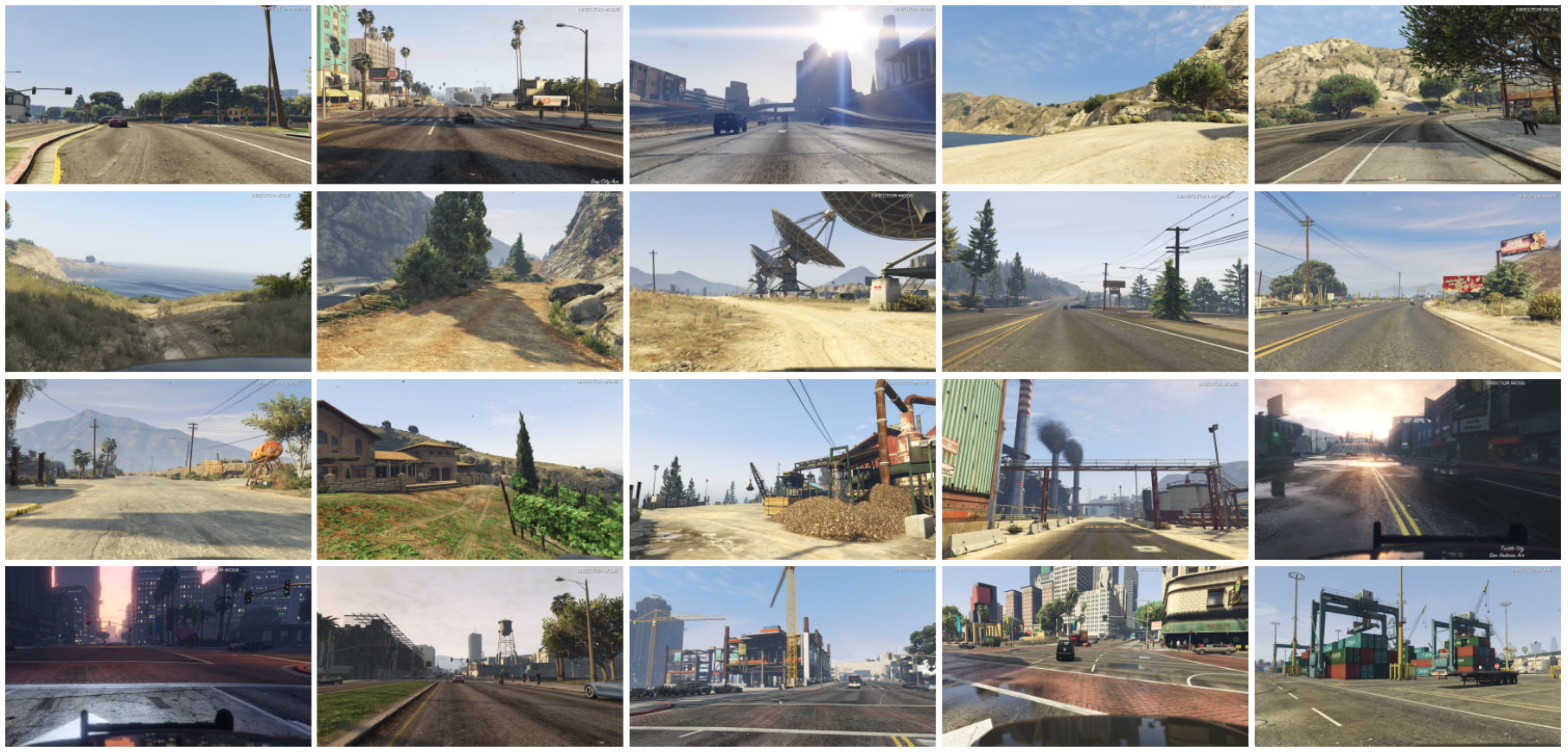}
\caption{Examples of different locations, including suburbs, downtown, off-road, nature, highway, and industrial locations.}
\label{fig:location}
\end{figure}

Note that, for all of our videos, the camera was mounted on the car so as to provide us with a front view. For heavy vehicles, the camera was mounted in a similar manner, but providing a higher viewpoint, since heavy vehicles are taller. In Figs.~\ref{fig:daytime},~\ref{fig:weather} and~\ref{fig:location}, we provide example images acquired at different daytimes, i.e., from pre-dawn to midnight, in different weather conditions and at different locations, respectively. We believe that the combination of these conditions, i.e., different vehicles, daytimes, weathers, and locations, provide a very diverse set of data which will reduce the effect of dataset bias.

\section{VIENA$^2$ Visualization: Scenarios}
\begin{figure}[!h]
\centering
\includegraphics[width=\textwidth]{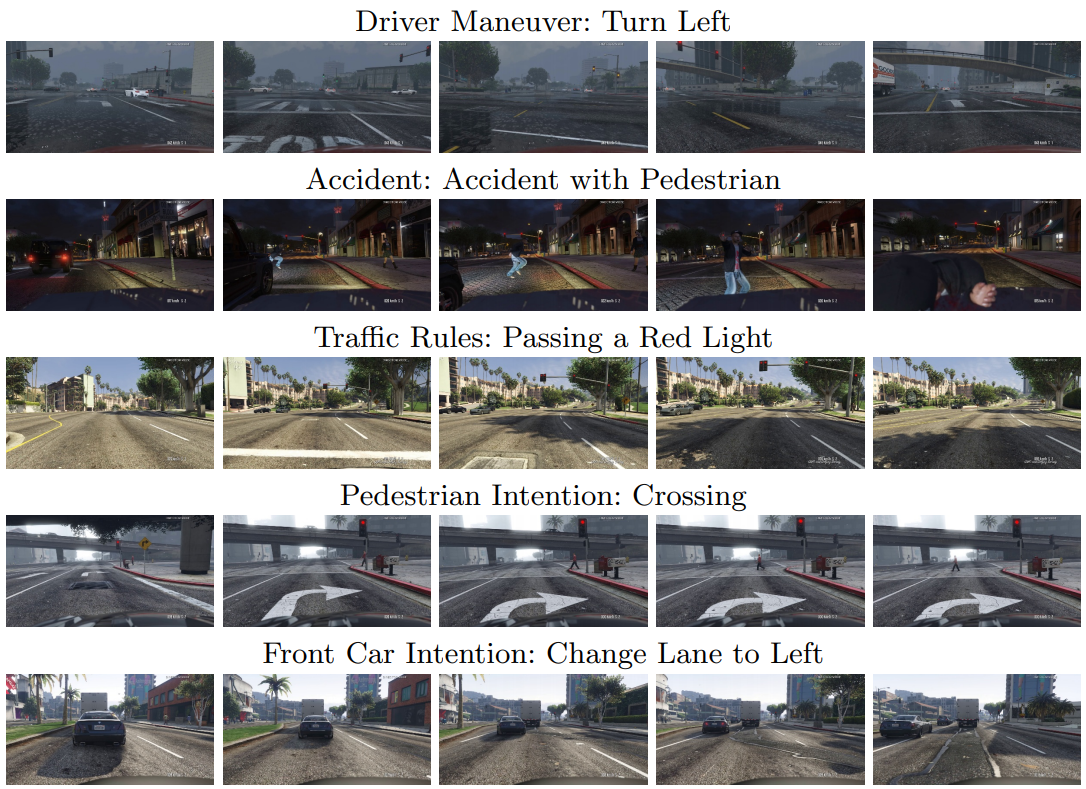}
\caption{Examples of our action classes in different scenarios. Each row represents an action example from one scenario.}
\label{fig:scenario}
\end{figure}

In Fig.~\ref{fig:scenario}, we visualize samples from the different classes of different scenarios of VIENA$^2$.

\section{Conclusion} 
\label{sec:conclusion}
In this chapter, we have introduced a new large-scale dataset for general action anticipation in driving scenarios, which covers a broad range of situations with a common set of sensors. Furthermore, we have proposed a new MM-LSTM architecture allowing us to learn the importance of multiple input modalities for action anticipation. Our experimental evaluation has shown the benefits of our new dataset and of our new model.
This chapter also concludes our contributions on the task of anticipating a discrete representation of a deterministic future. In the next chapter, we start studying the problem of anticipating a continuous representation of a stochastic future, with the focus on diverse human motion prediction.

%% file: chapter3.tex
\chapter{A Stochastic Conditioning Scheme for Diverse Human Motion Prediction}
\label{cha:cvpr}

In this chapter, we start studying the problem of anticipating continuous future representations of a stochastic process. That is, for instance, the case of human motion prediction. 
Human motion prediction, the task of predicting future 3D human poses given a sequence of observed ones, has been mostly treated as a deterministic problem. However, human motion is a stochastic process: Given an observed sequence of poses, multiple future motions are plausible. 
Existing approaches to modeling this stochasticity typically combine a random noise vector with information about the previous poses. This combination, however, is done in a deterministic manner, which gives the network the flexibility to learn to ignore the random noise.
Alternatively, in this chapter, we propose to stochastically combine the root of variations with previous pose information, so as to force the model to take the noise into account.
We exploit this idea for motion prediction by incorporating it into a recurrent encoder-decoder network with a conditional variational autoencoder block that learns to exploit the perturbations.
Our experiments on two large-scale motion prediction datasets demonstrate that our model yields high-quality pose sequences that are much more diverse than those from state-of-the-art stochastic motion prediction techniques.

\section{Introduction}
Human motion prediction aims to forecast the sequence of future poses of a person given past observations of such poses. To achieve this, existing methods typically rely on recurrent neural networks (RNNs) that encode the person's motion~\citep{martinez2017human,gui2018adversarial,walker2017pose,kundu2018bihmp,barsoum2018hp,pavllo2019modeling,pavllo2018quaternet}. While they predict reasonable motions, RNNs are deterministic models and thus cannot account for the highly stochastic nature of human motion; given the beginning of a sequence, multiple, diverse futures are plausible. To correctly model this, it is therefore critical to develop algorithms that can learn the \emph{multiple modes} of human motion, even when presented with only deterministic training samples.

Recently, several attempts have been made at modeling the stochastic nature of human motion~\citep{yan2018mt,barsoum2018hp,walker2017pose,kundu2018bihmp,lin2018human}. These methods rely on sampling a random vector that is then combined with an encoding of the observed pose sequence. In essence, this combination is similar to the conditioning of generative networks; the resulting models aim to generate an output from a random vector while taking into account additional information about the content.

While standard conditioning strategies, i.e., concatenating the condition to the latent variable, may be
effective for many tasks, as in~\citep{yan2016attribute2image,kulkarni2015deep,esser2018variational,engel2017latent,bao2017cvae,larsen2015autoencoding}, they are ill-suited for motion prediction. The reason is the following: In other tasks, the conditioning variable only provides auxiliary information about the output to produce,
such as the fact that a generated face should be smiling. By contrast, in motion prediction, it typically contains the core signal to produce the output, i.e., the information about the previous poses. We empirically observed that, since the prediction model is trained using deterministic samples (i.e., one condition per sample), it can then simply learn to ignore the random vector and still produce a meaningful output based on the conditioning variable only. In other words, the model can ignore the root of variations, and thus essentially become deterministic. 
This problem was discussed in~\citep{bowman2015generating} in the context of unconditional text generation, and we identified it in our own motion prediction experiments. 

We introduce a simple yet effective approach to counteracting this loss of diversity and thus to generating truly diverse future pose sequences. At the heart of our approach lies the idea of \emph{Mix-and-Match} perturbations: Instead of combining a noise vector with the conditioning variables in a deterministic manner, we randomly select and perturb a subset of these variables. By randomly changing this subset at every iteration, our strategy prevents training from identifying the root of variations and forces the model to take it into account in the generation process. Consequently, as supported by our experiments, our approach produces not only high-quality predictions but also truly diverse ones. 

In short, our contributions in this chapter are \emph{(i)} a novel way of imposing diversity into conditional VAEs, called \emph{Mix-and-Match perturbations}; \emph{(ii)} a new motion prediction model capable of generating multiple likely future pose sequences from an observed motion; \emph{(iii)} a new set of evaluation metrics for quantitatively measuring the quality and the diversity of generated motions, thus facilitating the comparison of different stochastic approaches; and \emph{(iv)} a curriculum learning paradigm for training generative models that use Mix-and-Match perturbation as the stochastic conditioning scheme. Despite its simplicity, curriculum learning of variation is essential to achieve optimal performance in case of imposing large variations.

\section{Related Work}
\label{sec:related_work}
\noindent\textbf{Deterministic Motion Prediction.}
Most motion prediction approaches are based on \emph{deterministic} models~\citep{pavllo2018quaternet,pavllo2019modeling,gui2018adversarial,jain2016structural,martinez2017human,gui2018few,fragkiadaki2015recurrent,ghosh2017learning,mao2019learning}, casting motion prediction as a regression task where only one outcome is possible given the observations. 
Due to the success of RNN-based methods at modeling sequence-to-sequence learning problems, many attempts have been made to address motion prediction within a recurrent framework~\citep{martinez2017human,gui2018adversarial,walker2017pose,kundu2018bihmp,barsoum2018hp,pavllo2019modeling,pavllo2018quaternet}. Typically, these approaches try to learn a mapping from the observed sequence of poses to the future sequence. Another group of study addresses this problem within feed-forward models~\citep{mao2019learning,li2018convolutional,butepage2017deep}, either with fully-connected~\citep{butepage2017deep}, convolutional~\citep{li2018convolutional}, or more recently, graph neural networks~\citep{mao2019learning}.
While a deterministic approach may produce accurate predictions, it fails to reflect the stochastic nature of human motion, where multiple plausible outcomes can be highly likely for a single given series of observations. Modeling this diversity is the topic of this chapter, and we therefore focus the discussion below on the other methods that have attempted to do so.

\noindent\textbf{Stochastic Motion Prediction.}
The general trend to incorporate variations in the predicted motions consists of combining information about the observed pose sequence with a random vector. In this context, two types of approaches have been studied: The techniques that directly incorporate the random vector into the RNN decoder, e.g., as in GANs, and those that make use of an additional Conditional Variational Autoencoder (CVAE)~\citep{sohn2015learning} to learn a latent variable that acts as the root of variation.

In the first class of methods,~\citep{lin2018human} sample a random vector $z_t\sim\mathcal{N}(0,I)$ at each time step and add it to the pose input to the RNN decoder. By relying on different random vectors at each time step, however, this strategy is prone to generating discontinuous motions. To overcome this,~\citep{kundu2018bihmp} make use of a single random vector to generate the entire sequence. This vector is both employed to alter the initialization of the decoder and concatenated with a pose embedding at each iteration of the RNN. By relying on concatenation as a mean to fuse the condition and the random vector, these two methods contain parameters that are specific to the random vector, and thus give the model the flexibility to ignore this information.
In~\citep{barsoum2018hp}, instead of using concatenation, the random vector is added to the hidden state produced by the RNN encoder. While addition prevents having parameters that are specific to the random vector, this vector is first transformed by multiplication with a parameter matrix, and thus can again be zeroed out so as to remove the source of diversity, as we observe empirically in Section~\ref{sec:ablation}. 

The second category of stochastic methods introduce an additional CVAE between the RNN encoder and decoder. This allows them to learn a more meaningful transformation of the noise, combined with the conditioning variables, before passing the resulting information to the RNN decoder. In this context,~\citep{walker2017pose} propose to directly use the pose as conditioning variable. As will be shown in our experiments, while this approach is able to maintain some degree of diversity, albeit less than ours, it yields motions of lower quality because of its use of independent random vectors at each time step. 
In~\citep{butepage2018anticipating}, an approach similar to that of~\citep{walker2017pose} is proposed, but with one CVAE per limb. As such, this method suffers from the same discontinuity problem as~\citep{walker2017pose,lin2018human}. Finally, instead of perturbing the pose, the recent work of~\citep{yan2018mt} uses the RNN decoder hidden state as conditioning variable in the CVAE, concatenating it with the random vector. While this approach generates high-quality motions, it suffers from the fact that the CVAE decoder gives the model the flexibility to ignore the random vector. 

Ultimately, both classes of methods suffer from the fact that they allow the model to ignore the random vector, thus relying entirely on the conditioning information to generate future poses. Here, we introduce an effective way to maintain the root of diversity by randomizing the combination of the random vector with the conditioning variable.

\section{Proposed Method}
In this section, we first introduce our \emph{Mix-and-Match} approach to introducing diversity in CVAE-based motion prediction. We then describe the motion prediction architecture we used in our experiments and propose a novel evaluation metric to quantitatively measure the diversity and  quality of generated motions.

\begin{figure}[!h]
    \centering
    \includegraphics[width=\textwidth]{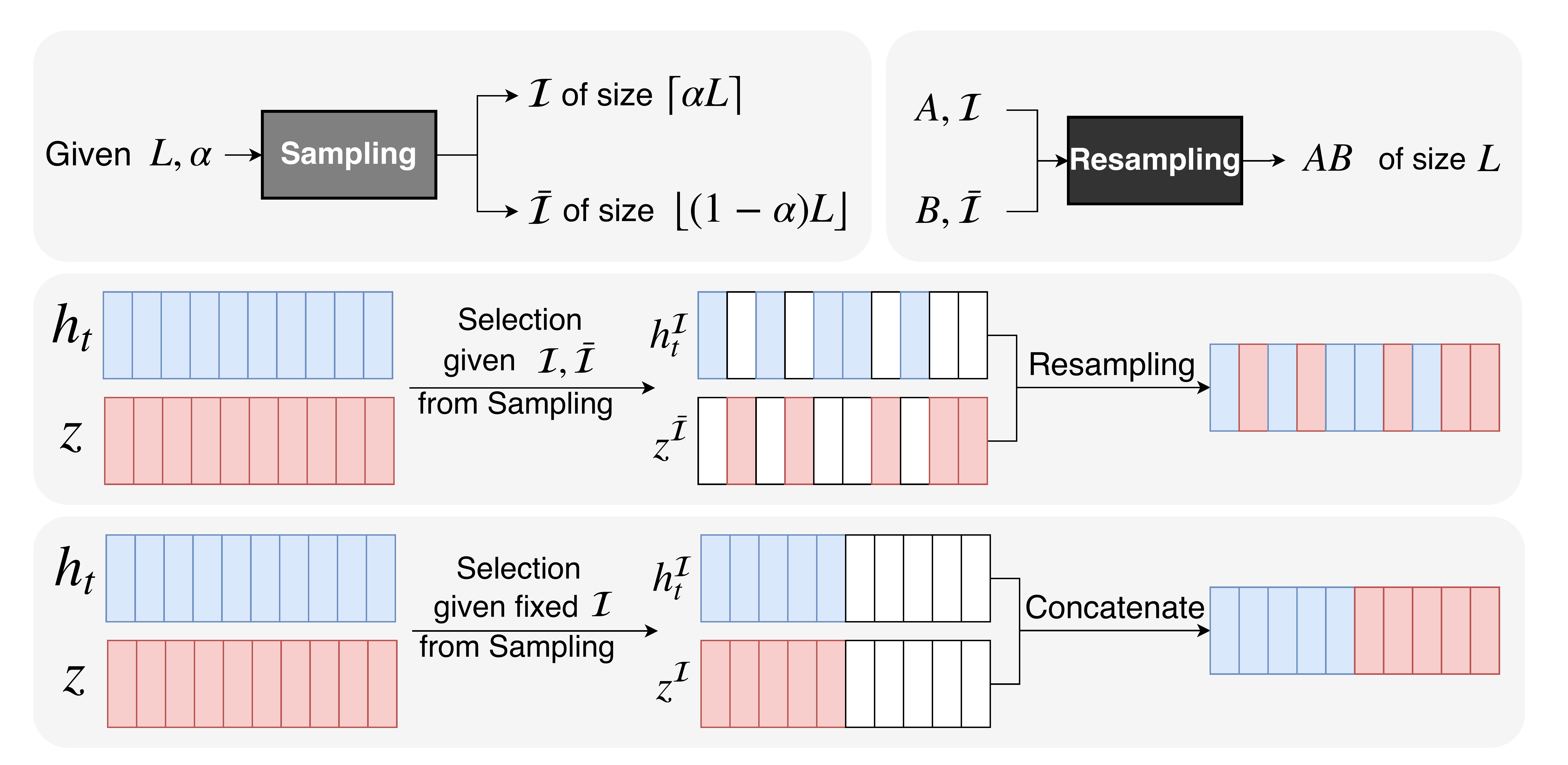}
    \caption{Mix-and-Match perturbation. \textbf{(Top)} Illustration of the \textit{Sampling} operation (left) and of the \textit{Resampling} one (right). Given a sampling rate $\alpha$ and a vector length $L$, the Sampling operation samples $\left\lceil\alpha L\right\rceil$ indices, say $\mathcal{I}$. The complementary, unsampled indices are denoted by $\mathcal{\bar{I}}$. Then, given two $L$-dimensional vectors and the corresponding  $\left\lceil\alpha L\right\rceil$ and $\left\lfloor(1-\alpha)L\right\rfloor$ indices, the Resampling operation mixes the two vectors to form a new $L$-dimensional one. \textbf{(Middle)} Example of Mix-and-Match perturbation. \textbf{(Bottom)} Example of perturbation by concatenation, as in~\citep{yan2018mt}. Note that, in Mix-and-Match perturbations, sampling is stochastic; the indices are sampled uniformly randomly for each mini-batch. By contrast, in~\citep{yan2018mt}, sampling is deterministic, and the indices in $\mathcal{I}$ are fixed and correspond to $\mathcal{I}=\{1,\dots,\frac{L}{2}\}$.
    }
    \vspace{-15pt}
    \label{fig:mix_match}
\end{figure}

\begin{figure*}[t]
    \centering
    \includegraphics[width=\textwidth]{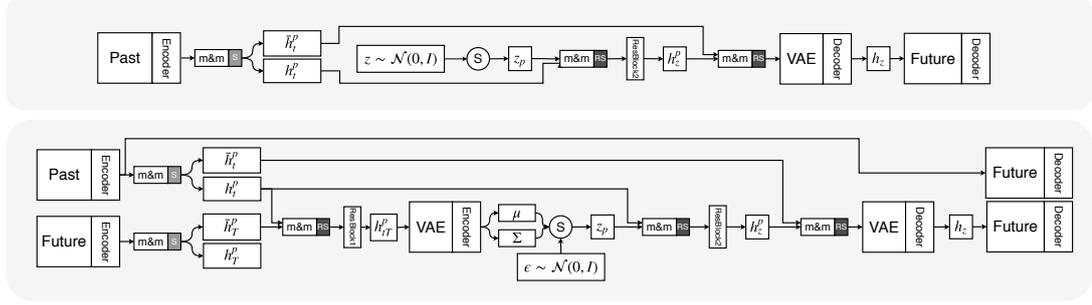}
    \caption{Overview of our approach. \textbf{(Top): Overview of the model during inference}. During inference, given past information and a random vector sampled from a Normal distribution, the model generates new motions.
    \textbf{(Bottom): Overview of the model during training.} During training, we use a future pose autoencoder with a CVAE  between the encoder and the decoder. The RNN encoder-decoder network mapping the past to the future then aims to generate good conditioning variables for the CVAE. }
    \label{fig:stochastic}
    \vspace{-10pt}
\end{figure*}

\subsection{Mix-and-Match Perturbation}
\label{sec:mix_match}

The main limitation of prior work in the area of stochastic motion modeling, such as~\citep{walker2017pose,barsoum2018hp,yan2018mt}, lies in the way they fuse the random vector with the conditioning variable, i.e., RNN hidden state or pose, which causes the model to learn to ignore the  randomness and solely exploit the deterministic conditioning information to generate motion. 
To overcome this, we propose to make it harder for the model to decouple the random variable from the deterministic information. Specifically, we observe that the way the random variable and the conditioning one are combined in existing methods is deterministic. We therefore propose to make this process stochastic.

Similarly to~\citep{yan2018mt}, we propose to make use of the hidden state as the conditioning variable and generate a perturbed hidden state by combining a part of the original hidden state with the random vector. However, as illustrated in Fig.~\ref{fig:mix_match}, instead of assigning predefined, deterministic indices to each piece of information, such as the first half for the hidden state and the second one for the random vector, we assign the values of the hidden state to \textit{random} indices and the random vector to the complementary ones. 

More specifically, as depicted in Fig.~\ref{fig:mix_match}, a mix-and-match perturbation takes two vectors of size $L$ as input, say $h_t$ and $z$, and combines them in a stochastic manner. To this end, it relies on two operations. The first one, called \textit{Sampling}, chooses $\left\lceil\alpha L\right\rceil$ indices uniformly at random among the $L$ possible values, given a sampling rate $0\leq \alpha \leq 1$. Let us denote by ${\cal I} \subseteq\{1,\ldots,L\}$, the resulting set of indices and by ${\cal \bar{I}}$ the complementary set. The second operation, called \textit{Resampling}, then creates a new $L$-dimensional vector whose values at indices in $\mathcal{I}$ are taken as those at corresponding indices in the first input vector and the others at the complementary indices, of dimension $\left\lfloor(1-\alpha)L\right\rfloor$, in the second input vector. 

\subsection{M\&M Perturbation for Motion Prediction}
Let us now describe the way we use our mix-and-match perturbation strategy for motion prediction. To this end, we first discuss the network we rely on during inference, and then explain our training strategy.

\noindent\textbf{Inference.} The high-level architecture we use at inference time is depicted by Fig.~\ref{fig:stochastic} (Top). It consists of an RNN encoder that takes $t$ poses $x_{1:t}$ as input and outputs an $L$-dimensional hidden vector $h_t$. A random $\left\lceil\alpha L\right\rceil$-dimensional portion of this hidden vector, $h_t^\mathcal{I}$, is then combined with an $\left\lfloor(1-\alpha)L\right\rfloor$-dimensional random vector $z\sim\mathcal{N}(0,I)$ via our mix-and-match perturbation strategy.
The resulting $L$-dimensional output is passed through a small neural network (i.e., \textit{ResBlock2} in Fig.~\ref{fig:stochastic}) that reduces its size to $\left\lceil\alpha L\right\rceil$, and then fused with the remaining $\left\lfloor(1-\alpha)L\right\rfloor$-dimensional portion of the hidden state, $h_t^\mathcal{\bar{I}}$. This, in turn, is passed through the VAE decoder to produce the final hidden state $h_z$, from which the future poses $x_{t+1:T}$ are obtained via the RNN decoder.

\noindent\textbf{Training.} During training, we aim to learn both the RNN parameters and the CVAE ones. Because the CVAE is an \emph{auto}encoder, it needs to take as input information about future poses. To this end, we complement our inference architecture with an additional RNN future encoder, yielding the training architecture depicted in Fig.~\ref{fig:stochastic} (Bottom). Note that, in this architecture, we incorporate an additional mix-and-match perturbation that fuses the hidden state of the RNN past encoder $h_t$ with that of the RNN future encoder $h_T$ and forms $h_{tT}^p$. This allows us to condition the VAE encoder in a manner similar to the decoder.
Note that, for each mini batch, we use the same set of sampled indices for all mix-and-match perturbation steps throughout the network. Furthermore, following the standard CVAE strategy, during training, the random vector $z_p$ is sampled from the approximate posterior distribution $\mathcal{N}(\mu_\theta(x), \Sigma_\theta(x))$, whose mean $\mu_\theta(x)$ and covariance matrix $\Sigma_\theta(x)$ are produced by the CVAE encoder with parameters $\theta$. 
This, in practice, is done by the reparameterization technique~\citep{kingma2013auto}.
Note that, during inference, $z_p=\epsilon\sim\mathcal{N}(0,I)$ since we do not have access to $x$, hence to $\mu_\theta(x)$ and $\Sigma_\theta(x)$.

To learn the parameters of our model, we rely on the availability of a dataset $D=\{X_1, X_2, ..., X_N\}$ containing $N$ videos $X_i$ depicting a human performing an action. Each video consists of a sequence of $T$ poses, $X_i=\{x_i^1, x_i^2, ..., x_i^T\}$, and each pose comprises $J$ joints forming a skeleton, $x_i^t=\{x_{i, 1}^t, x_{i, 2}^t, ..., x_{i, J}^t\}$. The pose of each joint is represented as a 4D quaternion. Given this data, we train our model by minimizing a loss function of the form

\begin{align}
\mathcal{L} = \frac{1}{N} \sum_{i=1}^N\Big(\mathcal{L}_{rot}(X_i) + \mathcal{L}_{skl}(X_i)\Big) + \lambda\mathcal{L}_{prior}\;.
\label{eq:stochastic_loss}
\end{align}

The first term in this loss compares the output of the network with the ground-truth motion using the squared loss. That is, 

\begin{align}
\mathcal{L}_{rot}(X_i) = -\sum_{k=t+1}^T\sum_{j=1}^{J}{\|\hat{x}^{k}_{i,j} - x^{k}_{i,j}\|}^2\;,
\label{eq:loss_quat}
\end{align}

where $\hat{x}^{k}_{i,j}$ is the predicted 4D quaternion for the $j^{th}$ joint at time $k$ in sample $i$, and $x^{k}_{i,j}$ the corresponding ground-truth one. The main weakness of this loss is that it treats all joints equally. However, when working with angles, some joints have a much larger influence on the pose than others. For example, because of the kinematic chain, the pose of the shoulder affects that of the rest of the arm, whereas the pose of the wrists has only a minor effect.
To take this into account, we define our second loss term as the error in 3D space. That is, 
\begin{align}
\mathcal{L}_{skl}(X_i) = -\sum_{k=t+1}^{T}\sum_{j=1}^{J} \|\hat{p}^{k}_{i,j} - p^{k}_{i,j}\|^2\;,
\label{eq:loss_skl}
\end{align}
where $\hat{p}^{k}_{i,j}$ is the predicted 3D position of joint $j$ at time $k$ in sample $i$ and $p^{k}_{i,j}$ the corresponding ground-truth one. These 3D positions can be computed using forward kinematics, as in~\citep{pavllo2018quaternet,pavllo2019modeling}. Note that, to compute this loss, we first perform a global alignment of the predicted pose and the ground-truth one by rotating the root joint to face [0, 0, 0].
Finally, following standard practice in training VAEs, we define our third loss term as the KL divergence 
\begin{align}
    \mathcal{L}_{prior} = -KL\Big(\mathcal{N}(\mu_\theta(x), \Sigma_\theta(x))) \| \mathcal{N}(0,I)\Big) \nonumber \\
    =-\frac{1}{2}\sum_{j=1}^d \Big(1+\log(\sigma_\theta(x)_{j}^2) - \mu_\theta(x)_j^2 - \sigma_\theta(x)_{c_j}^2 \Big)\;.
    \label{eq:kl}
\end{align}
where $\Sigma_\theta(x)=diag(\sigma_\theta(x)^2)$ and $d$ is the length of the diagonal of the covariance matrix. In practice, since our VAE appears within a recurrent model, we weigh $\mathcal{L}_{prior}$ by a function $\lambda$ corresponding to the KL annealing weight of~\citep{bowman2015generating}. We start from $\lambda=0$, forcing the model to encode as much information in $z$ as possible, and gradually increase it to $\lambda=1$, following a logistic curve. 

\begin{figure}
    \centering
    \includegraphics[width=\textwidth]{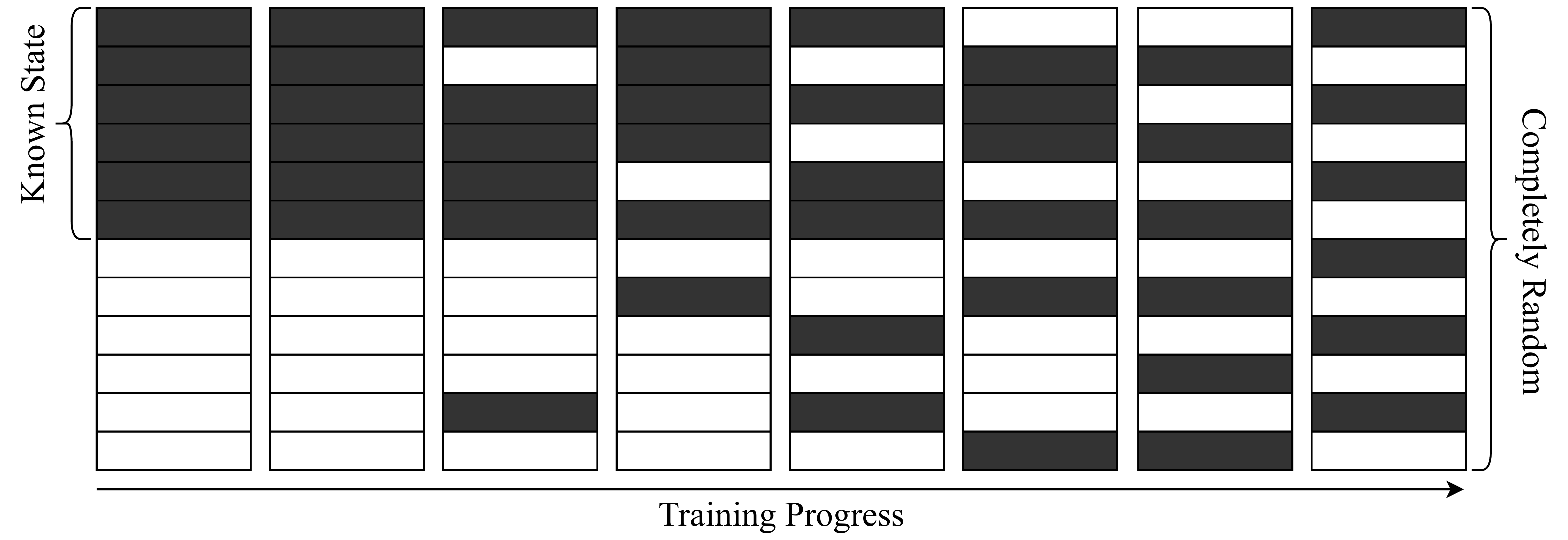}
    \caption{Example of curriculum perturbation of the hidden state. 
    }
    \label{fig:curriculum}
\end{figure}

\subsection{Curriculum Learning of Variation} 
\label{sec:curriculum}
The parameter $\alpha$ in our mix-and-match perturbation scheme determines a trade-off between stochasticity and motion quality. The larger $\alpha$, the larger the portion of the original hidden state that will be perturbed. Thus, the model incorporates more randomness and less information from the original hidden state. As such, given a large $\alpha$, it becomes harder for the model to deliver motion information from the observation to the future representation since a large portion of the hidden state is changing randomly. In particular, we observed that training becomes unstable if we use a large $\alpha$ from the beginning, with the motion-related loss terms fluctuating while the prior loss $\mathcal{L}_{prior}$ quickly converges to zero. 
To overcome this while still enabling the use of sufficiently large values of $\alpha$ to achieve high diversity, we introduce the curriculum learning strategy depicted by Fig.~\ref{fig:curriculum}. In essence, we initially select $\left\lceil\alpha L\right\rceil$ indices in a deterministic manner and gradually increase the randomness of these indices as training progresses. More specifically, given a set of $\left\lceil\alpha L\right\rceil$ indices, we replace $c$ indices from the sampled ones with the corresponding ones from the remaining $\left\lfloor(1-\alpha)L\right\rfloor$ indices. Starting from $c=0$, we gradually increase $c$ to the point where all $\left\lceil\alpha L\right\rceil$ indices are sampled uniformly randomly. More details, including the pseudo-code of this approach, are provided in the supplementary material. This strategy helps the motion decoder to initially learn and incorporate information about the observations (as in~\citep{yan2018mt}), yet, in the long run, still prevents it from ignoring the random vector.

\subsection{Quality and Diversity Metrics}
\label{sec:eval}
When dealing with multiple plausible motions, or in general diverse solutions to a problem, evaluation is a challenge. The standard metrics used for deterministic motion prediction models are ill-suited to this task, because they typically compare the predictions to the ground truth, thus inherently penalizing diversity. For multiple motions, two aspects are important: the \emph{diversity} and the \emph{quality}, or realism, of each individual motion.
Prior work typically evaluates these aspects via human judgement. While human evaluation is highly valuable, and we will also report human results, it is very costly and time-consuming. Here, we therefore introduce two metrics that facilitate the quantitative evaluation of both quality and diversity of generated human motions. We additionally extend the Inception-Score~\citep{salimans2016improved} to our task.

To measure the quality of generated motions, we propose to rely on a binary classifier trained to discriminate real (ground-truth) samples from fake (generated) ones. The accuracy of this classifier on the test set is thus inversely proportional to the quality of the generated motions. In other words, high-quality motions are those that are not distinguishable from real ones. Note that we do not rely on adversarial training, i.e., we do not define a loss based on this classifier when training our model.
To measure the diversity of the generated motions, a naive approach would consist of relying on the distance between the generated motion and a reference one. However, generating identical motions that are all far from the reference one would therefore yield a high value, while not reflecting diversity. To prevent this, we propose to make use of the average distance between all pairs of generated motions. A similar idea has been investigated to measure the diversity of solutions in other domains~\citep{yuan2019diverse,yang2018diversitysensitive}.

The quality and diversity metrics can reliably evaluate a stochastic motion prediction model. While providing valuable information, drawing conclusion about the performance of a model is always easier with a single measure. To this end, we extend the Inception-Score (IS)~\citep{salimans2016improved} used to measure the quality of images produced by a generative model. Our extension to IS is twofold: \textbf{(1)} Inspired by~\citep{huang2018multimodal}, we extend IS to the conditional case, where the condition provides the core signal to generate the sample; \textbf{(2)} Our extended IS measures the quality and diversity of \emph{sequential} solutions. To this end, we first train a strong skeleton-based action classifier~\citep{li2018co} on ground-truth motions. With then compute the IS of each of the multiple motions generated for a given condition (observed motion), and report the mean IS and its standard deviation over all conditions. The reason behind reporting the mean IS over all conditions is to evaluating the diversity of generated motions given each observation. 
Note that studying IS only makes it hard to evaluate quality and diversity separately, and thus we still believe that all three metrics are required. Importantly, we show empirically that our proposed metrics are in line with human judgement, at considerably lower cost.

\begin{table*}[!h]
    \centering
    \footnotesize
    \begin{tabular}{c}
    \begin{tabular}{l c c c c c c}
    \multicolumn{5}{c}{\bf Quantitative results on Human3.6M dataset}\\
    \hline
    Method & ELBO $\downarrow$ (KL $\uparrow$) & Diversity $\uparrow$ & Quality $\uparrow$ & IS $\uparrow$ & Tr KL $\uparrow$ \\
    \hline
    \citep{yan2018mt} & 0.51 (0.06) & 0.26 & 0.45 & 1.9$\pm$0.4  & 0.08 \\
    \citep{walker2017pose} & 2.08 (N/A) & 1.70 & 0.13  & 1.8$\pm$0.6 & N/A \\
    \citep{barsoum2018hp} & 0.61 (N/A) & 0.48 & 0.47 & 2.1$\pm$1.3 & N/A\\
    \hline
    \texttt{Mix-and-Match} & 0.55 (2.03)  &  3.52 & 0.42 & 7.3$\pm$1.4 & 1.98\\
    \hline
    \end{tabular}
    \\
    \begin{tabular}{l c c c c c c}
    \multicolumn{5}{c}{\bf Quantitative results on CMU Mocap dataset}\\
    \hline
    Method & ELBO $\downarrow$ (KL $\uparrow$) & Diversity $\uparrow$ & Quality $\uparrow$ & IS $\uparrow$ & Tr KL $\uparrow$ \\
    \hline
    \citep{yan2018mt} & 0.25 (0.08) & 0.41 & 0.46 &  2.4$\pm$0.1 & 0.01 \\
    \citep{walker2017pose} & 1.93 (N/A) & 3.00 & 0.18 & 1.4$\pm$0.4 & N/A \\
    \citep{barsoum2018hp} & 0.24 (N/A) & 0.43 & 0.45 & 2.0$\pm$1.0 & N/A\\
    \hline
    \texttt{Mix-and-Match} & 0.25 (2.92)  &  2.63 & 0.46 & 9.0$\pm$1.7 & 2.20\\
    \hline
    
    \end{tabular}
    \end{tabular}
    \caption{Comparison of our approach with the stochastic motion prediction baselines on Human3.6M dataset (left) and CMU Mocap dataset (right). Tr KL stands for KL term at training convergence.}
    \label{tab:stoch_h36m}
\end{table*}

\begin{figure*}[!h]
    \centering
    \includegraphics[width=0.97\textwidth]{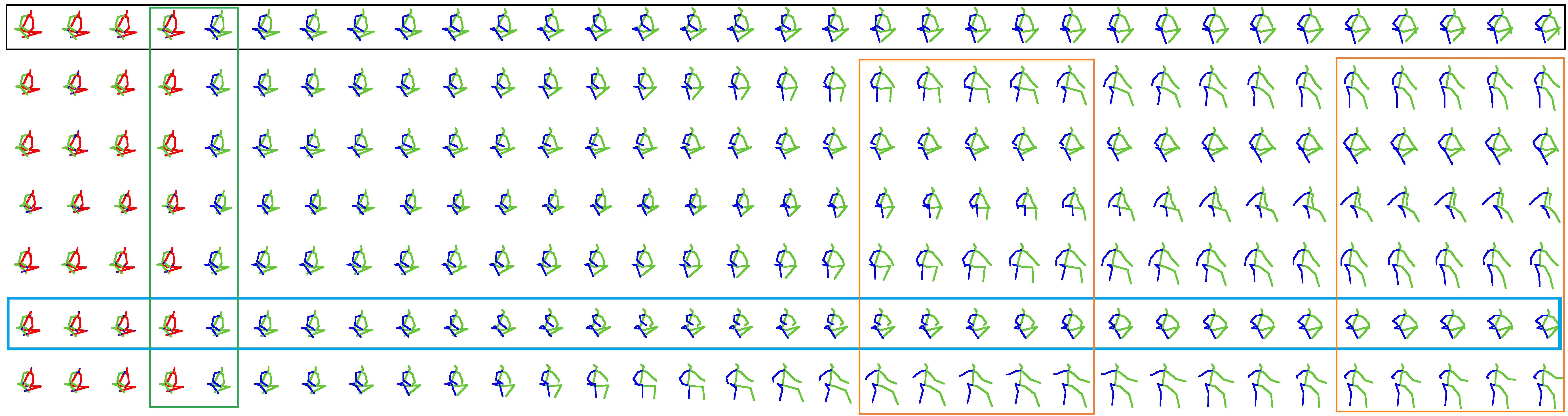}
    \caption{Qualitative evaluation of diversity. The first row (black box) shows the ground-truth motion. The next six rows depict six randomly generated motions (not cherry-picked) given the same observations (the first four poses of each motion). The green box shows the last observed frame and the first generated one, illustrating the consistency of the generated motions. The orange boxes show the diversity of the generated motions in different temporal windows. The blue box shows a randomly sampled motion whose poses are similar to the ground-truth ones. Best seen in color and zoomed in.}
    \label{fig:h36m_div}
\end{figure*}{}

\begin{figure}[!h]
    \centering
    \small
        \includegraphics[width=\textwidth]{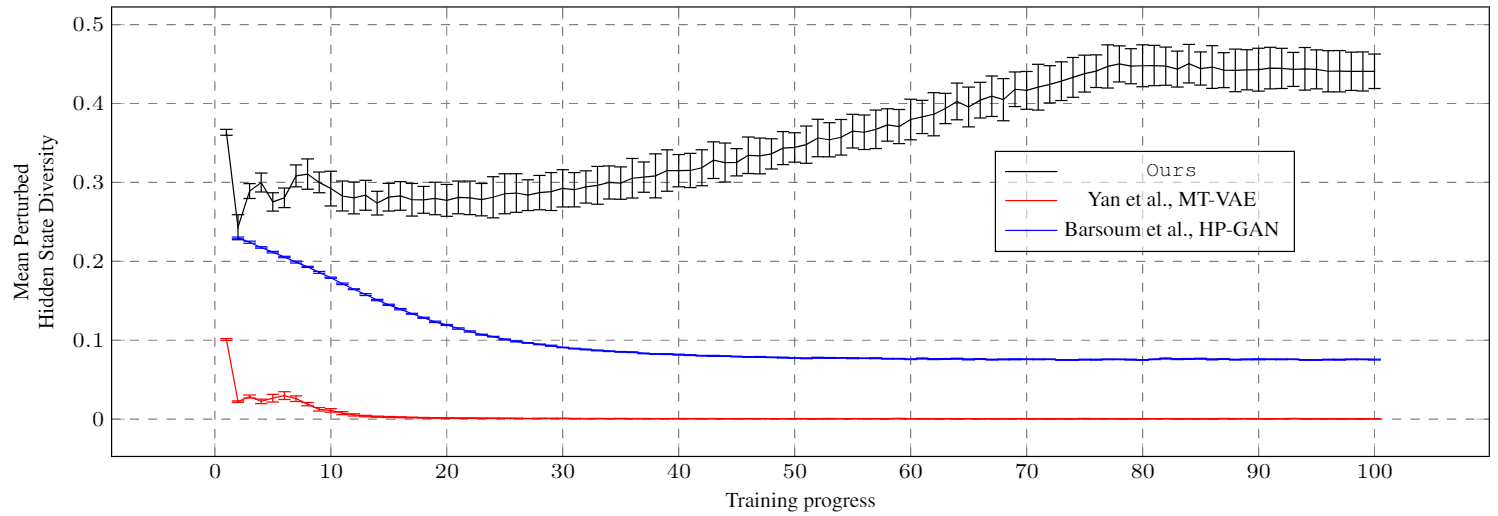} 
    \caption{Diversity of $K$ RNN decoder inputs,
    generated with $K=50$ different random vectors. We report the mean diversity over $N=50$ samples and the corresponding standard deviation. 
    }
    \label{fig:baseline_RHP}
\end{figure}

\begin{figure*}[!ht]
\scriptsize
    \centering
    \tabcolsep=0.05cm
    \renewcommand{\arraystretch}{0.6} 
    \begin{tabular}{c}
         \includegraphics[width=\textwidth]{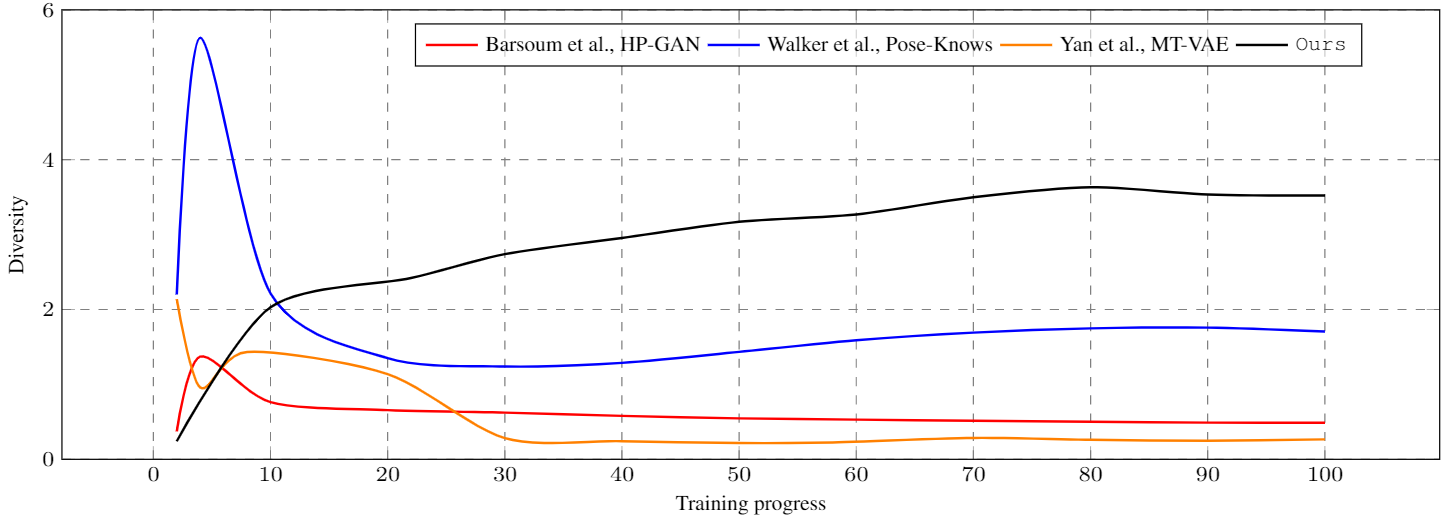}  \\
         \includegraphics[width=\textwidth]{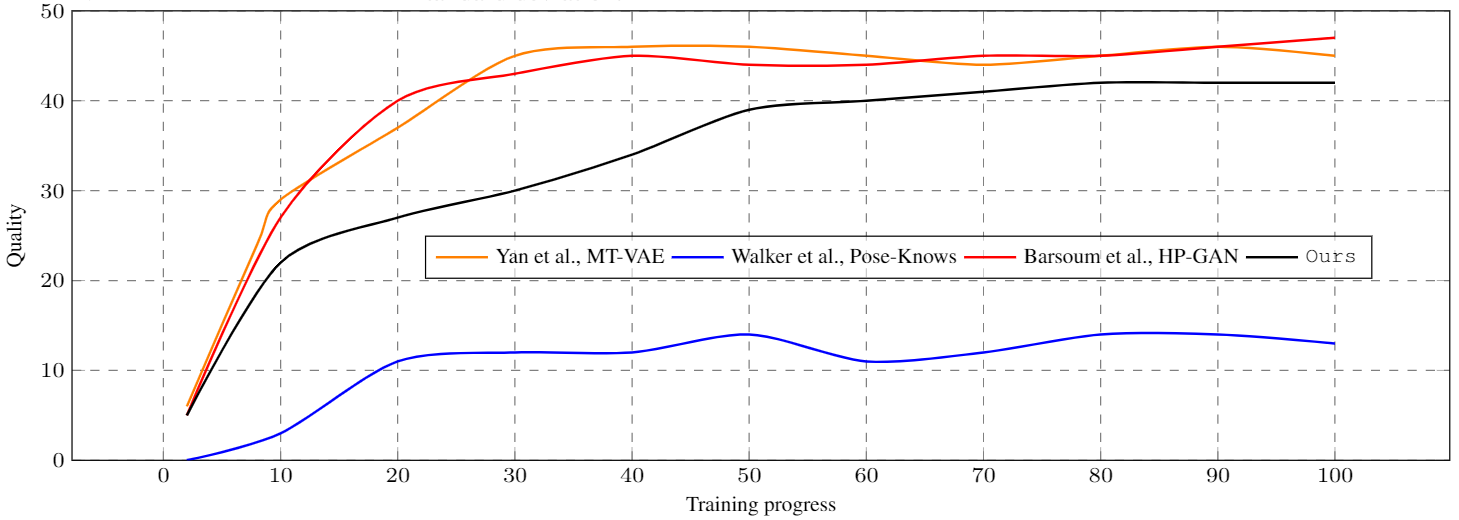}  \\
         \includegraphics[width=\textwidth]{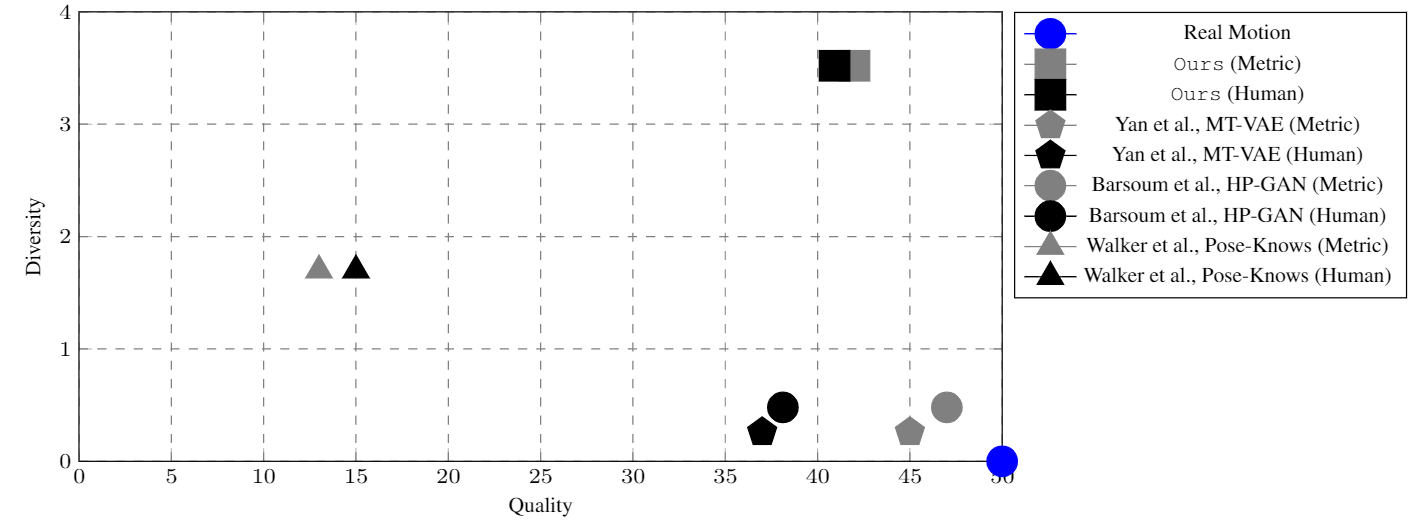} 
    \end{tabular}{}
    \caption{(Top) Diversity of our approach and the stochastic baselines. (Middle) Quality of our approach and the stochastic baselines. (Bottom) Comparing classifier-based and human evaluation of quality for our approach and the baselines, where the statistics correspond to evaluation after the models are fully trained. The numbers are provided in the supplementary material to facilitate future comparisons.
    }
    \label{fig:QD}
\end{figure*}

\section{Experiments}

We now evaluate the effectiveness of our approach at generating multiple plausible motions. To this end, we use Human3.6M~\citep{h36m_pami} and the CMU Mocap dataset\footnote{Available at \texttt{http://mocap.cs.cmu.edu/}.}, two large publicly available motion capture datasets. In this section, we introduce the baselines and give information about the implementation details and evaluation metrics. We then provide all the experimental results.

\paragraph{Baselines.} We compare our Mix-and-Match approach with the different means of imposing variation in motion prediction discussed in Section~\ref{sec:related_work}, i.e., concatenating the hidden state to a learned latent variable, Yan et al.,~\citep{yan2018mt}, concatenating the pose to a learned latent variable at each time-step, Walker et al.,~\citep{walker2017pose}, and adding a (transformed) random noise to the hidden state, Barsoum et al.,~\citep{barsoum2018hp}. For the comparison to be fair, we use 16 frames (i.e., 640ms) as observation to generate the next 60 frames (i.e., 2.4sec) for all baselines. All models are trained with the same motion representation, annealing strategy, backbone network, and  losses, except for Barsoum et al.,~\citep{barsoum2018hp} which cannot make use of $\mathcal{L}_{prior}$.

\paragraph{Implementation Details.} The motion encoders and decoders in our model are single layer GRU~\citep{cho2014learning} networks, comprising 1024 hidden units each. For the decoders, we use a teacher forcing technique~\citep{williams1989learning} to decode motion. At each time-step, the network chooses with probability $P_{tf}$ whether to use its own output at the previous time-step or the ground-truth pose as input. We initialize $P_{tf}=1$, and decrease it linearly at each training epoch such that, after a certain number of epochs, the model becomes completely autoregressive, i.e., uses only its own output as input to the next time-step. 
We train our model on a single GPU with the Adam optimizer~\citep{kingma2014adam} for 100K iterations. We use a learning rate of 0.001 and a mini-batch size of 64. To avoid exploding gradients, we use the gradient-clipping technique of~\citep{pascanu2013difficulty} for all layers in the network. We implemented our model using the Pytorch framework of~\citep{paszke2017automatic}.

\paragraph{Evaluation Metrics.} 
In addition to the metrics discussed in Section~\ref{sec:eval}, we also report the standard ELBO metric (approximated by the reconstruction loss and the KL on the test set) and the sampling loss (S-MSE) of our approach and the state-of-the-art stochastic motion prediction techniques. However, evaluating only against one ground-truth motion (i.e., one sample from multi-modal distribution), as in MSE or S-MSE, can lead to a high score for one sample while penalizing other plausible modes. This behavior is undesirable since it cannot differentiate a multi-modal solution from a good, but uni-modal one. 
Similarly, the metrics in~\citep{yan2018mt} or the approximate ELBO only evaluate quality given one single ground truth. 
While the ground truth has high quality, there exist multiple high quality continuations of an observation, which our proposed metric accounts for.
As discussed in Section~\ref{sec:eval}, we evaluate the quality and diversity of the predicted motions. Note, these metrics should be considered together, since each one taken separately does not provide a complete picture of how well a model can predict \emph{multiple plausible} future motions. For example, a model can generate diverse but unnatural motions, or, conversely, realistic but identical motions.
To evaluate quality, as discussed in Section~\ref{sec:eval}, we use a recurrent binary classifier whose task is to determine whether a sample comes from the ground-truth data or was generated by the model.
We train such a classifier for each method, using 25K samples generated at different training steps together with 25K real samples, forming a binary dataset of 50K motions for each method. 
To evaluate diversity, as discussed in Section~\ref{sec:eval}, we compute the mean Euclidean distance from each motion to all other $K-1$ motions when generating $K=50$ motions. 
To compute IS, we trained an action classifier~\citep{li2018co} with 50K real motions. We then compute the IS for $K=50$ samples per condition for 50 different conditions. We followed Section~\ref{sec:eval} to report IS.
Furthermore, we also performed a human evaluation to measure the quality of the motions generated by each method. To this end, we asked eight users to rate the quality of 50 motions generated by each method, for a total of 200 motions. The ratings were defined on a scale of 1-5, 1 representing a low-quality motion and 5 a high-quality, realistic one. We then scaled the values to the range 0-50 to make them comparable with those of the binary classifier.

\subsection{Comparison to the State-of-the-Art}
In this section, we quantitatively compare our approach to the state-of-the-art stochastic motion prediction techniques in terms of approximate ELBO, Diversity, Quality, and IS on a held-out test set, as well as the training KL term at convergence. Table~\ref{tab:stoch_h36m} shows the results on the Human3.6M and CMU Mocap datasets.

These results show that Mix-and-Match is highly capable of learning the variation in human motion while maintaining a good motion quality. This is shown by IS, Diversity, and Quality metrics, which should be considered together. It is also evidenced by the low reconstruction loss and higher KL term on the test set. The training KL term at  convergence also shows that, in Mix-and-Match, the posterior does not collapse to the prior distribution, i.e., the model does not ignore the latent variable. While the MSE of our approach is slightly higher than that of Yan et al.,~\citep{yan2018mt} on Human3.6M and Barsoum et al.,~\citep{barsoum2018hp} on the CMU Mocap dataset,  we effectively exploit the latent variables, as demonstrated by the KL term on the test set, the IS and diversity metric and the qualitative results provided in Fig.~\ref{fig:h36m_div} and in the supplementary material. As evidenced by the examples of diverse motions generated by our model in Fig.~\ref{fig:h36m_div}, given a single observation, Mix-and-Match is able to generate diverse, but natural motions\footnote{See the video of our results in the supplementary material.}.

\subsection{Analysis on Diversity and Quality}
\label{sec:ablation}
To provide a deeper understanding of our approach, we evaluate different aspects of Mix-and-Match. All these experiments were done on Human3.6M. In the following, we first analyze the diversity in the hidden state space, i.e., the first part of the model where variation is imposed. We then evaluate the quality and diversity of prediction when tested at different stages of the training. We also perform a human evaluation on the quality of the generated motions, comparing it with our inexpensive, automatic quality metric. Finally, we compare Mix-and-Match with other stochastic techniques in terms of sampling error (S-MSE), i.e., by computing the error of the best of $K$ generated motions given the ground-truth one. More experiments and visualizations are provided in the supplementary material.

\paragraph{Diversity in Hidden State Space.}
In Fig.~\ref{fig:baseline_RHP}, we plot the diversity of the representations used as input to the RNN decoders of~\citep{yan2018mt} and~\citep{barsoum2018hp}, two state-of-the-art methods that are closest in spirit to our approach. Here, diversity is measured as the average pairwise distance across the $K=50$ representations produced for a single series of observations. We report the mean diversity over 50 samples and the corresponding standard deviation. As can be seen from the figure, the diversity of~\citep{yan2018mt} and~\citep{barsoum2018hp} decreases as training progresses, thus supporting our observation that these models learn to ignore the 
perturbations. As evidenced by the black curve, which shows an increasing diversity as training progresses, our approach produces not only high-quality predictions but also truly diverse ones. The gradual but steady increase in diversity of our approach is due to our curriculum learning strategy described in Section~\ref{sec:curriculum}. Without it, training is less stable, with large diversity variations.

\begin{table*}[ht]
    \centering
    \scalebox{0.97}{
    \begin{tabular}{c}
        \begin{tabular}{l c c c c c c}
    & \multicolumn{6}{c}{Walking}\\
    \hline
    Method & 80ms  & 160ms & 320ms & 400ms & 560ms & 1000ms \\
        \hline
    MT-VAE~\citep{yan2018mt} & 
    0.73 & 0.79 & 0.90 & 0.93 & 0.95 & 1.05 \\
    HP-GAN~\citep{barsoum2018hp} & 
    0.61 & 0.62 & 0.71 & 0.79 & 0.83 & 1.07 \\
    Pose-knows~\citep{walker2017pose} & 
    0.56 & 0.66 & 0.98 & 1.05 & 1.28 & 1.60 \\
    \hline
    \texttt{Mix-and-Match} & 
    0.33 & 0.48 & 0.56 & 0.58 & 0.64 & 0.68 \\
    \hline
    \end{tabular}
    \\
    \begin{tabular}{l c c c c c c}
    \hline
    & \multicolumn{6}{c}{Eating}\\
    \hline
    Method & 80ms  & 160ms & 320ms & 400ms & 560ms & 1000ms \\
    \hline
    MT-VAE~\citep{yan2018mt} & 
    0.68 & 0.74 & 0.95 & 1.00 & 1.03 & 1.38 \\
    HP-GAN~\citep{barsoum2018hp} & 
    0.53 & 0.67 & 0.79 & 0.88 & 0.97 & 1.12 \\
    Pose-knows~\citep{walker2017pose} & 
    0.44 & 0.60 & 0.71 & 0.84 & 1.05 & 1.54 \\
    \hline
    \texttt{Mix-and-Match} & 
    0.23 & 0.34 & 0.41 & 0.50 & 0.61 & 0.91 \\
    \hline
    \end{tabular}
    \\
    \begin{tabular}{l c c c c c c}
    \hline
    & \multicolumn{6}{c}{Smoking}\\
    \hline
    Method & 80ms  & 160ms & 320ms & 400ms & 560ms & 1000ms \\
    \hline
    MT-VAE~\citep{yan2018mt} & 
    1.00 & 1.14 & 1.43 & 1.44 & 1.68 & 1.99 \\
    HP-GAN~\citep{barsoum2018hp} & 
    0.64 & 0.78 & 1.05 & 1.12 & 1.64 & 1.84 \\
    Pose-knows~\citep{walker2017pose} & 
    0.59 & 0.83 & 1.25 & 1.36 & 1.67 & 2.03 \\
    \hline
    \texttt{Mix-and-Match} & 
    0.23 & 0.42 & 0.79 & 0.77 & 0.82 & 1.25 \\
\hline
    \end{tabular} 
    \\
    \begin{tabular}{l c c c c c c}
    \hline
    & \multicolumn{6}{c}{Discussion}\\
    \hline
  Method & 80ms  & 160ms & 320ms & 400ms & 560ms & 1000ms \\
    \hline
    MT-VAE~\citep{yan2018mt} & 
    0.80 & 1.01 & 1.22 & 1.35 & 1.56 & 1.69 \\
    HP-GAN~\citep{barsoum2018hp} & 
    0.79 & 1.00 & 1.12 & 1.29 & 1.43 & 1.71 \\
    Pose-knows~\citep{walker2017pose} & 
    0.73 & 1.10 & 1.33 & 1.34 & 1.45 & 1.85 \\
    \hline
    \texttt{Mix-and-Match} & 
    0.25 & 0.60 & 0.83 & 0.89 & 1.12 & 1.30 \\
\hline
    \end{tabular}
    
    \end{tabular}
    }
    
    \caption{Quantitative comparison of the S-MSE against stochastic baselines for four actions of the Human3.6M dataset.}
    \label{tab:det}
\end{table*}

\begin{table*}
    \centering
    \scalebox{0.97}{
    \begin{tabular}{c}
        \begin{tabular}{l c c c c c c}
    & \multicolumn{6}{c}{Walking}\\
    \hline
    Method & 80ms  & 160ms & 320ms & 400ms & 560ms & 1000ms \\
    \hline
    Zero Velocity & 
    0.39 & 0.86 & 0.99 & 1.15 & 1.35 & 1.32 \\
    AGED~\citep{gui2018adversarial} & 
    0.22 & 0.36 & 0.55 & 0.67 & 0.78 & 0.91 \\
    Imitation~\citep{wang2019imitation} &
    0.21 & 0.34 & 0.53 & 0.59 & 0.67 & 0.69 \\
    LSTM-3LR~\citep{fragkiadaki2015recurrent} & 
    1.18 & 1.50 & 1.67 & 1.76 & 1.81 & 2.20 \\
    SRNN~\citep{jain2016structural} & 
    1.08 & 1.34 & 1.60 & 1.80 & 1.90 & 2.13 \\
    DAE-LSTM~\citep{ghosh2017learning} & 
    1.00 & 1.11 & 1.39 & 1.48 & 1.55 & 1.39 \\
    GRU~\citep{martinez2017human} & 
    0.28 & 0.49 & 0.72 & 0.81 & 0.93 & 1.03 \\
    LTD~\citep{mao2019learning} & 
    0.18 & 0.31 & 0.49 & 0.56 & 0.65 & 0.67 \\
    \hline
    Mix-and-Match & 
    0.33 & 0.48 & 0.56 & 0.58 & 0.64 & 0.68 \\
    \hline
    \end{tabular} 
    \\
    \begin{tabular}{l c c c c c c}
    \hline
    & \multicolumn{6}{c}{Eating}\\
    \hline
    Method & 80ms  & 160ms & 320ms & 400ms & 560ms & 1000ms \\
    \hline
    Zero Velocity & 
    0.27 & 0.48 & 0.73 & 0.86 & 1.04 & 1.38 \\
    AGED~\citep{gui2018adversarial} & 
    0.17 & 0.28 & 0.51 & 0.64 & 0.86 & 0.93 \\
    Imitation~\citep{wang2019imitation} &
    0.17 & 0.30 & 0.52 & 0.65 & 0.79 & 1.13 \\
    LSTM-3LR~\citep{fragkiadaki2015recurrent} & 
    1.36 & 1.79 & 2.29 & 2.42 & 2.49 & 2.82 \\
    SRNN~\citep{jain2016structural} & 
    1.35 & 1.71 & 2.12 & 2.21 & 2.28 & 2.58 \\
    DAE-LSTM~\citep{ghosh2017learning} & 
    1.31 & 1.49 & 1.86 & 1.89 & 1.76 & 2.01 \\
    GRU~\citep{martinez2017human} & 
    0.23 & 0.39 & 0.62 & 0.76 & 0.95 & 1.08 \\
    LTD~\citep{mao2019learning} &
    0.16 & 0.29 & 0.50 & 0.62 & 0.76 & 1.12 \\
    \hline

    Mix-and-Match & 
    0.23 & 0.34 & 0.41 & 0.50 & 0.61 & 0.91 \\
    \hline
    \end{tabular}
    \\
    \begin{tabular}{l c c c c c c}
    \hline
    & \multicolumn{6}{c}{Smoking}\\
    \hline
    Method & 80ms  & 160ms & 320ms & 400ms & 560ms & 1000ms \\
    \hline
    Zero Velocity & 
    0.26 & 0.48 & 0.97 & 0.95 & 1.02 & 1.69 \\
    AGED~\citep{gui2018adversarial} & 
    0.27 & 0.43 & 0.82 & 0.84 & 1.06 & 1.21 \\
    Imitation~\citep{wang2019imitation} &
    0.23 & 0.44 & 0.86 & 0.85 & 0.95 & 1.63 \\
    LSTM-3LR~\citep{fragkiadaki2015recurrent} & 
    2.05 & 2.34 & 3.10 & 3.18 & 3.24 & 3.42 \\
    SRNN~\citep{jain2016structural} & 
    1.90 & 2.30 & 2.90 & 3.10 & 3.21 & 3.23 \\
    DAE-LSTM~\citep{ghosh2017learning} & 
    0.92 & 1.03 & 1.15 & 1.25 & 1.38 & 1.77 \\
    GRU~\citep{martinez2017human} & 
    0.33 & 0.61 & 1.05 & 1.15 & 1.25 & 1.50 \\
    LTD~\citep{mao2019learning} &
    0.22 & 0.41 & 0.86 & 0.80 & 0.87 &  1.57 \\
    \hline

    Mix-and-Match & 
    0.23 & 0.42 & 0.79 & 0.77 & 0.82 & 1.25 \\
\hline
    \end{tabular}   
    \\
    \begin{tabular}{l c c c c c c}
    \hline
    & \multicolumn{6}{c}{Discussion}\\
    \hline
   Method & 80ms  & 160ms & 320ms & 400ms & 560ms & 1000ms \\
    \hline
    Zero Velocity & 
    0.31 & 0.67 & 0.94 & 1.04 & 1.41 & 1.96 \\
    AGED~\citep{gui2018adversarial} & 
    0.27 & 0.56 & 0.76 & 0.83 & 1.25 & 1.30 \\
    Imitation~\citep{wang2019imitation} &
    0.27 & 0.56 & 0.82 & 0.91 & 1.34 & 1.81 \\
    LSTM-3LR~\citep{fragkiadaki2015recurrent} & 
    2.25 & 2.33 & 2.45 & 2.46 & 2.48 & 2.93 \\
    SRNN~\citep{jain2016structural} & 
    1.67 & 2.03 & 2.20 & 2.31 & 2.39 & 2.43 \\
    DAE-LSTM~\citep{ghosh2017learning} & 
    1.11 & 1.20 & 1.38 & 1.42 & 1.53 & 1.73 \\
    GRU~\citep{martinez2017human} & 
    0.31 & 0.68 & 1.01 & 1.09 & 1.43 & 1.69 \\
    LTD~\citep{mao2019learning} &
    0.20 & 0.51 & 0.77 & 0.85 & 1.33 & 1.70 \\
    \hline

    Mix-and-Match & 
    0.25 & 0.60 & 0.83 & 0.89 & 1.12 & 1.30 \\
\hline
    \end{tabular}
    
    \end{tabular}
    }
    \caption{Comparison against deterministic motion prediction techniques for four actions of the Human3.6M dataset.}
    \label{tab:det_baseline}
\end{table*}

\paragraph{Diversity and Quality in Motion Space.}
Now, we thoroughly compare our approach with state-of-the-art stochastic motion prediction models in terms of quality and diversity.
The results of the metrics of Section~\ref{sec:eval} are provided in  Fig.~\ref{fig:QD}(Left and Middle) and those of the human evaluation in Fig.~\ref{fig:QD}(Right). Below, we analyze the results of the different models.

As can be seen from Fig.~\ref{fig:QD},~\citep{yan2018mt} tends to ignore the random variable $z$, thus ignoring the root of variation. As a consequence, it achieves a low diversity, much lower than ours, but produces samples of high quality, albeit almost identical, which is also shown in qualitatively in Fig. 3 of the supplementary material. We empirically observed that the magnitude of the weights acting on $z$ to be orders of magnitude  smaller than that of acting on the condition, 0.008 versus 232.85 respectively.
Note that this decrease in diversity occurs after 16K iterations, indicating that the model takes time to identify the part of the hidden state that contains the randomness. Nevertheless, at iteration 16K, prediction quality is low, and thus one could not simply stop training at this stage. 
Note that the lack of diversity of~\citep{yan2018mt} is also evidenced by Fig.~\ref{fig:baseline_RHP}. 
As can be verified in Fig.~\ref{fig:QD}(Right), where~\citep{yan2018mt} appears in a region of high quality but low diversity, the results of human evaluation match those of our classifier-based quality metric.

Fig.~\ref{fig:QD} also evidences the limited diversity of the motions produced by~\citep{barsoum2018hp} despite its use of random noise during inference. Note that the authors of~\citep{barsoum2018hp} mentioned in their chapter that the random noise was added to the hidden state. Only by studying their publicly available code\footnote{\texttt{https://github.com/ebarsoum/hpgan}} did we understand the precise way this combination was done. In fact, the addition relies on a parametric, linear transformation of the noise vector. That is, the perturbed hidden state is obtained as $h_{perturbed} = h_{original} + W^{z\rightarrow h} z$.
Because the parameters $W^{z\rightarrow h}$ are \emph{learned}, the model has the flexibility to ignore $z$ (the magnitude of $W^{z\rightarrow h}$ is in the order of $O(1e^{-3})$), which causes the behavior observed in Figs.~\ref{fig:QD} and~\ref{fig:baseline_RHP}. Note that the authors of~\citep{barsoum2018hp} acknowledged that, despite their best efforts, they noticed very little variations between predictions obtained with different $z$ values. 
By depicting~\citep{barsoum2018hp} in a region of high quality but low diversity, the human evaluation results in Fig.~\ref{fig:QD}(Right) again match those of our classifier-based quality metric. 

As can be seen in Fig.~\ref{fig:QD}(Left and Middle),~\citep{walker2017pose} produces motions with higher diversity than~\citep{barsoum2018hp,yan2018mt}, but of much lower quality. The main reason behind this is that the random vectors that are concatenated to the poses at each time-step are sampled independently of each other, which translates to discontinuities in the generated motions. 
Human evaluation in Fig.\ref{fig:QD}(Right) further confirms that~\citep{walker2017pose}'s results lie in a low-quality, medium-diversity region.

The success of our approach is confirmed by Fig.~\ref{fig:QD}(Left and Middle). Our model generates diverse motions, even after a long training time, and the quality of these motions is high. While this quality is slightly lower than that of~\citep{barsoum2018hp,yan2018mt} when looking at our classifier-based metric, it is rated higher by IS and humans, as can be verified from Fig.~\ref{fig:QD}(Right) and Table~\ref{tab:stoch_h36m}. 
Altogether, these results confirm the ability of our approach to generate highly diverse yet realistic motions. 

\paragraph{Evaluating the Sampling Error.}
We now quantitatively compare our approach with other stochastic baselines in terms of sampling error (aka S-MSE). To this end, we follow the evaluation setting of deterministic motion prediction (as in ~\citep{fragkiadaki2015recurrent,pavllo2019modeling,pavllo2018quaternet,martinez2017human,gui2018adversarial}) which allows further comparisons to deterministic baselines.
We report the standard metric, i.e., the Euclidean distance between the generated and ground-truth Euler angles (aka MAE). To evaluate this metric for our method and the stochastic motion prediction models, which generate multiple, diverse predictions, we make use of the best sample among the $K$ generated ones with $K=50$ for the stochastic baselines and for our approach. This evaluation procedure aims to show that, among the $K$ generated motions, at least one is close to the ground truth. As shown in Table~\ref{tab:det}, by providing higher diversity, our approach outperforms the baselines. 
Similarly, in Table~\ref{tab:det_baseline}, we compare the best of $K=50$ sampled motions for our approach with the deterministic motion prediction techniques. Note that the goal of this experiment is not to provide a fair comparison to deterministic models, but to show that, among the diverse set of motions generated by our model, there exists at least one motion that is very close to the ground-truth one. The point of bringing the MAE of other deterministic methods, is to show how good deterministic models, with sophisticated architectures and complicated loss functions, perform on this task.

 \section{Conclusion}

In this chapter, we have proposed an effective way of perturbing the hidden state of an RNN such that it becomes capable of learning the multiple modes of human motions. Our evaluation of quality and diversity, based on both new quantitative metrics and human judgment, have evidenced that our approach outperforms existing stochastic methods. 
Generating diverse plausible motions given limited observations has many applications, especially when the motions are generated in an action-agnostic manner, as done here. For instance, our model can be used for human action forecasting~\citep{rodriguez2018action,aliakbarian2016deep,shi2018action}, as well as the methods introduced in Chapter 3 and Chapter 4, where one seeks to anticipate the action as early as possible, or for motion inpainting, where, given partial observations, one aims to generate multiple in-between solutions.
Although Mix-and-Match can successfully encourage diversity in the generate motions, it does not account for the fact that all generated motions should convey the semantic information exist in the past observation. This could be particularly important in certain applications, such as automatic animation generation. In the next chapter, we introduce another VAE-based framework that not only encourages diversity, but also encourages generating semantically plausible motions.

%% file: chapter4.tex
\chapter{Better Motion Generation via Variational Autoencoders with Learned Conditional Priors}
\label{cha:eccv}

Following previous chapter, we continue studying the problem of anticipating continuous representations of a stochastic future, with the focus on diverse human motion prediction. Similar to previous chapter, in this context, a popular approach consists of using Conditional Variational Autoencoders (CVAEs). In this chapter, we identify a major weakness of the CVAE frameworks used in existing techniques: During training, they rely on a global prior on the latent variable, thus making it independent of the conditioning signal. At inference, this translates to sampling latent variables that do not match the given CVAE condition.
In this chapter, we directly address this by conditioning the sampling of the latent variable on the CVAE condition, thus encouraging it to carry relevant information. Our experiments demonstrate that our approach not only yields samples of higher quality while retaining the semantic information contained in the observed 3D pose sequence, but also helps the model to avoid the posterior collapse, a known problem of VAEs with expressive decoders. We additionally show the generality of our approach by using it for diverse image captioning.

\section{Introduction}

Human motion prediction is the task of forecasting plausible 3D human motion continuation(s) given a sequence of past 3D human poses.
To address this problem, prior work mostly relies on recurrent encoder-decoder architectures, where the encoder processes the observed motion, and the decoder generates a single estimated future trajectory given the encoded representation of the past~\citep{martinez2017human,gui2018adversarial,walker2017pose,kundu2018bihmp,barsoum2018hp,pavllo2019modeling,pavllo2018quaternet,wei2019motion}. While this approach yields valid future motion, it tends to ignore the fact that human motion is stochastic in nature; given one single observation, multiple diverse continuations of the motion are likely and plausible. The lack of stochasticity of these encoder-decoder methods ensues from the fact that both the network operations and the sequences in the training dataset are deterministic. In this chapter, we introduce an approach to modeling this stochasticity by learning \emph{multiple modes} of human motion.

Recent attempts that account for human motion stochasticity rely on combining a random vector with an encoding of the observed pose sequence~\citep{butepage2018anticipating,yan2018mt,barsoum2018hp,walker2017pose,kundu2018bihmp,lin2018human,aliakbarian2019MixAndMatch}. In particular, the state-of-the-art approaches to diverse human motion prediction~\citep{aliakbarian2019MixAndMatch,yan2018mt} make use of conditional variational auto-encoders (CVAEs). In this chapter, we argue that standard CVAEs are ill-suited to this task for the following reason.

\begin{figure}[t]
    \centering
    \includegraphics[width=\textwidth]{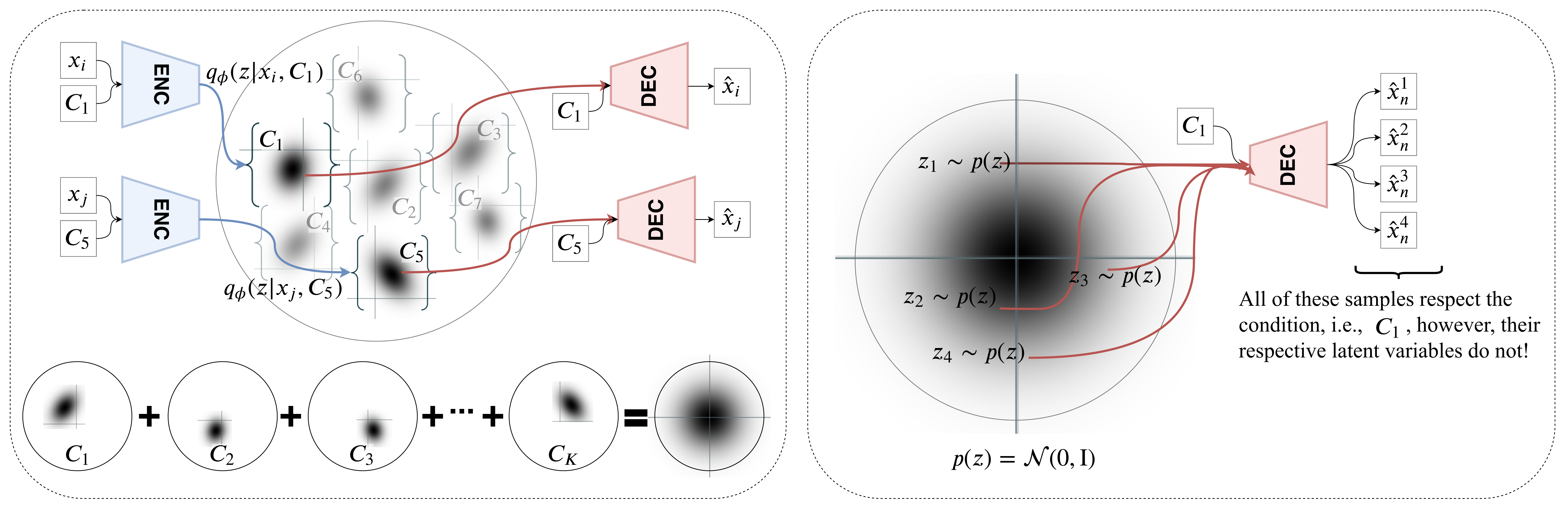}
    \caption{Training and inference for a standard CVAE. \textbf{(Left)} During the training phase, the encoder takes as input the 
    combination
    of the data and the corresponding condition and compresses it into the latent space. The decoder then samples a latent variable from the approximate posterior computed by the encoder, 
    combines
    the latent variable with the conditioning signal, and reconstructs the data. The approximate posterior distribution of \emph{all} training samples, i.e., for all conditioning signals, 
    is encouraged to match the prior distribution. \textbf{(Right)} During inference, the decoder samples different latent variables from the prior distribution, 
    combines
    them with the condition, and generates samples that all respect the conditioning signal. However, there is no guarantee that the latent variable is sampled from the region of the prior that corresponds to the given condition.}
    \label{fig:cvae_train_inference}
\end{figure}{}

In essence, VAEs utilize neural networks to learn the distribution of the data. To this end, VAEs first learn to generate a latent variable $z$ given the data $x$, i.e., approximate the posterior distribution $q_\phi(z|x)$, where $\phi$ are the parameters of a neural network, the encoder, whose goal is to model the variation of the data. From this latent random variable $z$, VAEs then generate a new sample $x$ by learning $p_\theta(x|z)$, where $\theta$ denotes the parameters of another neural network, the decoder, whose goal is to maximize the log likelihood of the data. These two networks, i.e., the encoder and the decoder, are trained jointly, using a prior over the latent variable. By using a variational approximation of the posterior, training translates to maximizing the variational lower bound of the log likelihood with respect to the parameters $\phi$ and $\theta$, given by
\begin{align}
    \log p_\theta(x) \geq \mathbb{E}_{q_\phi(z|x)}\Big[\log p_\theta(x|z)\Big] - KL\Big(q_\phi(z|x) \,||\, p(z)\Big)\;,
\end{align}
where the second term on the right hand side encodes the KL divergence between the posterior $q_\phi(z|x)$ and a chosen prior distribution $p(z)$. As an extension to VAEs, CVAEs use auxiliary information, i.e., the conditioning variable or observation, to generate the data $x$. In the standard setting, both the encoder and the decoder are conditioned on the conditioning variable $c$. That is, the encoder becomes $q_\phi(z|x,c)$ and the decoder $p_\theta(x|z,c)$. Then, in theory, the objective of the model should become 
\begin{align}
    \log p_\theta(x|c) \geq \mathbb{E}_{q_\phi(z|x,c)}\Big[\log p_\theta(x|z,c)\Big] - KL\Big(q_\phi(z|x,c) \,||\, p(z|c)\Big)\;.
    \label{eq:cvae_elbo}
\end{align}

In practice, however, \textit{the prior distribution of the latent variable is still assumed to be independent of $c$, i.e., $p(z \mid c) = p(z)$.} As illustrated by Fig.~\ref{fig:cvae_train_inference}, at test time, this translates to sampling a latent variable from a region of the prior that is unlikely to be highly correlated with the (observed) condition.

In this chapter, we overcome this limitation by explicitly making the sampling of the latent variable depend on the condition. In other words, instead of using $p(z)$ as prior distribution, we truly use $p(z|c)$. This not only respects the theory behind the design of CVAEs, but, as we empirically demonstrate, leads to generating motions of higher quality, that preserve the context of the conditioning signal, i.e., the observed past motion.  To achieve this, we develop a CVAE architecture that learns a distribution not only of the latent variable but also of the conditioning one. We then use this distribution as a prior over the latent variable, making its sampling explicitly dependent on the condition.
As such, we name our method \textbf{\texttt{LCP-VAE}}, for \textbf{\texttt{L}}earned \textbf{\texttt{C}}onditional \textbf{\texttt{P}}rior.

Our experiments show that
not only does \texttt{LCP-VAE} yield a much wider variety of plausible samples than 
state-of-the-art stochastic motion prediction methods, but it also preserves the semantic information of the condition, such as the type of action performed by the person,
without explicitly exploiting this information. 
We also show that, by unifying latent variable sampling and conditioning, we can mitigate the posterior collapse problem, a well-known issue for VAEs with expressive decoders~\citep{bowman2015generating,yang2017improved,kim2018semi,gulrajani2016pixelvae,liu2019cyclical,semeniuta2017hybrid,zhao2017infovae,tolstikhin2017wasserstein,chen2016variational,alemi2017fixing,he2019lagging,li2019surprisingly,goyal2017z,lucas2018auxiliary,dieng2018avoiding,van2017neural,guu2018generating,xu2018spherical,davidson2018hyperspherical,razavi2019preventing}, but unexplored for CVAEs. For more detail on this problem, we refer the reader to Appendix A.
Finally, we show that our approach generalizes to the task of diverse image captioning. 

\section{Related Work}
Because of space limitation, in this section, we focus on the human motion prediction literature. We nonetheless review of the literature on image captioning work and posterior collapse in Appendix A.

Most motion prediction methods are based on \emph{deterministic} models~\citep{pavllo2018quaternet,wei2019motion,pavllo2019modeling,gui2018adversarial,jain2016structural,martinez2017human,gui2018few,fragkiadaki2015recurrent,ghosh2017learning}, casting motion prediction as a regression task where only one outcome is possible given the observation. While this may produce accurate predictions, it fails to reflect the stochastic nature of human motion, where multiple futures can be highly likely for a single given series of observations. Modeling stochasticity is the topic of this chapter, and we therefore focus the discussion below on the other methods that have attempted to do so.

The general trend to incorporate variations in the predicted motions consists of combining information about the observed pose sequence with a random vector. In this context, two types of approaches have been studied: The techniques that directly incorporate the random vector into the RNN decoder and those that make use of an additional CVAE. 
In the first class of methods,~\citep{lin2018human} samples a random vector $z_t\sim\mathcal{N}(0,I)$ at each time step and adds it to the pose input of the RNN decoder. By relying on different random vectors at each time step, however, this strategy is prone to generating discontinuous motions. 
To overcome this,~\citep{kundu2018bihmp} makes use of a single random vector to generate the entire sequence. 
As we will show in our experiments, by relying on concatenation, these two methods contain parameters that are specific to the random vector, and thus give the model the flexibility to ignore this information. 
In~\citep{barsoum2018hp}, instead of using concatenation, the random vector is added to the hidden state produced by the RNN encoder. While addition prevents having parameters that are specific to the random vector, this vector is first transformed by multiplication with a learnable parameter matrix, and thus can again be zeroed out so as to remove the source of diversity, as observed in our experiments. 

The second category of stochastic methods introduce an additional CVAE between the RNN encoder and decoder. 
In this context,~\citep{walker2017pose} proposes to directly use the pose as conditioning variable. As will be shown in our experiments, while this approach is able to maintain some degree of diversity, albeit less than ours, it yields motions of lower quality because of its use of independent random vectors at each time step. 
Instead of perturbing the pose,~\citep{yan2018mt} uses the RNN decoder hidden state as conditioning variable in the CVAE, concatenating it with the random vector. While this approach generates high-quality motions, it suffers from the fact that the CVAE decoder gives the model the flexibility to ignore the random vector, which therefore yields low-diversity outputs. 
To overcome this,~\citep{aliakbarian2019MixAndMatch} perturbs the hidden states via a stochastic Mix-and-Match operation instead of concatenation. Through such a perturbation, the decoder is not able decouple the noise and the condition. However, since the perturbation is not learned and is a non-parametric operation, the quality and the context of the generated motion are inferior to those obtained with our approach.
More importantly, all of the above-mentioned CVAE-based approaches use priors that are independent of the condition. We will show in our experiments that such designs are ill-suited for human motion prediction. By contrast, our approach uses a conditional prior and is thus able to generate diverse motions of higher quality, carrying the contextual information of the conditioning signal.

\section{Unifying Sampling Latent Variable and Conditioning}

In this section, we introduce our approach as a general framework with a new conditioning scheme for CVAEs that is capable of generating diverse and \textit{plausible} samples, where the latent variables are sampled from an appropriate region of the prior distribution. In essence, our framework consists of two autoencoders, one acting on the conditioning signal and the other on the samples we wish to model. The latent representation of the condition then serves as a conditioning variable to generate samples from a learned distribution.

As discussed above, we are interested in problems that are stochastic in nature; given one condition, multiple plausible and natural samples are likely. However, for complicated tasks such as motion prediction, the training data is typically insufficiently sampled, in that, for any given condition, the dataset contains only a single sample, in effect making the data appear deterministic. For instance, in motion prediction, we never observed twice the same past motion with two different future ones. Moreover, for such problems, the condition provides the core signal to generate a good sample, even in a deterministic model. Therefore, it is highly likely that a CVAE trained for this task learns to ignore the latent variable and rely only on the condition to produce its output~\citep{aliakbarian2019MixAndMatch,yan2018mt}. This relates to the posterior collapse problem in strongly conditioned VAEs~\citep{aliakbarian2019MixAndMatch}, as discussed in more detail in Appendix A. 
Below, we address this by forcing the sampling of the random latent variable to depend on the conditioning one. By making this dependency explicit, we (1) sample an informative latent variable \textit{given} the condition, and thus generate a sample of higher quality, and (2) prevent the network from ignoring the latent variable in the presence of a strong condition, thus enabling it to generate diverse outputs.

Note that conditioning the VAE \emph{encoder} via standard strategies, e.g., concatenation, is perfectly fine, since the two inputs to the encoder, i.e., the data and the condition, are deterministic and useful to compress the sample into the latent space. However, conditioning the VAE \emph{decoder} requires special care, which is what we focus on below.

\begin{figure}[t]
\centering
\includegraphics[width=\textwidth]{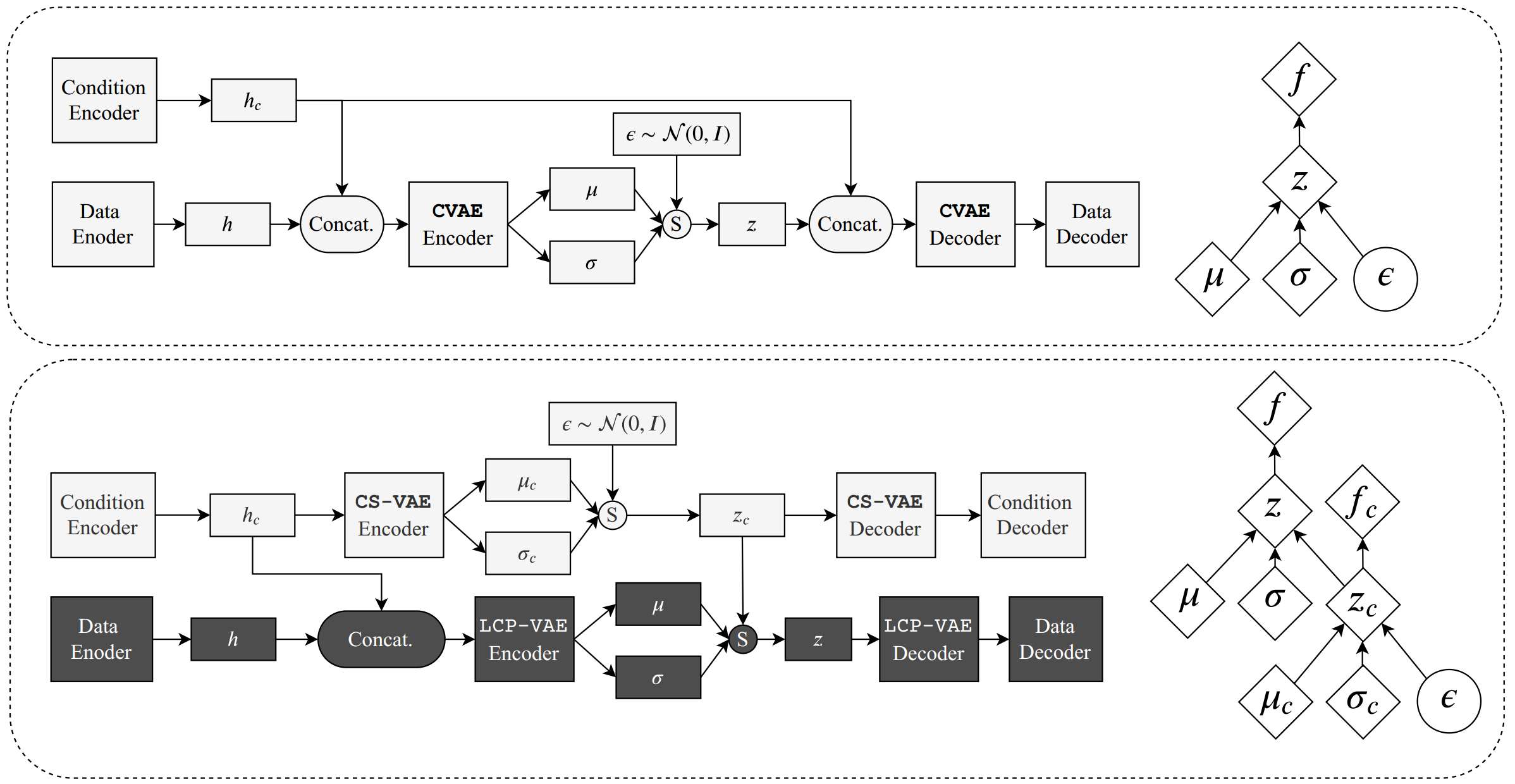}
\caption{Illustration of a CVAE architecture and an \texttt{LCP-VAE} architecture. \textbf{(Top)} In a standard CVAE architecture, the latent variable $z$ is sampled from a standard Normal distribution, independently of the condition. This leads to the reparametrization trick illustrated on the right. In this case, the approximate posterior is normally distributed.
\textbf{(Bottom)} In an \texttt{LCP-VAE}, the sampling of $z$ is conditioned on the CVAE condition via the latent variable $z_c$. Specifically, the posterior distribution of the condition acts as prior on the data posterior. This corresponds to the \textit{extended reparameterization trick} illustrated on the right.
Note that the approximate posterior is not normally distributed anymore.}
\label{fig:method}
\end{figure}{}

\subsection{Stochastically Conditioning the Decoder} 

We propose to make the sampling of the latent variable from the prior/posterior distribution explicitly depend on the condition instead of treating these two variables as independent. To this end, we first learn the distribution of the condition via a simple VAE, which we refer to as \texttt{CS-VAE} because this VAE acts on the conditioning signal. The goal of \texttt{CS-VAE} is to reconstruct the condition, e.g., the observed past motion, given its latent representation. We take the prior of \texttt{CS-VAE} as a standard Normal distribution $\mathcal{N}(0,I)$. Following Kingma and Welling~\citep{kingma2013auto}, this allows us to approximate the \texttt{CS-VAE} posterior with another sample from a Normal distribution $\epsilon\sim\mathcal{N}(0,I)$ via the reparametrization trick
\begin{align}
    z_c = \mu_c + \sigma_c \odot \epsilon\;,
    \label{eq:observation_reparam}
\end{align}
where $\mu_c$ and $\sigma_c$ are the parameter vectors of the posterior distribution generated by the VAE encoder, and thus  $z_c\sim\mathcal{N}(\mu_c, \text{diag}(\sigma_c)^2)$.

Following the same strategy for the data VAE 
translates to treating the conditioning and the data latent variables independently, which we seek to avoid. Therefore, as illustrated in Fig.~\ref{fig:method} (Bottom), we instead define the \texttt{LCP-VAE} posterior not as directly normally distributed but conditioned on the posterior of \texttt{CS-VAE}. To this end, we extend the standard reparameterization trick as 
\begin{align}
    z =  & \mu + \sigma \odot z_c \nonumber \\ = & \mu + \sigma \odot (\mu_c + \sigma_c \odot \epsilon) \nonumber \\ = & \underbrace{(\mu + \sigma\odot\mu_c)}_\text{\texttt{LCP-VAE}'s mean} + \underbrace{(\sigma\odot\sigma_c)}_\text{\texttt{LCP-VAE}'s std.}\odot \epsilon\;,
    \label{eq:future_reparam}
\end{align}
where $z_c$ comes from Eq.~\ref{eq:observation_reparam}, and $\mu$ and $\sigma$ are the parameter vectors generated by the \texttt{LCP-VAE} encoder. In fact, $z_c$ in Eq.~\ref{eq:observation_reparam} is a sample from the scaled and translated version of $\mathcal{N}(0,I)$ given $\mu_c$ and $\sigma_c$, and $z$ in Eq.~\ref{eq:future_reparam} is a sample from the scaled and translated version of $\mathcal{N}(\mu_c,\text{diag}(\sigma_c)^2)$ given $\mu$ and $\sigma$. Since we have access to the observations during both training and testing, we always sample $z_c$ from the condition posterior. As $z$ is sampled given $z_c$, one expects the latent variable $z$ to carry information about the strong condition, and thus a sample generated from $z$ to correspond to a plausible sample given the condition. This extended reparameterization trick lets us sample one single informative latent variable that contains information about both the data and the conditioning signal. This further allows us to avoid conditioning the \texttt{LCP-VAE} decoder by concatenating the latent variable with a deterministic representation of the condition.
As will be shown in our experiments, and in more detail in Appendix A, our sampling strategy not only yields higher-quality motions that retain the semantic information contained in the conditioning signal, but also helps the model to avoid posterior collapse.
However, it changes the variational family of the \texttt{LCP-VAE} posterior. In fact, the posterior is no longer $\mathcal{N}(\mu,\text{diag}(\sigma)^2)$, but a Gaussian distribution with mean  $\mu + \sigma\odot\mu_c$ and  covariance matrix $\text{diag}(\sigma\odot\sigma_c)^2$. This will be accounted for when designing the KL divergence loss discussed below.

\begin{figure}[t]
    \centering
    \begin{tabular}{cc}
            \includegraphics[width=.5\textwidth]{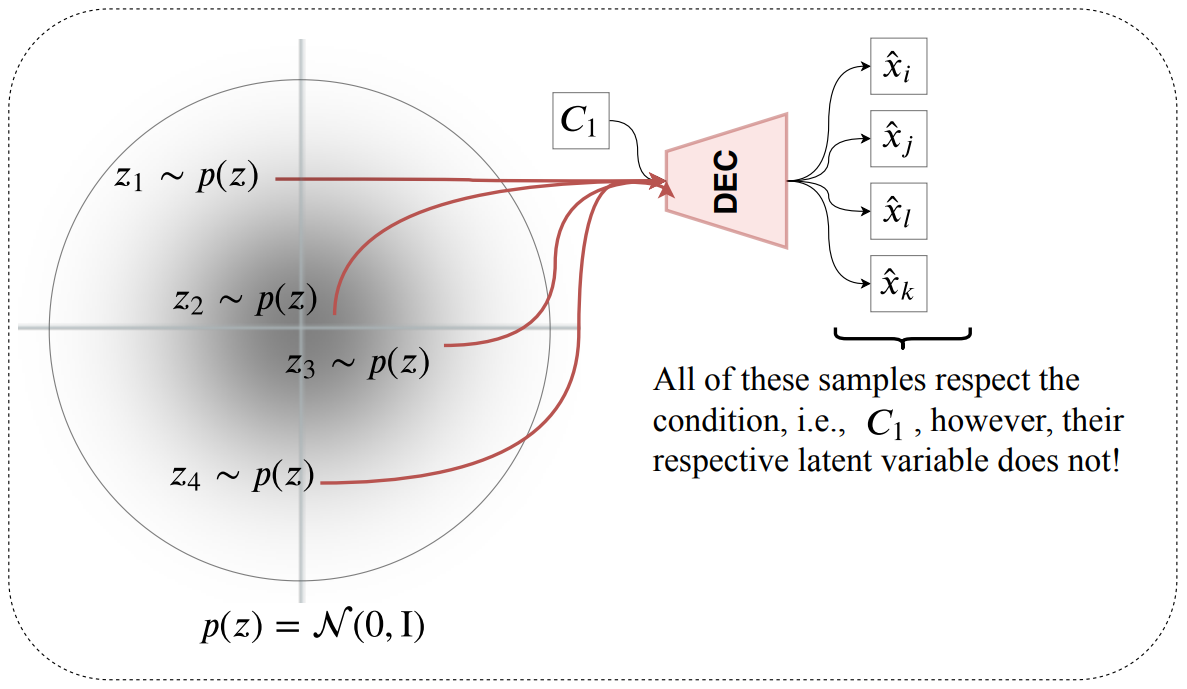} &
            \includegraphics[width=.475\textwidth]{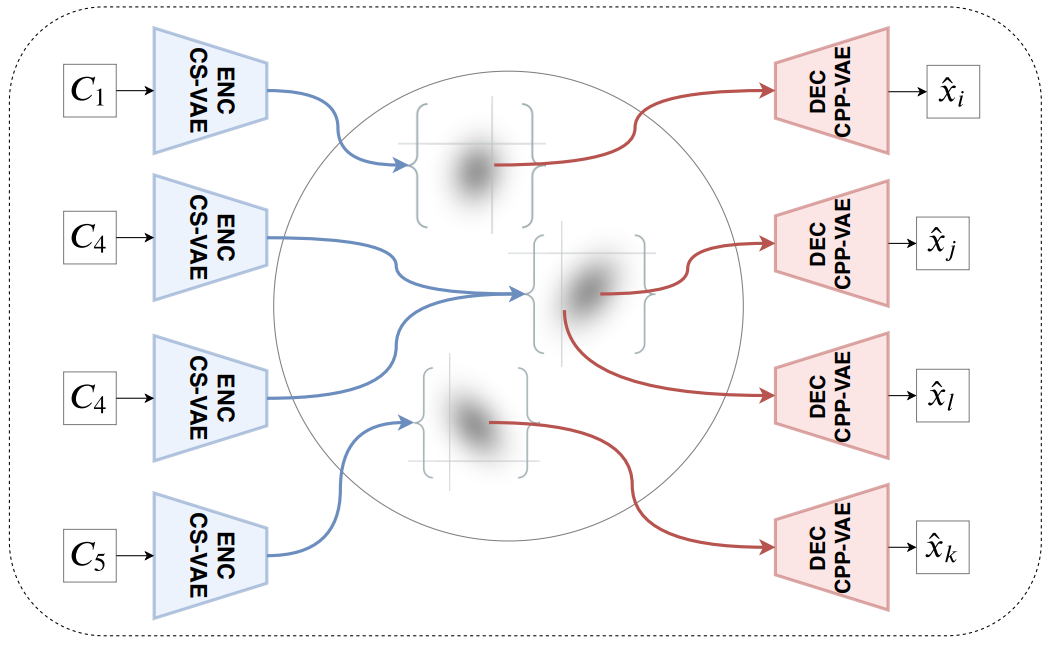}

    \end{tabular}
    \caption{Inference procedure in an \texttt{LCP-VAE} and a CVAE. \textbf{(Left)} In a standard CVAE, at inference time, since the prior distribution is independent of the conditioning signal, we are unlikely to sample a latent variable that corresponds to the region of the latent space that is related to the observed condition. \textbf{(Right)} In an \texttt{LCP-VAE}, at inference time, since we have access to the conditioning signal, we use the \texttt{CS-VAE} encoder to approximate the posterior of each condition. To generate a sample given a condition, the \texttt{LCP-VAE} decoder then samples a latent variable from the posterior of its condition, instead of using a general prior distribution as in CVAEs, and generate a sample. For instance, $\hat{x}_j$ and $\hat{x}_l$ are conditioned on $C_4$ and their corresponding latent variables are sampled from a very similar region. By contrast, $\hat{x}_k$ is generated by a latent variable sampled from a completely different region, i.e., the region corresponding to the approximate posterior of $C_5$. 
    }
    \label{fig:cvae_cpp_vae_inference}
\end{figure}{}

\subsection{Learning}
To learn the parameters of our model, we rely on the availability of a dataset $D=\{X_1, X_2, ..., X_N\}$ containing $N$ training samples $X_i$. Each training sample is a pair of condition and data sample. For \texttt{CS-VAE}, which learns the distribution of the condition, we define the loss as the KL divergence between its posterior and the standard Gaussian prior, that is,
\begin{align}
    \mathcal{L}_{prior}^{\texttt{CS-VAE}} = KL\Big(\mathcal{N}(\mu_c, \text{diag}(\sigma_c)^2) \Big\| \mathcal{N}(0,I)\Big) =  -\frac{1}{2}\sum_{j=1}^d \Big(1+\log(\sigma_{c_j}^2) - \mu_{c_j}^2 - \sigma_{c_j}^2 \Big)\;,
    \label{eq:cs-kl-loss}
\end{align}
where $d$ is the dimension of the latent variable $z_c$. By contrast, for \texttt{LCP-VAE}, we define the loss as the KL divergence between the posterior of \texttt{LCP-VAE} and the posterior of  \texttt{CS-VAE}, i.e., of the condition. To this end, we freeze the weights of \texttt{CS-VAE} before computing the KL divergence, since we do not want to move the posterior of the condition but that of the data. The KL divergence is then computed as the divergence between two multivariate Normal distributions, encoded by their mean vectors and covariance matrices as
\begin{align}
    \mathcal{L}_{prior}^{\texttt{LCP-VAE}} = KL\Big(\mathcal{N}(\mu + \sigma\odot\mu_c, \text{diag}(\sigma\odot\sigma_c)^2) \Big\| \mathcal{N}(\mu_c, \text{diag}(\sigma_c)^2)\Big)\;.
    \label{eq:cpp-kl-loss}
\end{align}
Let $\Sigma=\text{diag}(\sigma)^2$,  $\Sigma_c=\text{diag}(\sigma_c)^2$, $d$ be the dimensionality of the latent space and $tr\{\cdot\}$ the trace of a square matrix. The loss in Eq.~\ref{eq:cpp-kl-loss} can be written as\footnote{See Appendix A for more detail on the KL divergence between two multivariate Gaussians and the derivation of Eq.~\ref{eq:cpp-vae-simplified-kl}.}
\begin{align}
    \mathcal{L}_{prior}^{\texttt{LCP-VAE}} = -\frac{1}{2}\Big[\log\frac{1}{|\Sigma|}-d+
    tr\{\Sigma\} + (\mu_c-(\mu+\Sigma\mu_c))^T\Sigma_c^{-1}(\mu_c-(\mu+\Sigma\mu_c))\Big]\;. 
    \label{eq:cpp-vae-simplified-kl}
\end{align}

After computing the loss in Eq.~\ref{eq:cpp-vae-simplified-kl}, we unfreeze \texttt{CS-VAE} and update it with its previous gradient. Trying to match the posterior of \texttt{LCP-VAE} to that of \texttt{CS-VAE} allows us to effectively use our extended reparameterization trick in Eq.~\ref{eq:future_reparam}. Furthermore, we use the standard reconstruction loss for both \texttt{CS-VAE} and \texttt{LCP-VAE}, thus minimizing the mean squared error (MSE) in the case of human motion prediction and the negative log-likelihood (NLL) in the case of image captioning. 

We refer to the reconstruction losses as $\mathcal{L}_{rec}^{\texttt{CS-VAE}}$ and $\mathcal{L}_{rec}^{\texttt{LCP-VAE}}$ for \texttt{CS-VAE} and \texttt{LCP-VAE}, respectively.
Thus, our complete loss is 
\begin{align}
\mathcal{L} =  \lambda(\mathcal{L}_{prior}^{\texttt{CS-VAE}} + \mathcal{L}_{prior}^{\texttt{LCP-VAE}}) + \mathcal{L}_{rec}^{\texttt{CS-VAE}} + \mathcal{L}_{rec}^{\texttt{LCP-VAE}} \;.
\label{eq:stochastic_loss}
\end{align}
In practice, since the nature of our data is sequential (e.g., sequence of human poses in human motion prediction and sequence of words in image captioning), the \texttt{LCP-VAE} encoder is a recurrent model. Thus, we weigh the KL divergence terms by a function $\lambda$ corresponding to the KL annealing weight of~\citep{bowman2015generating}. We start from $\lambda=0$, forcing the model to encode as much information in $z$ as possible, and gradually increase it to $\lambda=1$ during training, following a logistic curve. We then continue training with $\lambda=1$.

In short, our method can be interpreted as a simple yet effective framework (designed for CVAEs) for altering the variational family of the posterior such that (1) a latent variable from this posterior distribution is explicitly sampled given the condition, both during training and inference (as illustrated in Fig.~\ref{fig:cvae_cpp_vae_inference}), and (2) the model is much less likely to suffer from posterior collapse because the mismatch between the posterior and prior distributions makes it harder for learning to drive the KL divergence of Eq.~\ref{eq:cpp-vae-simplified-kl} towards zero.

\begin{figure}[t]
    \centering
    \scriptsize
    \begin{tabular}{c}
    \toprule
    \includegraphics[width=0.91\textwidth]{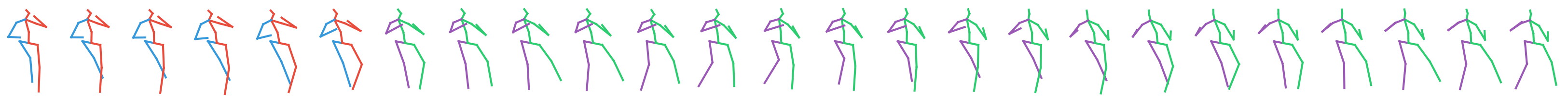} \\
    \midrule
    \includegraphics[width=0.91\textwidth]{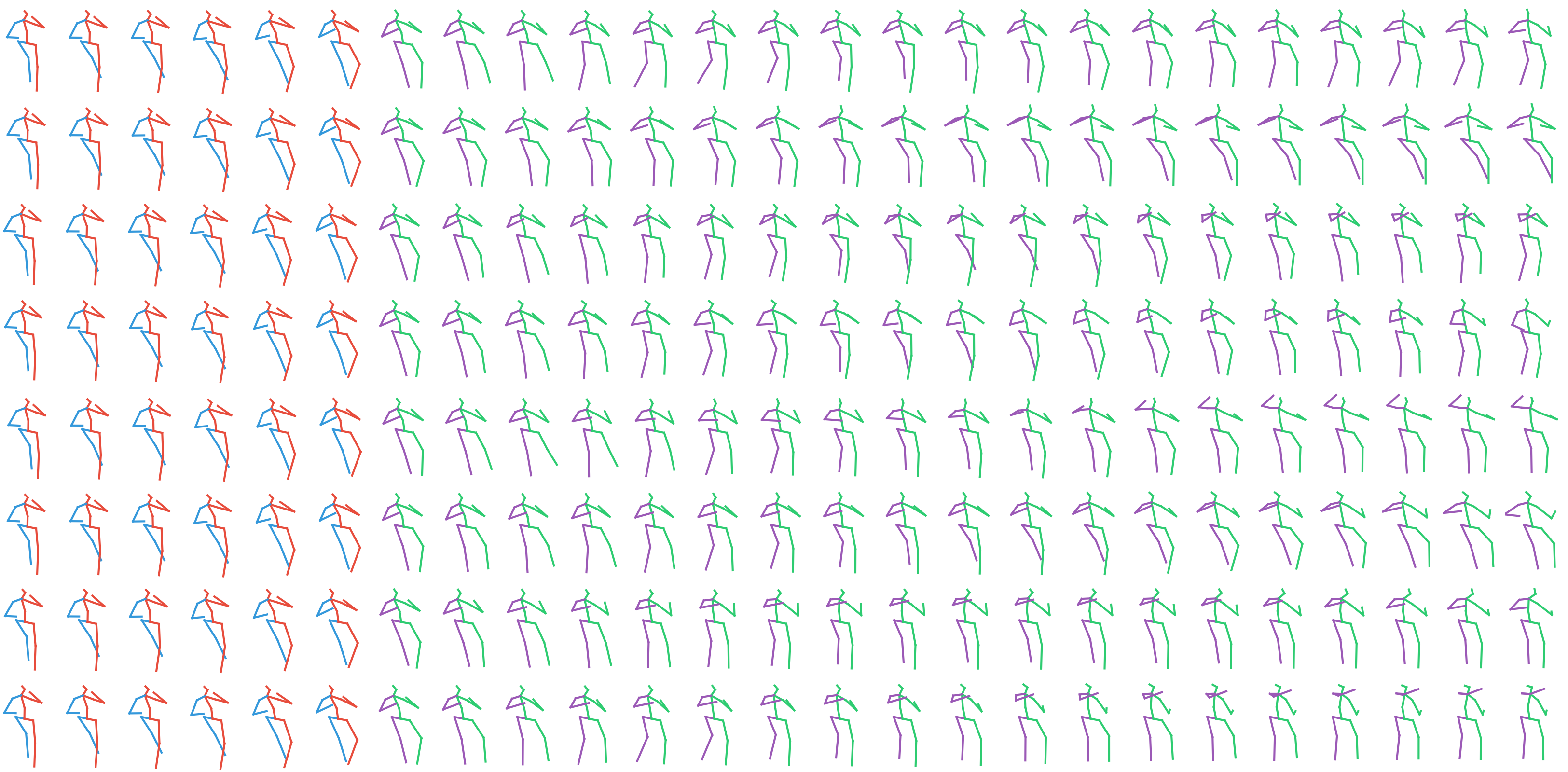} \\
    \bottomrule
    \end{tabular}{}
    \caption{Qualitative evaluation of the diversity in human motion. The first row depicts the ground-truth motion. The first six poses of each row correspond to the observation (the condition) and the rest are sampled from our model. Each row is a randomly sampled motion (not cherry picked). Note that all sampled motions are natural, with a smooth transition from the observed poses to the generated ones. Diversity increases with the length of the generated sequence.}
    \label{fig:motion_qualitative}
\end{figure}{}

\begin{table}[t]
    \centering
    \caption{Comparison of \texttt{LCP-VAE} with the stochastic motion prediction baselines on Human3.6M (top) and CMU MoCap (bottom). }
    \scriptsize
    \setlength\extrarowheight{-3pt}
    \begin{tabular}{l @{ }@{ } c@{ }@{ } c@{ }@{ } c@{ }@{ } c@{ }@{ } c}
    
    \toprule
    \multicolumn{6}{c}{\textbf{Results on Human3.6M}}\\
    \midrule
     & ELBO  (KL) & Diversity & Quality & Context & Training KL \\
     Method & \scriptsize{(Reconstructed)} & \scriptsize{(Sampled)} & \scriptsize{(Sampled)} & \scriptsize{(Sampled)} & \scriptsize{(Reconstructed)} \\
    \midrule
    MT-VAE~\citep{yan2018mt} & 0.51 (0.06) & 0.26 & 0.45 & 0.42  & 0.08 \\
    Pose-Knows~\citep{walker2017pose} & 2.08 (N/A) & 1.70 & 0.13 & 0.08 & N/A \\
    HP-GAN~\citep{barsoum2018hp} & 0.61 (N/A) & 0.48 & 0.47 & 0.35 & N/A\\
    Mix-and-Match~\citep{aliakbarian2019MixAndMatch} & 0.55 (2.03)  &\textbf{ 3.52 }& 0.42 & 0.37 & 1.98\\
    
    \texttt{LCP-VAE} & \textbf{0.41 (8.07)} &  3.12 & \textbf{0.48} & \textbf{0.54} & \textbf{6.93}\\
    \\
    \toprule
    \multicolumn{6}{c}{\textbf{Results on CMU MoCap}}\\
    \midrule
     & ELBO  (KL) & Diversity & Quality & Context & Training KL \\
     Method & \scriptsize{(Reconstructed)} & \scriptsize{(Sampled)} & \scriptsize{(Sampled)} & \scriptsize{(Sampled)} & \scriptsize{(Reconstructed)} \\
    \midrule
    MT-VAE~\citep{yan2018mt} & 0.25 (0.08) & 0.41 & 0.46 & 0.80  & 0.01 \\
    Pose-Knows~\citep{walker2017pose} & 1.93 (N/A) & 3.00 & 0.18 & 0.27 & N/A \\
    HP-GAN~\citep{barsoum2018hp} & 0.24 (N/A) & 0.43 & 0.45 & 0.73 & N/A\\
    Mix-and-Match~\citep{aliakbarian2019MixAndMatch} & 0.25 (2.92)  &\textbf{2.63} & 0.46 & 0.78 & 2.00\\
    
    \texttt{LCP-VAE} & \textbf{0.23 (4.13)} &  2.36 & \textbf{0.48} & \textbf{0.88} & \textbf{3.91}\\
    \bottomrule
    \end{tabular}
    
    \label{tab:stoch}
\end{table}

\begin{table}[t]
\scriptsize
\setlength\extrarowheight{-3pt}
    \caption{Comparison of the generated motions with the ground-truth future motions in terms of context. 
    The gap between the performance of the state-of-the-art pose-based action classifier~\citep{li2018co} with and without true future motions is 0.22 / 0.54. Using our predictions, this gap decreases to 0.06 / 0.08, showing that our predictions reflect the class label. H3.6M / CMU represent Human3.6M and CMU MoCap, respectively.}
    \label{tab:context_upperbound}
    \centering
        \begin{tabular}{l c c c}
    \toprule
    Setting & Obs. & Future Motion & Context (H3.6M / CMU) \\
    \midrule
    Lower bound & GT & Zero velocity & 0.38 / 0.42\\
    Upper bound (GT poses as future motion) & GT & GT & 0.60 / 0.96\\
    
    Ours (sampled motions as future motion) & GT & Sampled from \texttt{LCP-VAE} & 0.54 / 0.88\\
    \bottomrule
    \end{tabular}
\end{table}{}

\section{Experiments}

In this chapter, we mainly focus on stochastic human motion prediction, where the goal is to generate diverse and plausible continuations of given past observations. Additionally, to show that our \texttt{LCP-VAE} generalizes to other domains, we tackle the problem of stochastic image captioning, where, given an image representation, the task is to generate diverse yet related captions. 

\subsection{Diverse Human Motion Prediction}
\paragraph{Datasets.}
To evaluate the effectiveness of our approach on the task of stochastic human motion prediction, we use the Human3.6M~\citep{h36m_pami} and CMU MoCap\footnote{Available at \texttt{http://mocap.cs.cmu.edu/}.} datasets, two large publicly-available motion capture (mocap) datasets. 
Human3.6M comprises more than 800 long indoor motion sequences performed by 11 subjects, leading to 3.6M frames. Each frame contains a person annotated with 3D joint positions and rotation matrices for all 32 joints. In our experiments, for our approach and the replicated VAE-based baselines, we represent each joint in 4D quaternion space. We follow the standard preprocessing and evaluation settings used in~\citep{martinez2017human,gui2018adversarial,pavllo2018quaternet,jain2016structural}.
The CMU MoCap dataset is another large-scale motion capture dataset covering diverse human activities, such as jumping, running, walking, and playing basketball. Each frame contains a person annotated with 3D joint rotation matrices for all 38 joints. As for Human3.6M, and following standard practice~\citep{wei2019motion,li2018convolutional}, we represent each joint in 4D quaternion space.
We also evaluate our approach on the real-world Penn Action dataset~\citep{zhang2013actemes}, which contains 2326 sequences of 15 different actions, where for each person, 13 joints are annotated in 2D space. The results on Penn Action are provided in Appendix A.

\begin{figure}[t]
    \centering
    \scriptsize
    \begin{tabular}{c@{}c@{}c@{}c}
         \includegraphics[width=0.125\textwidth]{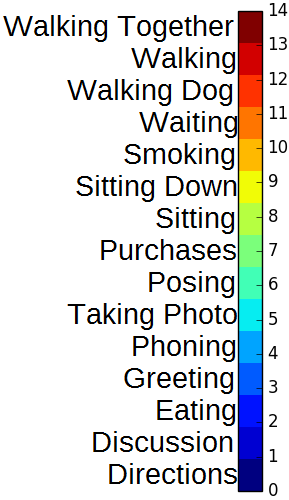} & 
         \includegraphics[width=0.29\textwidth]{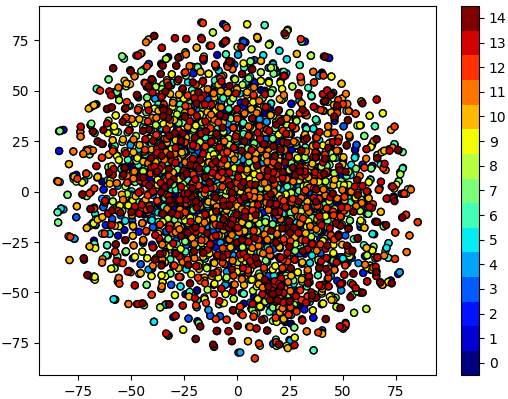} & 
         \includegraphics[width=0.29\textwidth]{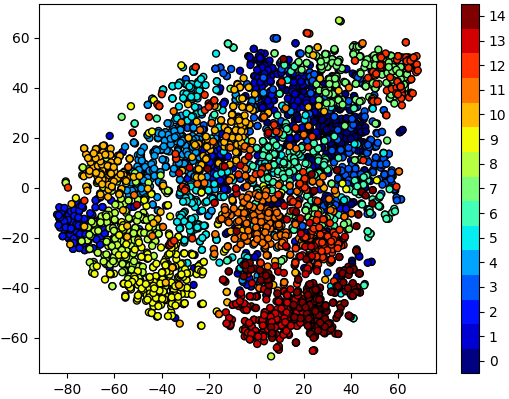} &
         \includegraphics[width=0.29\textwidth]{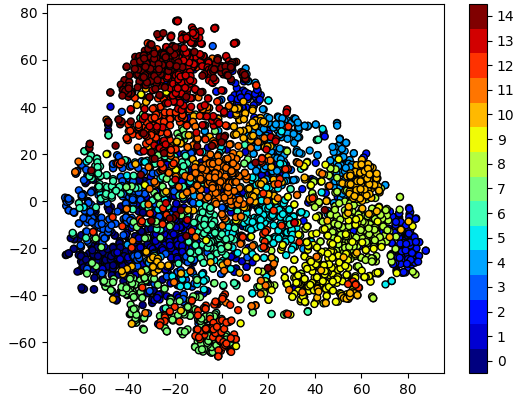} \\
         & MT-VAE~\citep{yan2018mt} & \texttt{LCP-VAE} ($\mu$) & \texttt{LCP-VAE} ($\mu+\mu_c\odot\sigma_c$)\\
    \end{tabular}
    \caption{t-SNE plots of the posterior mean for 3750 test motions. With MT-VAE~\citep{yan2018mt}, all classes are mixed, suggesting that the latent variable carries little information about the motions. By contrast, our condition-dependent sampling allows \texttt{LCP-VAE} to better preserve context. Note that some actions, such as discussion and directions, are very hard to identify and are thus spread over other actions. Others, such as walking, walking with dog, and walking together, or sitting and sitting down overlap due to their similarity.}
    \label{fig:tsne}
\end{figure}{}

\paragraph{Evaluation Metrics.}
To quantitatively evaluate our approach and other stochastic motion prediction baselines~\citep{yan2018mt,barsoum2018hp,walker2017pose,aliakbarian2019MixAndMatch}, we report the reconstruction error, commonly referred to as ELBO, along with the KL-divergence on the held-out test set. Additionally, we report quality~\citep{aliakbarian2019MixAndMatch} and diversity~\citep{yang2018diversitysensitive,aliakbarian2019MixAndMatch,yuan2019diverse} metrics, which should be considered together. Specifically, to measure the diversity of the motions generated by a stochastic model, we make use of the average distance between all pairs of the $K$ motions generated from the same observation. To measure quality, we train a binary classifier~\citep{aliakbarian2019MixAndMatch} to discriminate real (ground-truth) samples from fake (generated) ones. The accuracy of this classifier on the test set is inversely proportional to the quality of the generated motions. Furthermore, we report a context metric measured as the performance of a strong action classifier~\citep{li2018co} trained on ground-truth motions. Specifically, the classifier is tested on each of the $K$ motions generated from each observation. For $N$ observations and $K$ continuations per observation, the accuracy is measured by computing the argmax over each prediction's probability vector, and we report context as the mean class accuracy on the $K\times N$ motions. Finally, we report the training KL at convergence to show that no posterior collapse occurred. For all metrics, we use $K=50$ motions per test observation. For all experiments, we use 16 frames (i.e., 640ms) as observation to generate the next 60 frames (i.e., 2.4sec). 

\paragraph{Evaluating Stochasticity.} 
In Table~\ref{tab:stoch}, we compare our approach (whose detailed architecture is described in Appendix A) with the state-of-the-art stochastic motion prediction models~\citep{yan2018mt,aliakbarian2019MixAndMatch,walker2017pose,barsoum2018hp}. Note that one should consider the reported metrics jointly to truly evaluate a stochastic model. For instance, while MT-VAE~\citep{yan2018mt} and HP-GAN~\citep{barsoum2018hp} generate high-quality motions, they are not diverse. Conversely, while Pose-Knows~\citep{walker2017pose} generates diverse motions, they are of low quality. By contrast, our approach generates both high quality and diverse motions. This is also the case of Mix-and-Match~\citep{aliakbarian2019MixAndMatch}, which, however, preserves much less context. In fact, none of the baselines effectively conveys the context of the observation to the generated motions. As shown in Table~\ref{tab:context_upperbound}, the upper bound for context on Human3.6M is 0.60 (i.e., the classifier~\citep{li2018co} performance given the ground-truth motions). Our approach yields a context of 0.54 when given only about 20\% of the data. We observe a similar behavior on the CMU MoCap dataset, shown in Table~\ref{tab:context_upperbound}. Altogether, as also supported by the qualitative results of Fig.~\ref{fig:motion_qualitative} and in Appendix A, our approach yields diverse, high-quality and context-preserving predictions. This is further evidenced by the t-SNE~\citep{maaten2008visualizing} plots of Fig.~\ref{fig:tsne}, where different samples of various actions are better separated for our approach than for, e.g., MT-VAE~\citep{yan2018mt}.
For further discussion of the baselines, a deeper insight of their behavior under different evaluation metrics, and a discussion of the relation of each method's performance to posterior collapse, we refer the reader to Appendix A.

\begin{table}[t]
\scriptsize
\setlength\extrarowheight{-3pt}
    \centering
        \caption{Comparison with the state-of-the-art stochastic motion prediction models for 4 actions of Human3.6M (all methods use the best of $K=50$ sampled motions).}
    
    \scalebox{0.93}{
    \begin{tabular}{l @{ }@{ } c c c c c c @{ }@{ } @{ }@{ } c c c c c c}
\toprule
    & \multicolumn{6}{c}{Walking} & \multicolumn{6}{c}{Eating} \\
    \midrule
    Method & 80  & 160 & 320 & 400 & 560 & 1000  & 80  & 160 & 320 & 400 & 560 & 1000  \\
    \midrule
    
    MT-VAE~\citep{yan2018mt} & 
    0.73 & 0.79 & 0.90 & 0.93 & 0.95 & 1.05 & 
    0.68 & 0.74 & 0.95 & 1.00 & 1.03 & 1.38 \\
    
    HP-GAN~\citep{barsoum2018hp} & 
    0.61 & 0.62 & 0.71 & 0.79 & 0.83 & 1.07 & 
    0.53 & 0.67 & 0.79 & 0.88 & 0.97 & 1.12 \\
    
    Pose-Knows~\citep{walker2017pose} & 
    0.56 & 0.66 & 0.98 & 1.05 & 1.28 & 1.60 & 
    0.44 & 0.60 & 0.71 & 0.84 & 1.05 & 1.54 \\ 
    
    Mix\&Match~\citep{aliakbarian2019MixAndMatch} & 
    0.33 & 0.48 & 0.56 & 0.58 & 0.64 & \textbf{0.68} &
    0.23 & 0.34 & 0.41 & \textbf{0.50} & 0.61 & 0.91  \\
    
    \texttt{LCP-VAE} & 
    \textbf{0.22} & \textbf{0.36} & \textbf{0.47} & \textbf{0.52} & \textbf{0.58} & 0.69 & 
    \textbf{0.19} & \textbf{0.28} & \textbf{0.40} & 0.51 & \textbf{0.58} & \textbf{0.90} \\ 
    \midrule
    
    & \multicolumn{6}{c}{Smoking} & \multicolumn{6}{c}{Discussion}\\
    \midrule
    Method & 80  & 160 & 320 & 400 & 560 & 1000  & 80  & 160 & 320 & 400 & 560 & 1000  \\
    \midrule
    
    MT-VAE~\citep{yan2018mt} & 
    1.00 & 1.14 & 1.43 & 1.44 & 1.68 & 1.99 & 
    0.80 & 1.01 & 1.22 & 1.35 & 1.56 & 1.69 \\
    
    HP-GAN~\citep{barsoum2018hp} & 
    0.64 & 0.78 & 1.05 & 1.12 & 1.64 & 1.84 & 
    0.79 & 1.00 & 1.12 & 1.29 & 1.43 & 1.71 \\
    
    Pose-Knows~\citep{walker2017pose} & 
    0.59 & 0.83 & 1.25 & 1.36 & 1.67 & 2.03 & 
    0.73 & 1.10 & 1.33 & 1.34 & 1.45 & 1.85 \\
    
    Mix\&Match~\citep{aliakbarian2019MixAndMatch} & 
    \textbf{0.23} & \textbf{0.42} & 0.79 & 0.77 & 0.82 & 1.25 & 
    0.25 & 0.60 & 0.83 & 0.89 & 1.12 & 1.30 \\
    
    \texttt{LCP-VAE} & 
    \textbf{0.23} & 0.43 & \textbf{0.77} &\textbf{ 0.75} & \textbf{0.78} & \textbf{1.23} & 
    \textbf{0.21} & \textbf{0.52} & \textbf{0.81} & \textbf{0.84} & \textbf{1.04} & \textbf{1.28} \\
    
\bottomrule
    \end{tabular}
    }
    \label{tab:mae_stoch}
    \vspace{-5pt}
\end{table}
\begin{table}[t]
    \centering
    \scriptsize
    \caption{Comparison with the state-of-the-art deterministic models for 4 actions of Human3.6M. }
    \scalebox{0.9}{
    \begin{tabular}{l  c c c c c c  c c c c c c}
\toprule
    & \multicolumn{6}{c}{Walking} & \multicolumn{6}{c}{Eating} \\
    \midrule
    Method & 80  & 160 & 320 & 400 & 560 & 1000  & 80  & 160 & 320 & 400 & 560 & 1000  \\
    \midrule
    Zero Velocity & 
    0.39 & 0.86 & 0.99 & 1.15 & 1.35 & 1.32 & 
    0.27 & 0.48 & 0.73 & 0.86 & 1.04 & 1.38 \\
    
    LSTM-3LR~\citep{fragkiadaki2015recurrent} & 
    1.18 & 1.50 & 1.67 & 1.76 & 1.81 & 2.20  & 
    1.36 & 1.79 & 2.29 & 2.42 & 2.49 & 2.82 \\
    
    SRNN~\citep{jain2016structural} & 
    1.08 & 1.34 & 1.60 & 1.80 & 1.90 & 2.13 & 
    1.35 & 1.71 & 2.12 & 2.21 & 2.28 & 2.58 \\ 
    
    DAE-LSTM~\citep{ghosh2017learning} & 
    1.00 & 1.11 & 1.39 & 1.48 & 1.55 & 1.39 & 
    1.31 & 1.49 & 1.86 & 1.89 & 1.76 & 2.01 \\ 
    
    GRU~\citep{martinez2017human} & 
    0.28 & 0.49 & 0.72 & 0.81 & 0.93 & 1.03 & 
    0.23 & 0.39 & 0.62 & 0.76 & 0.95 & 1.08 \\

    AGED~\citep{gui2018adversarial} & 
    0.22 & 0.36 & 0.55 & 0.67 & 0.78 & 0.91 & 
    0.17 & 0.28 & 0.51 & 0.64 & 0.86 & 0.93 \\

    DCT-GCN~\citep{wei2019motion} & 
    \textbf{0.18} & \textbf{0.31} & 0.49 & 0.56 & 0.65 & \textbf{0.67} & 
    \textbf{0.16} & 0.29 & 0.50 & 0.62 & 0.76 & 1.12 \\ 
    
    \texttt{LCP-VAE} ($z=\mu_c$) & 
    0.20 & 0.34 & \textbf{0.48} & \textbf{0.53} & \textbf{0.57} & 0.71 & 
    0.20 & \textbf{0.26} & \textbf{0.44} & \textbf{0.52} & \textbf{0.61} & \textbf{0.92}\\
    
    \midrule
    & \multicolumn{6}{c}{Smoking} & \multicolumn{6}{c}{Discussion}\\
    \midrule
    Method & 80  & 160 & 320 & 400 & 560 & 1000  & 80  & 160 & 320 & 400 & 560 & 1000  \\
    \midrule
    Zero Velocity & 
    0.26 & 0.48 & 0.97 & 0.95 & 1.02 & 1.69 & 
    0.31 & 0.67 & 0.94 & 1.04 & 1.41 & 1.96 \\
    
    LSTM-3LR~\citep{fragkiadaki2015recurrent} & 
    2.05 & 2.34 & 3.10 & 3.18 & 3.24 & 3.42  & 
    2.25 & 2.33 & 2.45 & 2.46 & 2.48 & 2.93 \\
    
    SRNN~\citep{jain2016structural} & 
    1.90 & 2.30 & 2.90 & 3.10 & 3.21 & 3.23 & 
    1.67 & 2.03 & 2.20 & 2.31 & 2.39 & 2.43 \\
    
    DAE-LSTM~\citep{ghosh2017learning} & 
    0.92 & 1.03 & 1.15 & 1.25 & 1.38 & 1.77 & 
    1.11 & 1.20 & 1.38 & 1.42 & 1.53 & 1.73 \\
    
    GRU~\citep{martinez2017human} & 
    0.33 & 0.61 & 1.05 & 1.15 & 1.25 & 1.50 & 
    0.31 & 0.68 & 1.01 & 1.09 & 1.43 & 1.69 \\
    
    AGED~\citep{gui2018adversarial} & 
    0.27 & 0.43 & 0.82 & 0.84 & 1.06 & 1.21 & 
    0.27 & 0.56 & \textbf{0.76} & 0.83 & 1.25 & 1.30 \\

    DCT-GCN~\citep{wei2019motion} & 
    0.22 & \textbf{0.41} & 0.86 & 0.80 & 0.87 & 1.57 & 
    \textbf{0.20} & \textbf{0.51} & 0.77 & 0.85 & 1.33 & 1.70 \\
    
    \texttt{LCP-VAE} ($z=\mu_c$) & 
    \textbf{0.21} & 0.43 & \textbf{0.79} & \textbf{0.79} & \textbf{0.77} & \textbf{1.15} & 
    0.22 & 0.55 & 0.79 & \textbf{0.81} & \textbf{1.05} & \textbf{1.28} \\
    \bottomrule
    \end{tabular}
    }
    \label{tab:deterministic}
\end{table}

\paragraph{Evaluating Sampling Quality.}
To further evaluate the sampling quality, we evaluate stochastic baselines using the standard mean angle error (MAE) metric in Euler space. To this end, we use the best of the $K=50$ generated motions for each observation (referred to as S-MSE in~\citep{yan2018mt}). A model that generates more diverse motions has higher chances of producing a motion  close to the ground-truth one. As shown in Table~\ref{tab:mae_stoch}, this is the case with our approach and Mix-and-Match~\citep{aliakbarian2019MixAndMatch}, which both yield higher diversity. However, our approach performs better thanks to its context-preserving latent representation and its higher quality of the generated motions.

In Table~\ref{tab:deterministic}, we compare our approach with the state-of-the-art deterministic motion prediction models~\citep{martinez2017human,jain2016structural,gui2018few,fragkiadaki2015recurrent,gui2018adversarial} using the MAE metric in Euler space. 
To have a fair comparison, we generate one motion per observation by setting the latent variable to the distribution mode, i.e., $z=\mu_c$. This allows us to generate a plausible motion without having access to the ground truth. To compare against the deterministic baselines, we follow the standard setting, and thus use 50 frames (i.e., 2sec) as observation to generate the next 25 frames (i.e., 1sec). Surprisingly, despite having a very simple motion decoder architecture (one-layer GRU network) with a very simple reconstruction loss function (MSE), this motion-from-mode strategy yields results that are competitive with those of the baselines that use sophisticated architectures and advanced loss functions. We argue that learning a good, context-preserving latent representation of human motion is the contributing factor to the success of our approach. This, however, could be used in conjunction with sophisticated motion decoders and reconstruction losses, which we leave for future research.

In Appendix A, we study alternative designs to condition the VAE encoder and decoder.

\subsection{Diverse Image Captioning}
For the task of conditional text generation, we focus on stochastic image captioning. To demonstrate the effectiveness of our approach, we report results on the MSCOCO~\citep{lin2014microsoft} captioning task, with the original train/test splits of 83K and 41K images, respectively. The MSCOCO dataset has five captions per image. However, we make it deterministic by removing four captions per image, yielding a Deterministic-MSCOCO captioning dataset. Note that the goal of this experiment is  not to advance the state of the art in image captioning, but rather to explore the effectiveness of our approach on a different task
where we have strong conditioning signal and an expressive decoder in the presence of a deterministic dataset. 
A brief review of the recent work on diverse text generation is given in Appendix A.

\begin{figure}[t]
    \centering
    \scriptsize
    \scalebox{0.7}{
    \begin{tabular}{c @{ } @{ } @{ }c @{ } @{ } @{ }c}
    Image & Groundtruth and Baselines & \texttt{LCP-VAE}\\
    \midrule
    \begin{minipage}{.3\linewidth} 
    \includegraphics[width=\textwidth]{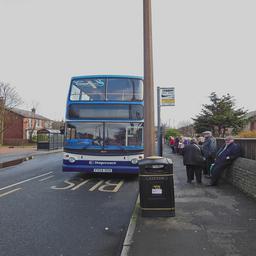}
    \end{minipage}
    
    &
    \begin{minipage}{.45\linewidth} 
    \textbf{Ground-truth Captions}: \\
    {1.A big blue two level bus at a bus stop with people.}\\
    {2.a blue and white double decker bus parked at a bus stop}\\
    {3.A blue bus s parked on a curb where people are standing.}\\
    {4.A blue double decker bus parked on the side of a road.}\\
    {5.A bus pulls up to a station where people wait.}\\
    \textbf{Caption from Autoregressive}: \\
 {a blue bus is parked on the side of the road . }\\
    \textbf{Caption from Conditional VAE}: \\
{a bus is parked on the side of the road .} \\
    \end{minipage}
    
    &
    \begin{minipage}{.56\linewidth} 
    \textbf{Captions from \texttt{LCP-VAE}}:\\
    {1. a large bus is going down the street next to a huge crowd of people .} \\
    {2. a double decker bus is driving down the street .} \\
    {3. a blue and white bus on a street next to a building .} \\
    {4. a bus is driving down the street with a car on the opposite side of the road .} \\
    {5. a bus is driving down the street with a group of people on the side of it .} \\
    {6. a street with a lot of people on it and a street sign on it .}\\
    {7. a group of people standing on a street near a bus .}\\
    {8. a street sign on a pole with a bus in the background .} \\
    \textbf{Caption from the \texttt{LCP-VAE} mode}: \\
    {a double decker bus is driving down the street in pursuit of passengers .}
    \end{minipage}
\end{tabular}
}
    \caption{Qualitative evaluation of the diversity in generated captions. While captions generated by our approach are diverse, they all describe the image properly. The caption from the mode also typically provides an accurate description.}
    \label{fig:qualitative_captioning5}
\end{figure}{}

\begin{table}[t]
    \centering
    \scriptsize
    \setlength\extrarowheight{-3pt}

    \caption{Quantitative evaluation of stochastic image captioning on the MSCOCO Captioning dataset.}
    \label{tab:captioning}
    \scalebox{0.9}{
    \begin{tabular}{l c c c c c c}
    \toprule
         Model & ELBO (KL) & Perplexity & Quality & Diversity & Context & Training KL\\
         & \scriptsize{(Reconstructed)} & \scriptsize{(Reconstructed)} & \scriptsize{(Sampled)} & \scriptsize{(Sampled)} & \scriptsize{(Sampled)} & \scriptsize{(Reconstructed)} \\
         \midrule
         Autoregressive & 3.01 (N/A) & 20.29 & 0.40 & N/A & 0.46 & N/A\\
         Conditional VAE & 2.86 (0.00) & 17.46 & 0.39 & 0.00 & 0.44 & 0.00 \\
         \texttt{LCP-VAE} & 0.21 (3.28) & 1.23 & 0.40 & 0.53 & 0.43 & 3.11 \\
         \bottomrule
    \end{tabular}
    }
    \vspace{-5pt}
\end{table}{}

We compare \texttt{LCP-VAE} (with the architecture described in Appendix A) with a standard CVAE and with its autoregressive, non-variational counterpart\footnote{Note that \texttt{LCP-VAE} is agnostic to the choice of data encoder/decoder architecture. We leave the use of more sophisticated architectures for future research.}. 
For quantitative evaluation, we report the ELBO (the negative log-likelihood), along with the KL-divergence and the Perplexity of the reconstructed captions on the held-out test set. We also quantitatively measure the diversity, the quality, and the context of sampled captions. To measure the context, we rely on the BLEU1 score, making sure that the sampled captions represent  elements that appear in the image. For CVAE and \texttt{LCP-VAE}, we compute the average BLEU1 score for $K=50$ captions sampled per image and report the mean over the images. To measure diversity, we compte the BLEU4 score between every pair of $K=50$ sampled captions per image. The smaller the BLEU4 is, the more diverse the captions are. The diversity metric is then 1-BLEU4, i.e., the higher the better. To measure quality, 
we use a metric similar to that in our human motion prediction experiments, obtained by training a binary classifier to discriminate real (ground-truth) captions from fake (generated) ones. The accuracy of this classifier on the test set is inversely proportional to the quality of the generated captions. 
We provide qualitative examples for all the methods in Fig.~\ref{fig:qualitative_captioning5} and in Appendix A. As shown in Table~\ref{tab:captioning}, a CVAE learns to ignore the latent variable as it can minimize the caption reconstruction loss given solely the image representation. 
By doing so, all the generated captions at test time are identical, despite sampling multiple latent variables. This can be further seen in the ELBO and Perplexity of the reconstructed captions. We expect a model that takes as input the captions and the image to have a much lower reconstruction loss compared to the autoregressive baseline, which takes only the image as input. However, this is not the case with CVAE, indicating that the connection between the encoder and the decoder, i.e., the latent variable, does not carry essential information about the input caption. However, the quality of the generated sample is reasonably good. This is also evidenced by the qualitative evaluations in Appendix A. By contrast, \texttt{LCP-VAE} effectively handles this situation by unifying the sampling of the latent variable and the conditioning, leading to diverse but high quality captions, as reflected by the ELBO of our approach in Table~\ref{tab:captioning} and the qualitative results in Appendix A. Additional quantitative evaluations and ablation studies for image captioning are provided in Appendix A.

\section{Conclusion}
In this chapter, we have studied the problem of conditionally generating diverse sequences  
with a focus on scenarios where the conditioning signal is strong enough such that an expressive decoder can generate plausible samples from it only. 
We have addressed this problem by forcing the sampling of the latent variable to depend on the conditioning one, which contrasts with standard CVAEs. By making this dependency explicit, the model receives a latent variable that carries information about the condition during both training and test time. This further prevents the network from ignoring the latent variable in the presence of a strong condition, thus enabling it to generate diverse outputs. 
To demonstrate the effectiveness of our approach, we have investigated two application domains: Stochastic human motion prediction and diverse image captioning. In both cases, our \texttt{LCP-VAE} model was able to generate diverse and plausible samples, as well as to retain contextual information, leading to semantically-meaningful predictions. 
We believe our approach will have great impact on practical applications such as pedestrian intention forecasting~\cite{aliakbarian2018viena2}, trajectory forecasting and human tracking~\cite{saleh2020artist,kosaraju2019social}.
In the future, we will also apply our approach to other problems that rely on strong conditions, such as image inpainting and super-resolution, for which only deterministic datasets are available.

%% file: conclusion.tex
\chapter{Conclusion}
\label{cha:conc}
In this thesis we have tackled the problem video anticipation by learning sequential representations. We have started from anticipating a discrete representation of a deterministic future, e.g., as in action anticipation and then move toward a more complex task of anticipating multiple plausible continuous representations of a stochastic task, e.g., human motion prediction.

The major contributions of this thesis are outlined below:
\begin{enumerate}
    \item We propose a novel action anticipation framework (that can be also seen as an \textit{early recognition} model). In particular, we introduce a novel loss that encourages making correct predictions very early. Our loss models the intuition that some actions, such as running and high jump, are highly ambiguous after seeing only the first few frames, and false positives should therefore not be penalized too strongly in the early stages. By contrast, we would like to predict a high probability for the correct class as early as possible, and thus penalize false negatives from the beginning of the sequence. Our experiments demonstrate that, for a given model, our new loss yields significantly higher accuracy than existing ones on the task of early prediction. We also propose a novel multi-stage Long Short Term Memory (MS-LSTM) architecture for action anticipation. This model effectively extracts and jointly exploits context- and action-aware features. This is in contrast to existing methods that typically extract either global representations for the entire image or video sequence, thus not focusing on the action itself, or localize the feature extraction process to the action itself via dense trajectories, optical flow or actionness, thus failing to exploit contextual information. This work has been published in ICCV 2017, Venice, Italy.
    
    \item We improve and extend our previous contribution by focusing on driving scenarios, encompassing common the subproblems of anticipating ego car's driver maneuvers, front car's driver maneuver, accidents, violating or respecting traffic rules, and pedestrian intention, with a fixed, sensible set of sensors. To this end, we introduce the VIrtual ENvironment  for Action Analysis  (VIENA$^2$) dataset. Altogether, these subproblems encompass a total of 25 distinct action classes. VIENA$^2$ is acquired using the GTA V video game. It contains more than 15K full HD, 5s long videos, corresponding to more than 600 samples per action class, acquired in various driving conditions, weathers, daytimes, and environments. This amounts to more than 2.25M frames, each annotated with an action label. These videos are complemented by basic vehicle dynamics measurements, reflecting well the type of information that one could have access to in practice\footnote{Our dataset is publicly available at \url{https://sites.google.com/view/viena2-project/}}. We then benchmark state-of-the-art action anticipation algorithms on VIENA2, and as another contribution, introduce a new multi-modal, LSTM-based  architecture that  generalizes  out previous contribution to an arbitrary number of modalities,  together  with  a  new  anticipation  loss,  which  outperforms existing approaches in our driving anticipation scenarios. This work has been published in ACCV 2018, Perth, Australia.

    \item For continuous, stochastic anticipation task, we address the problem of stochastic human motion prediction. As introduced earlier in this thesis, human motion prediction aims to forecast the sequence of future poses of a person given past observations of such poses. To achieve this, existing methods typically rely on recurrent neural networks (RNNs) that encode the person’s motion. While they predict reasonable motions, RNNs are deterministic models and thus cannot account for the highly stochastic nature of human motion; given the beginning of a sequence, multiple, diverse futures are plausible. To correctly model this, it is therefore critical to develop algorithms that can learn the multiple modes of human motion, even when presented with only deterministic training samples. We introduce an approach to effectively learn the stochasticity in human motion. At the heart of our approach lies the idea of Mix-and-Match perturbations: Instead of combining a noise vector with the conditioning variables in a deterministic manner (as usually done in standard practices), we randomly select and perturb a subset of these variables. By randomly changing this subset at every iteration, our strategy prevents training from identifying the root of variations and forces the model to take it into account in the generation process. This is a highly effective conditioning scheme in scenarios when (1) we are dealing with a deterministic dataset, i.e., one sample per condition, (2) the conditioning signal is very strong and representative, e.g., the sequence of past observations, and (3) the model has an expressive decoder that can generate a plausible sample given only the condition. We utilize Mix-and-Match by incorporating it into a recurrent encoder-decoder network with a conditional variational autoencoder (CVAE) block that learns to exploit the perturbations. Mix-and-Match then acts as the stochastic conditioning scheme instead of concatenation that usually appears in standard CVAEs. This work has been published in CVPR 2020, Seattle, Washington, USA.

    \item In our previous contribution, we identified one limitation of a standard CVAEs when dealing deterministic datasets and strong conditioning signals. In this work, we further investigates this problem from a more theoretical point of view. Specifically, in this contribution we tackle the task of diverse sequence generation in which all the diversely generated sequences carry the same semantic as in the conditioning signal. We observe that in standard CVAE, conditioning and sampling the latent variable are two independent processes, leading to generating samples that are not necessarily carry all the contextual information about the condition. To address this, we propose to explicitly make the latent variables depend on the observations (the conditioning signal). To achieve this, we develop a CVAE architecture that learns a distribution not only of the latent variables, but also of the observations, the latter acting as prior on the former. By doing so, we change the variational family of the posterior distribution of the CVAE, thus, as a side effect, our approach can mitigate posterior collapse to some extent. This work will be submitted to CVPR 2021.
\end{enumerate}

\section{Future Work}
\label{sec:future}
In this thesis we focused on the task of video anticipation. Although there has been great progress in recent years to address the task of action anticipation, diverse human motion prediction remained a relatively less studied problem. Below, we list some potential future direction based on our research, mostly focusing on diverse sequence generation:
\begin{itemize}
    \item Incorporating sequential latent variables: In our proposed frameworks, we use a single latent variable to represent a sequence, e.g., human motion or an image caption. However, a more expressive alternative could be a sequential latent variable, considering a latent variable for the representations at each time-step. Specifically, instead of having $z$ sampled from the approximate posterior $\mathcal{N}(\mu_x, \text{diag}(\sigma_x^2))$, we consider a sequence of latent variables $z_{1:T}$ where $z_i$ is sampled from $\mathcal{N}(\mu_{i_x}, \text{diag}(\sigma_{i_x}^2))$ and given $z_{<i}$. We believe that this not only provides a more expressive latent space, it allows us to have more control on the variations at different time-steps.
    \item Incorporating Auxiliary information: Our proposed methods for diverse human motion prediction are completely unsupervised; no extra annotation/feedback are required to train our models. However, one can certainly benefit from additional information to guide the latent space to learn better (more semantically meaningful) representations. For instance, in the case of human motion prediction, the motion capture datasets are often come with additional action label; the action that the subject is performing while his/her motion is being captured. One can utilize this information with a shallow action classifier on top of the latent space, forcing the latent space to learn not only the information that is useful for motion reconstruction, but also the ones that are discriminative enough to distinguish multiple actions. 
    \item Going beyond poses: Our motion prediction approaches works on abstract pose representation of the human at each time-step. That is a representation of e.g., 32 body joints represented in either 3D position or 4D Quaternion representation. One potential future direction is to evaluate how our methods perform when dealing with very high dimensional data, e.g., video frames or fine-grained human shapes.
\end{itemize}

%% file: appendix.tex
\appendix
\chapter{Appendix A}

\section{Technical Background on Evidence Lower Bound}
\label{appendix:elbo}
Our goal is to solve a maximum likelihood problem.
To this end, as discussed in the main paper, we rely on \emph{Variational Inference}, which approximates the true posterior $p_\theta(z|x)$ with another distribution $q_\phi(z|x)$. This distribution is computed via another neural network parameterized by $\phi$ (called variational parameters), such that $q_\phi(z|x)\simeq p_\theta(z|x)$. Using such an approximation, \emph{Variational Autoencoders}~\citep{kingma2013auto}, or VAEs in short, are able to optimize the marginal likelihood in a tractable way. The optimization objective of the VAEs is a variational lower bound, also known as evidence lower bound, or ELBO in short. 

Specifically, to find an approximation of the posterior that represents the true one, variational inference minimizes the Kullback-Leibler (KL) divergence between the approximate and true posteriors. This divergence can be written as
\begin{align}
    \mathcal{D}_{KL}\Big[q_\phi(z|x) || p_\theta(z|x)\Big] = \sum_{z\sim q_\phi(z|x)}q_\phi(z|x) \log \frac{q_\phi(z|x)}{p_\theta(z|x)}\;.
\end{align}{}
This can further be seen as an expectation, yielding
\begin{align}
    \mathcal{D}_{KL}\big[q_\phi(z|x) || p_\theta(z|x)\big] = \mathbb{E}_{z\sim q_\phi(z|x)} \bigg[\log \frac{q_\phi(z|x)}{p_\theta(z|x)}\bigg] \nonumber \\ =  \mathbf{E}_{z\sim q_\phi(z|x)} \Big[ \log q_\phi(z|x) - \log p_\theta(z|x) \Big]\;.
\end{align}{}
According to Bayes' theorem, the second term above, i.e., the true posterior, can be written as $p_\theta(z|x) = \frac{p_\theta(x|z)p(z)}{p_\theta(x)}$.
The data distribution $p_\theta(x)$ is independent of the latent variable $z$, and can thus be pulled out of the expectation term, giving
{\small
\begin{align}
    \mathcal{D}_{KL}\big[q_\phi(z|x) || p_\theta(z|x)\big] = \mathbb{E}_{z\sim q_\phi(z|x)} \Big[\log q_\phi(z|x) - \log p_\theta(x|z) - \log p(z) \Big] + \log p_\theta(x)\;.
\end{align}{}
}
By moving the $\log p_\theta(x)$ term to the right-hand side of the above equation, we can write
{\small
\begin{align}
    \mathcal{D}_{KL}\big[q_\phi(z|x) || p_\theta(z|x)\big] - \log p_\theta(x)= \mathbb{E}_{z\sim q_\phi(z|x)} \Big[\log q_\phi(z|x) - \log p_\theta(x|z) - \log p(z) \Big] \nonumber \\
    \log p_\theta(x) - \mathcal{D}_{KL}\big[q_\phi(z|x) || p_\theta(z|x)\big] = \mathbb{E}_{z\sim q_\phi(z|x)} \Big[\log p_\theta(x|z) - \big(\log q_\phi(z|x) - \log p(z)\big) \Big] \nonumber \\
    = \mathbb{E}_{z\sim q_\phi(z|x)} \Big[\log p_\theta(x|z) \Big] - \mathbb{E}_{z\sim q_\phi(z|x)} \Big[\log q_\phi(z|x) - \log p(z) \Big]\;.
\end{align}{}
}The second expectation term in the resulting equation is, by definition, the KL divergence between the approximate posterior $q_\phi(z|x)$ and the prior $p(z)$. This lets us write
\begin{align}
    \log p_\theta(x) - D_{KL}\big[q_\phi(z|x) || p_\theta(z|x)\big] = \mathbb{E}_{z\sim q_\phi(z|x)} \big[\log p_\theta(x|z) \big] - \mathcal{D}_{KL}\big[q_\phi(z|x) || p(z)\big].
\end{align}{}
In this equation, $\log p_\theta(x)$ is the log-likelihood of the data, which we would like to maximize; $D_{KL}\big[q_\phi(z|x) || p_\theta(z|x)\big]$ is the KL divergence between the approximate and the true posterior distributions, which, while not computable, is by definition non-negative; $\mathbf{E}_{z\sim q_\phi(z|x)} \big[\log p_\theta(x|z) \big]$ is the reconstruction loss; and $D_{KL}\big[q_\phi(z|x) || p(z)\big]$ is the KL divergence between the approximate posterior distribution and a prior over the latent variable. This last term can be seen as a regularizer of the latent representation. Altogether, the intractability and non-negativity of $D_{KL}\big[q_\phi(z|x) || p_\theta(z|x)\big]$ only allows us to optimize a lower bound of the log-likelihood of the data, given by
\begin{align}
\log p_\theta(x) \geq \mathbb{E}_{z\sim q_\phi(z|x)} \big[\log p_\theta(x|z) \big] - \mathcal{D}_{KL}\big[q_\phi(z|x) || p(z)\big]\;.
\end{align}{}
This is referred to as variational or evidence lower bound (ELBO).

\section{Derivation of LCP-VAE's KL Divergence Loss}
\label{appendix:KL}
In our approach, the model encourages the posterior of \texttt{LCP-VAE} to be close to the one of \texttt{CS-VAE}. In general, the KL divergence between two distributions $P_1$ and $P_2$ is defined as
\begin{align}
    \mathcal{D}_{KL}(P_1 || P_2) = \mathbb{E}_{P_1} \bigg[\log\frac{P_1}{P_2}\bigg]\;.
\end{align}
Let us now consider the case where the distributions are multivariate Gaussians $\mathcal{N}(\mu, \Sigma)$  in $\mathbb{R}^d$, where $\Sigma=\text{diag}(\sigma^2)$, with $\sigma$ and $\mu$ are $d$-dimensional vectors predicted by the encoder network of the VAE. The density function of such a distribution is 
\begin{align}
    p(x) = \frac{1}{(2\pi)^{\frac{d}{2}} det(\Sigma)^{\frac{1}{2}}} exp \bigg(-\frac{1}{2}(x-\mu)^T\Sigma^{-1}(x-\mu)\bigg)\;.
\end{align}{}
Thus, the KL divergence between two multivariate Gaussians is computed as
{\small
\begin{flalign}
    & \mathcal{D}_{KL}(P_1 || P_2) 
    \nonumber &\\
    & = \frac{1}{2} \mathbb{E}_{P_1}\bigg[-\log \det \Sigma_1 - (x-\mu_1)^T\Sigma_1^{-1}(x-\mu_1) + \log \det\Sigma_2+(x-\mu_2)^T\Sigma_2^{-1}(x-\mu_2)\bigg]
    \nonumber & \\
    & = \frac{1}{2}\log\frac{\det\Sigma_2}{\det\Sigma_1} + \frac{1}{2} \mathbb{E}_{P_1}\bigg[ - (x-\mu_1)^T\Sigma_1^{-1}(x-\mu_1) +(x-\mu_2)^T\Sigma_2^{-1}(x-\mu_2)\bigg]
    \nonumber & \\
    & = \frac{1}{2}\log\frac{\det\Sigma_2}{\det\Sigma_1} + \frac{1}{2} \mathbb{E}_{P_1}\bigg[-tr\{\Sigma_1^{-1}(x-\mu_1)(x-\mu_1)^T\} + tr\{\Sigma_2^{-1}(x-\mu_2)(x-\mu_2)^T\}\bigg]
    \nonumber & \\
    & = \frac{1}{2}\log\frac{\det\Sigma_2}{\det\Sigma_1} + \frac{1}{2} \mathbb{E}_{P_1}\bigg[-tr\{\Sigma_1^{-1}\Sigma_1\} +tr\{\Sigma_2^{-1}(xx^T-2x\mu^T_2+\mu_2\mu_2^T)\}\bigg]
    \nonumber & \\
    & = \frac{1}{2}\log\frac{\det\Sigma_2}{\det\Sigma_1} - \frac{1}{2}d + \frac{1}{2}tr\{\Sigma_2^{-1}(\Sigma_1+\mu_1\mu_1^T-2\mu_2\mu_1^T+\mu_2\mu_2^T)\}
    \nonumber & \\
    & = \frac{1}{2}\Bigg[\log\frac{\det\Sigma_2}{\det\Sigma_1} - d + tr\{\Sigma_2^{-1}\Sigma_1\}+tr\{\mu_1^T\Sigma_2^{-1}\mu_1 - 2\mu_1^T\Sigma_2^{-1}\mu_2 + \mu_2^T\Sigma_2^{-1}\mu_2\}\Bigg]
    \nonumber & \\
    & = \frac{1}{2}\Bigg[\log\frac{|\Sigma_2|}{|\Sigma_1|}-d+tr\{\Sigma_2^{-1}\Sigma_1\} + (\mu_2 - \mu_1)^T\Sigma_2^{-1}(\mu_2 - \mu_1)\Bigg]\;.
    \label{eq:kl}
\end{flalign}
}
where $tr\{\cdot\}$ denotes the trace operator. In Eq.~\ref{eq:kl}, the covariance matrix $\Sigma_1$ and mean $\mu_1$ correspond to distribution $P_1$ and the covariance matrix $\Sigma_2$ and mean $\mu_2$ correspond to distribution $P_2$. 
Given 
this result, we can then compute the KL divergence of the \texttt{LCP-VAE} and the posterior distribution with mean  $\mu + \sigma\odot\mu_c$ and covariance matrix $\text{diag}((\sigma\odot\sigma_c)^2)$. Let $\Sigma=\text{diag}(\sigma^2)$,  $\Sigma_c=\text{diag}(\sigma_c^2)$, and $d$ be the dimensionality of the latent space. The loss in Eq. 6 of the main paper can then be written as

\begin{align}
    \mathcal{L}_{prior}^{\texttt{LCP-VAE}} = -\frac{1}{2}\Big[\log\frac{|\Sigma_c|}{|\Sigma_c||\Sigma|}-d+
    tr\{\Sigma_c^{-1}\Sigma_c\Sigma\}+ \nonumber \\(\mu_c-(\mu+\Sigma\mu_c))^T\Sigma_c^{-1}(\mu_c-(\mu+\Sigma\mu_c))\Big]\;.
\end{align}

Since $\Sigma_c^{-1}\Sigma_c=I$, $|\Sigma_c|$ will be cancelled out in the $\log$ term, which yields
\begin{align}
    \mathcal{L}_{prior}^{\texttt{LCP-VAE}} = -\frac{1}{2}\Big[\log\frac{1}{|\Sigma|}-d+
    tr\{\Sigma\} + (\mu_c-(\mu+\Sigma\mu_c))^T\Sigma_c^{-1}(\mu_c-(\mu+\Sigma\mu_c))\Big]\;. 
\end{align}

\section{Mitigating Posterior Collapse: Related Work}
\label{appendix:posterior_related}
Deep generative models offer promising results in generating diverse, realistic samples, such as images, text, motion, and sound, from purely unlabeled data. One example of such successful generative models are variational autoencoders~\citep{kingma2013auto} (VAEs), the stochastic variant of autoencoders,
which, thanks to strong and expressive decoders, can generate high-quality samples. Training such models, however, may often result in posterior collapse: the posterior distribution $q(z|x)$ of the latent variable $z$ given the input $x$ becomes equal to the prior distribution, resulting in a latent variable carrying no information about the input. In other words, the model learns to ignore the latent variable.

The most common approaches to tackling posterior collapse consist of weighing the KL divergence between the posterior and prior during training by an annealing function~\citep{bowman2015generating,yang2017improved,kim2018semi,gulrajani2016pixelvae,liu2019cyclical}, weakening the decoder~\citep{semeniuta2017hybrid,zhao2017infovae}, or changing the training objective~\citep{zhao2017infovae,tolstikhin2017wasserstein}. All of them are based on the perspective that the solution to posterior collapse can be found in a good local optimum in terms of evidence lower bound~\citep{chen2016variational,alemi2017fixing}. However, they each suffer from drawbacks: Any annealing weight that does not become and remain equal to one at some point during training yields an improper statistical model; 
weakening the decoder tends to degrade the quality of the generated samples; changing the objective does not optimize the true variational lower bound.
As an alternative, some methods modify the training strategy to more strongly encourage the inference network to  approximate the model's true posterior~\citep{he2019lagging,li2019surprisingly}.
Other methods add auxiliary tasks either with non-autoregressive models~\citep{lucas2018auxiliary} or that exploit the latent variable~\citep{goyal2017z,lucas2018auxiliary,dieng2018avoiding}. While this encourages the latent variable to carry some information, it may not be directly useful for the main task. 
Alternatively, several techniques incorporate constraints in VAEs. In this context, VQ-VAE~\citep{van2017neural} introduces a discrete latent variable obtained by vector quantization of the latent one that, given a uniform prior over the outcome, yields a fixed KL divergence equal to log $K$, with $K$ the size of the codebook; several recent  works  use  the von~Mises-Fisher  distribution  to  obtain  a  fixed  KL  divergence, thus  mitigating  the  posterior collapse problem~\citep{guu2018generating,xu2018spherical,davidson2018hyperspherical}; more recently, delta-VAE~\citep{razavi2019preventing} modifies the posterior such that it maintains a minimum distance between the prior and the posterior. 

Although successful at handling the posterior collapse in the presence of expressive decoders, e.g., LSTM, GRUs, PixelCNN, all of these approaches were designed for standard VAEs, not \emph{conditional} VAEs (CVAEs). 
As such, they do not address the problem of mitigating the influence of a strong conditioning signal in ignoring the latent variable, which is our focus here. In this context, we observed that a strong condition provides enough information for an expressive decoder to reconstruct the data, thus allowing the decoder to ignore the latent variable at no loss in reconstruction quality. 

\section{Further Discussion on the Performance of Stochastic Baselines}
\label{appendix:motionbaselines}
The MT-VAE model~\citep{yan2018mt} tends to ignore the random variable $z$, thus ignoring the root of variation. As a consequence, it achieves a low diversity, much lower than ours, but produces samples of high quality, albeit almost identical (see the diversity of the generated motions in quantitative evaluations). We empirically observed this by analysing the weights acting on the latent variable $z$ and the ones acting on the conditioning signal (i.e., the hidden state of the past motion). We observed the magnitude of the weights acting on $z$ to be orders of magnitude  smaller than that of acting on the condition, 0.008 versus 232.85, respectively. To further confirm that the MT-VAE ignores the latent variable, we performed an additional experiment where, at test time, we sampled each element of the random vector independently from $\mathcal{N}(50, 50)$ instead of from the prior $\mathcal{N}(0, I)$. This led to neither loss of quality nor increase of diversity of the generated motions. 

Our experiments with the HP-GAN model~\citep{barsoum2018hp} evidence the limited diversity of the sampled motions despite its use of random noise during inference. Note that the authors of~\citep{barsoum2018hp} mentioned in their paper that the random noise was added to the hidden state. Only by studying their publicly available code\footnote{\texttt{https://github.com/ebarsoum/hpgan}} did we understand the precise way this combination was done. In fact, the addition relies on a parametric, linear transformation of the noise vector. That is, the perturbed hidden state is obtained as
\begin{align}
    h_{perturbed} = h_{original} + W^{z\rightarrow h} z\;.
    \label{eq:whz}
\end{align}
Because the parameters $W^{z\rightarrow h}$ are \emph{learned}, the model has the flexibility to ignore $z$, which leads to the low diversity of sampled motions. Note that the authors of~\citep{barsoum2018hp} acknowledged that, despite their best efforts, they noticed very little variation between predictions obtained with different $z$ values. We further analysed this phenomenon and observed that the magnitude of $W^{z\rightarrow h}$ is in the order of $O(1e^{-3})$, confirming the fact that it provides the model with the flexibility to ignore $z$. Since the perturbation is ignored, however, the quality of the generated motions is high. 

Pose-Knows~\citep{walker2017pose}, produces motions with higher diversity than the aforementioned two baselines, but of much lower quality. The main reason behind this is that the random vectors that are concatenated to the poses at each time-step are sampled independently of each other, which translates to discontinuities in the generated motions. This problem might be mitigated by sampling the noise in a time-dependent, autoregressive manner, as in~\citep{videoFlow} for video generation. Doing so, however, goes beyond the scope of our analysis. 

The Mix-and-Match approach~\citep{aliakbarian2019MixAndMatch} yields sampled motions with higher diversity and reasonable quality. The architecture of Mix-and-Match is very close to that of MT-VAE, but replaces the deterministic concatenation operation with a stochastic perturbation of the hidden state with the noise. Through such a perturbation, the decoder is not able decouple the noise and the condition, by contrast with concatenation. However, since the perturbation is not learned and is a non-parametric operation, the quality of the generated motion is lower than ours and of other baselines (except for Pose-Knows). We see the Mix-and-Match perturbation as a workaround to the posterior collapse problem, which nonetheless sacrifices the quality and the context of the sampled motions. 

\section{Ablation Study on Different Means of Conditioning}
\label{appendix:motionablation}
In addition to the experiments in the main paper, we also study various designs to condition the VAE encoder and decoder. As discussed before, conditioning the VAE encoder can be safely done via concatenating two deterministic sources of information, i.e., the representations of the past and the future, since both sources are useful to compress the future motion into the latent space. In Table~\ref{tab:arch_design}, we use both a deterministic representation of the observation, $h_t$, and a stochastic one, $z_c$, as a conditioning variable for the encoder. Similarly, we compare the use of either of these variables via concatenation with that of our modified reparameterization trick (explained in the main paper). This shows that, to condition the decoder, reparameterization is highly effective at addressing posterior collapse.  Furthermore, for the encoder, a deterministic condition works better than a stochastic one. When both the encoder and decoder are conditioned via deterministic conditioning variables, i.e., row 2 in Table~\ref{tab:arch_design}, the model learns to ignore the latent variable and rely solely on the condition, as evidenced by the KL term tending to zero.
\begin{table}[!t]
    \centering
    \small
     
    \caption{Evaluation of various architecture designs for a CVAE. A smaller KL value, indicating posterior collapse, leads to less diversity.}
   \tabcolsep=0.20cm
   
    \begin{tabular}{l l c}
    \toprule
        Encoder Conditioning & Decoder Conditioning & \texttt{LCP-VAE}'s Training KL\\
        \midrule
        Concatenation ($z_c$) & Reparameterization ($z_c$) & 6.92\\
        Concatenation ($h_t$) & Concatenation ($h_t$) & 0.04\\
        Concatenation ($z_c$) & Concatenation ($z_c$) & 0.61\\
        
        Concatenation ($h_t$) & Reparameterization ($z_c$) & 8.07 \\
        \bottomrule
    \end{tabular}
    \label{tab:arch_design}
\end{table}{}

\section{Experimental Results on the Penn Action Dataset}
\label{appendix:penn}
As a complementary experiment, we evaluate our approach on the Penn Action dataset, which contains 2326 sequences of 15 different actions, where for each person, 13 joints are annotated in 2D space. Most sequences have less than 50 frames and the task is to generate the next 35 frames given the first 15. Results are provided in Table~\ref{tab:penn}. Note that the upper bound for the Context metric is 0.74, i.e., the classification performance given the Penn Action ground-truth motions.

\begin{table}[!t]
    \centering
    \caption{Quantitative evaluation on the Penn Action dataset. Note that a diversity of 1.21 is reasonably high for normalized 2D joint positions, i.e., values between 0 and 1, normalized with the width and the height of the image.}
    \begin{tabular}{l c c c c c}
    \toprule
     & ELBO (KL) & Diversity & Quality & Context & Training KL \\
      Method & \scriptsize{(Reconstructed)} & \scriptsize{(Sampled)} & \scriptsize{(Sampled)} & \scriptsize{(Sampled)} & \scriptsize{(Reconstructed)} \\
     \midrule
        \texttt{LCP-VAE} &  0.034 (6.07) & 1.21 & 0.46 & 0.70 & 4.84\\
        Autoregressive Counterpart & 0.048 (N/A) & 0.00 & 0.46 & 0.51 & N/A\\
        \bottomrule
    \end{tabular}
    \label{tab:penn}
\end{table}{}

\section{Stochastic Human Motion Prediction Architecture}
\label{appendix:motionarch}
Our motion prediction model follows the architecture depicted in Fig. 2 of the main paper. Below, we describe the architecture of each component in our model. Note that human poses, consisting of 32 joints in case of the Human3.6M dataset, are represented in 4D quaternion space. Thus, each pose at each time-step is represented with a vector of size $1\times 128$. All the tensor sizes described below ignore the mini-batch dimension for simplicity. 

The \textbf{observed motion encoder}, or \texttt{CS-VAE} motion encoder, is a single layer GRU~\citep{chung2014empirical} network with 1024 hidden units. If the observation sequence has length $T_{obs}$, the observed motion encoder maps $T_{obs}\times 128$ into a single hidden representation of size $1\times 1024$, i.e., the hidden state of the last time-step. This hidden state, $h_t$, acts as the condition to the \texttt{LCP-VAE} encoder and the direct input to the \texttt{CS-VAE} encoder.

\textbf{\texttt{CS-VAE}}, similarly to any variational autoencoder, has an encoder and a decoder. The \texttt{CS-VAE} encoder is a fully-connected network with ReLU non-linearities, mapping the hidden state of the motion encoder, i.e., $h_t$, to an embedding of size $1\times 512$. Then, to generate the mean and standard deviation vectors, we use two fully connected branches. They map the embedding of size $1\times 512$ to a mean vector  of size $1\times 128$ and a standard deviation vector of size $1\times 128$, where 128 is the length of the latent variable. Note that we apply a ReLU non-linearity to the vector of standard deviations to ensure that it is non-negative. We then use the reparameterization trick~\citep{kingma2013auto} to sample a latent variable of size $1\times 128$. The \texttt{CS-VAE} decoder consists of multiple fully-connected layers, mapping the latent variable to a variable of size $1\times 1024$, acting as the initial hidden state of the observed motion decoder. Note that we apply a Tanh non-linearity to the generated hidden state to mimic the properties of a GRU hidden state.

The \textbf{observed motion decoder}, or \texttt{CS-VAE} motion decoder, is similar to its motion encoder, except for the fact that it reconstructs the motion autoregressively. Additionally, it is initialized with the reconstructed hidden state, i.e., the output of the \texttt{CS-VAE} decoder. The output of each GRU cell at each time-step is then fed to a fully-connected layer, mapping the GRU output to a vector of size $1\times 128$, which represents a human pose with 32 joints in 4D quaternion space. To decode the motions, we use a teacher forcing technique~\citep{williams1989learning} during training. At each time-step, the network chooses with probability $P_{tf}$ whether to use its own output at the previous time-step or the ground-truth pose as input. We initialize $P_{tf}=1$, and decrease it linearly at each training epoch such that, after a certain number of epochs, the model becomes completely autoregressive, i.e., uses only its own output as input to the next time-step. Note that, at test time, the motions are generated completely autoregressively, i.e., with $P_{tf}=0$.

Note that the future motion encoder and decoder have exactly the same architectures as the observed motion ones. The only difference is their input, where the future motion is represented by poses from $T_{obs}$ to $T_{end}$ in a sequence. In the following, we describe the architecture of \texttt{LCP-VAE} for motion prediction.

\textbf{\texttt{LCP-VAE}} is a conditional variational encoder. Its encoder's input is a representation of future motion, i.e., the last hidden state of the future motion encoder, $h_T$, conditioned on $h_t$. The conditioning is done by concatenation, thus the input to the encoder is a representation of size $1\times 2048$. The \texttt{LCP-VAE} encoder, similarly to the \texttt{CS-VAE} encoder, maps its input representation to an embedding of size $1\times 512$. Then, to generate the mean and standard deviation vectors, we use two fully connected branches, mapping the embedding of size $1\times 512$ to a mean vector of size $1\times 128$ and a standard deviation vector of size $1\times 128$, where 128 is the length of the latent variable. Note that we apply a ReLU non-linearity to the vector of standard deviations to ensure that it is non-negative. To sample the latent variable, we use our extended reparameterization trick, explained in the main paper. This unifies the conditioning and sampling of the latent variable. Then, similarly to \texttt{CS-VAE}, the latent variable is fed to the \texttt{LCP-VAE} decoder, which is a fully connected network that maps the latent representation of size $1\times 128$ to a reconstructed hidden state of size $1\times 1024$ for future motion prediction. Note that we apply a Tanh non-linearity to the generated hidden state to mimic the properties of a GRU hidden state.

\section{Diverse Image Captioning Architecture}
\label{appendix:textarch}
Our diverse image captioning model follows the architecture depicted in Fig.2 of the main paper. Below, we describe the architecture of each component in our model. Note that all tensor sizes described below ignore the mini-batch dimension for simplicity.  

The \textbf{image encoder} is a ResNet152~\citep{he2016deep} pretrained on ImageNet~\citep{krizhevsky2012imagenet}. Given the encoder, the conditioning signal is a $1\times 2048$ feature representation. Note that, to avoid an undesirable equilibrium in the reconstruction loss of the \texttt{CS-VAE}, we freeze the ResNet152 during training.

\textbf{\texttt{CS-VAE}} is a standard variational autoencoder. Its encoder maps the input representation of size $1\times 2048$ to an embedded representation of size $1\times 1024$. Then, to generate the mean and standard deviation vectors, we use two fully connected branches, mapping the embedding of size $1\times 1024$ to a mean vector of size $1\times 256$ and a standard deviation vector of size $1\times 256$, where 256 is the length of the latent variable. The decoder of the \texttt{CS-VAE} maps the sampled latent variable of size $1\times 256$ to a representation of size $1\times 2048$. The generated representation acts as a reconstructed image representation. During training, we learn the reconstruction by computing the smoothed $L_1$ loss between the generated representation and the image feature of the frozen ResNet152.

The \textbf{caption encoder} is a single layer GRU network with a hidden size of 1024. Each word in the caption is represented using a randomly initialized embedding layer that maps each word to a representation of size $1\times 1024$. 

\textbf{\texttt{LCP-VAE}} is a conditional variational autoencoder. As input to its encoder, we first concatenate the image representation of size $1\times 2048$ to the caption representation of size $1\times 1024$. The encoder then maps this representation to an embedded representation of size $1\times 1024$. Then, to generate the mean and standard deviation vectors, we use two fully connected branches, mapping the embedding of size $1\times 1024$ to a mean vector of size $1\times 256$ and a standard deviation vector of size $1\times 256$, where 256 is the length of the latent variable. To sample the latent variable, we make use of our extended reparameterization trick, explained in the main paper. This unifies the conditioning and sampling of the latent variable. The \texttt{LCP-VAE} decoder then maps this latent representation to a vector of size $1\times 1024$ via a fully-connected layer. We then apply batch normalization~\citep{ioffe2015batch} on the representation, which acts as the first token to the caption decoder.

The \textbf{caption decoder} is also a single layer GRU network with a hidden size of 1024. Its first token is the representation generated by the \texttt{LCP-VAE} decoder, while the remaining tokens are the words in the corresponding caption. To decode the caption, we use a teacher forcing technique during training. At each time-step, the network chooses with probability $P_{tf}$ whether to use its own output at the previous time-step or the ground-truth token as input. We initialize $P_{tf}=1$, and decrease it linearly at each training epoch such that, after a certain number of epochs, the model becomes completely autoregressive, i.e., uses only its own output as input to the next time-step. Note that, at test time, the captions are generated completely autoregressively, i.e., with $P_{tf}=0$.

\section{Diverse Text Generation: Related Work}
\label{appendix:textrelatedwork}

A number of studies utilize generative models for language modeling. For instance,~\citep{fang2019implicit} uses VAEs and LSTMs in an unconditional language modeling problem, where posterior collapse may occur if the VAE is not trained well. To handle the problem of posterior collapse in language modeling, the authors of~\citep{fang2019implicit} try to directly match the aggregated posterior to the prior. This can be considered an extension of variational autoencoders with a regularization to maximize the mutual information, addressing the posterior collapse. VAEs are also used for language modeling in~\citep{li2019surprisingly}. In this context, it was observed that VAEs make it hard to find a good balance between language modeling and representation learning. To improve the training of VAEs in such scenarios, the authors of~\citep{li2019surprisingly} first pretrain the inference network in an autoencoder fashion, such that the inference network learns a good representation of the data in a deterministic manner. Then, they train the whole VAE while considering a weight for the KL term during training. However, the second step modifies the way VAEs optimize the variational lower bound. Furthermore, this approach prevents the model from being trained end-to-end.
Unlike these approaches, our method considers the case of \emph{conditional} text generation, where the conditioning signal (the image to be captioned in our case) is strong enough such that the caption generator can rely solely on that.

The recent work of~\citep{cho2019mixture} proposes to separate diversification from generation for the tasks of sequence generation and language modeling. The diversification stage uses a mixture of experts (MoE) to sample different binary masks on the source sequence for diverse content selection. The generation stage uses a standard encoder-decoder model taking as input the selected content from the source sequence. While effective at generating diverse sequences, this approach relies heavily on the selection part, where one needs to select the most important information in the source to generate the target sequence. Thus, the diversity of the generated target sequence depends on the diversity of the selected parts of the source sequence. Similarly, the authors of~\citep{shen2019mixture} utilize an MoE for the task of diverse machine translation. While this task is considered to be a diverse text generation one, with indeed diverse translations generated from each source sentence, the methods addressing it rely on the availablity of the a stochastic dataset, i.e., having access to multiple target sequences for each source sentence during training. By contrast, here, we design an approach that works with deterministic datasets, as those available for human motion prediction.

\section{Ablation Study on Diverse Image Captioning}
\label{appendix:captionablation}
In addition to the experiments in the main paper, in Table~\ref{tab:bleus}, we evaluate our approach, the autoregressive baseline and the CVAE, in terms of BLEU scores, i.e., BLEU1, BLEU2, BLEU3, and BLEU4 of the captions generated at test time. For the autoregressive baseline, the model generates one caption per image, which makes it straightforward to compute the BLEU scores. For the CVAE, we consider the best BLEU score among all $K=50$ sampled captions, according to the best-matching ground-truth caption. For our model, we consider the caption from the mode, i.e., the one sampled from $z=\mu_c$. Although the caption sampled from \texttt{LCP-VAE} is not chosen based on the best match with the ground-truth caption, it yields promising quality in terms of BLEU scores. For the sake of completeness and fairness, we also provide the results obtained with the best of $K$ captions for our approach, which outperform all the baselines.

\begin{table}[!t]
    \centering
        \caption{BLEU scores of different orders for sampled captions from our model as well as the baselines.}
    \label{tab:bleus}
    \begin{tabular}{l c c c c}
        \toprule
        Model & BLEU1 & BLEU2 & BLEU3 & BLEU4 \\
        \midrule
        Autoregressive (deterministic) & 0.46 & 0.39 & 0.21 & 0.16\\
        Conditional VAE (best of $K$ captions) & 0.44 & 0.38 & 0.20 & 0.17\\
        \texttt{LCP-VAE} (caption from mode) & 0.44 & 0.37 & 0.20 & 0.14\\
        \texttt{LCP-VAE} (best of $K$ captions) & 0.45 & 0.39 & 0.23 & 0.18\\
        \bottomrule
    \end{tabular}
\end{table}{}

\section{Additional Motion Prediction Qualitative Results}
\label{appendix:motion_qualitative}
\begin{figure}
    \centering
    \includegraphics[width=\textwidth]{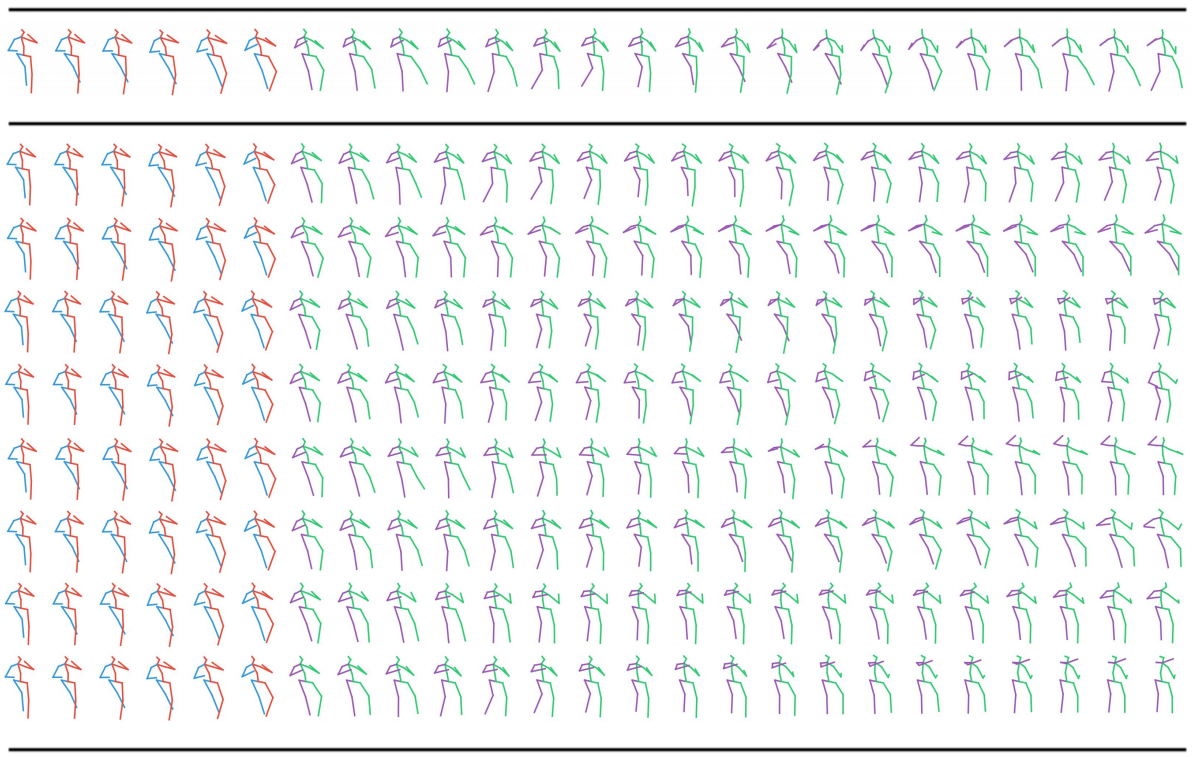}
    \caption{Qualitative evaluation of the diversity in human motion. The first row illustrates
the ground-truth motion. The first six poses of each row depict the observation (the
condition) and the rest are sampled from our model. Each row is a randomly sampled
motion (not cherry picked). As can be seen, all sampled motions are natural, with a
smooth transition from the observed to the generated ones. The diversity increases as
we increase the sequence length.}
    \label{fig:pose1}
\end{figure}

\begin{figure}
    \centering
    \includegraphics[width=\textwidth]{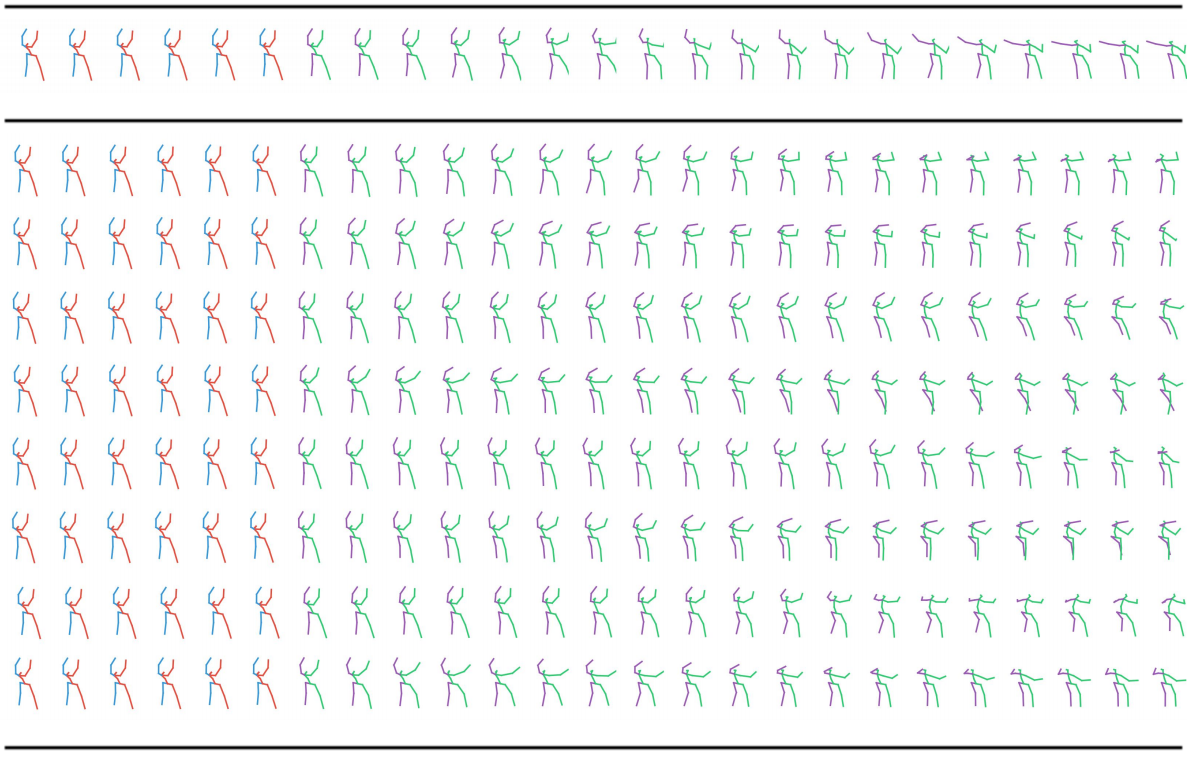}
    \caption{Additional qualitative evaluation of the diversity in human motion.}
    \label{fig:pose2}
\end{figure}

\begin{figure}
    \centering
    \includegraphics[width=\textwidth]{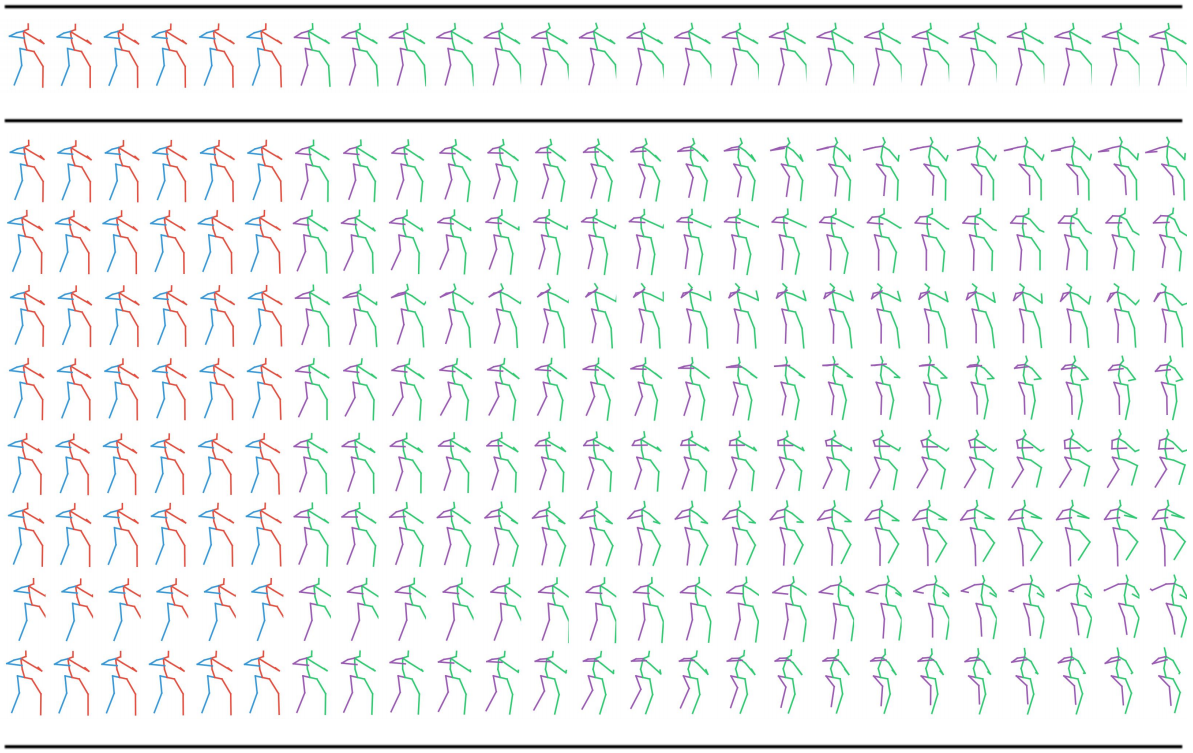}
    \caption{Additional qualitative evaluation of the diversity in human motion.}
    \label{fig:pose3}
\end{figure}

\begin{figure}
    \centering
    \includegraphics[width=\textwidth]{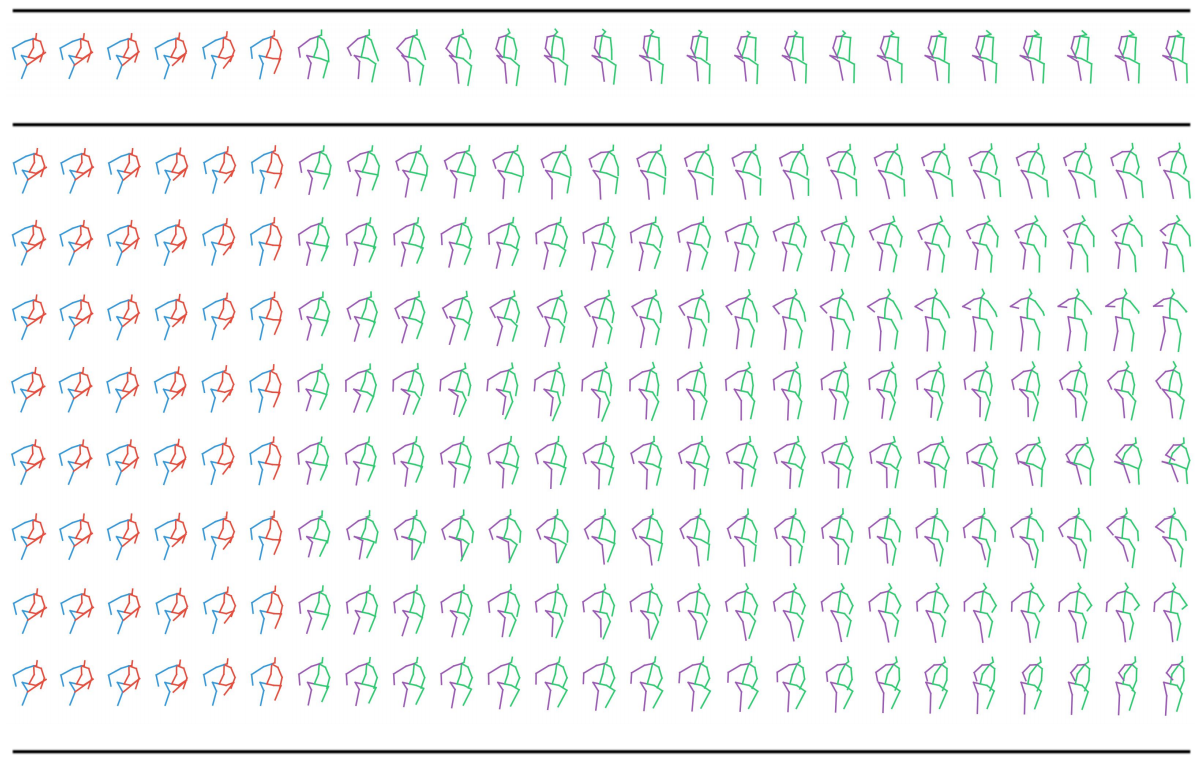}
    \caption{Additional qualitative evaluation of the diversity in human motion.}
    \label{fig:pose4}
\end{figure}

\begin{figure}
    \centering
    \includegraphics[width=\textwidth]{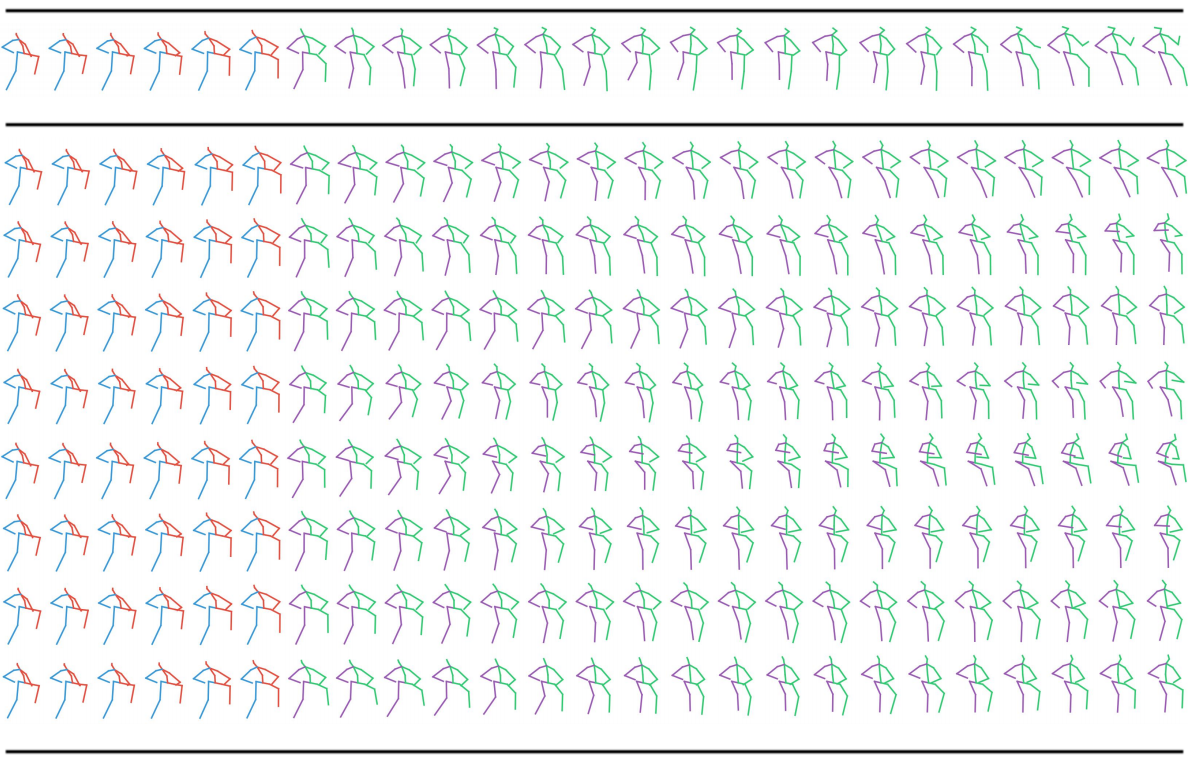}
    \caption{Additional qualitative evaluation of the diversity in human motion.}
    \label{fig:pose5}
\end{figure}

In this section we provide additional qualitative results for our LCP-VAE approach, illustrated in Fig.~\ref{fig:pose1} to Fig.~\ref{fig:pose5}.

\section{Additional Image Captioning Qualitative Results}
\label{appendix:motion_qualitative}
\begin{figure}
    \centering
    \includegraphics[width=\textwidth]{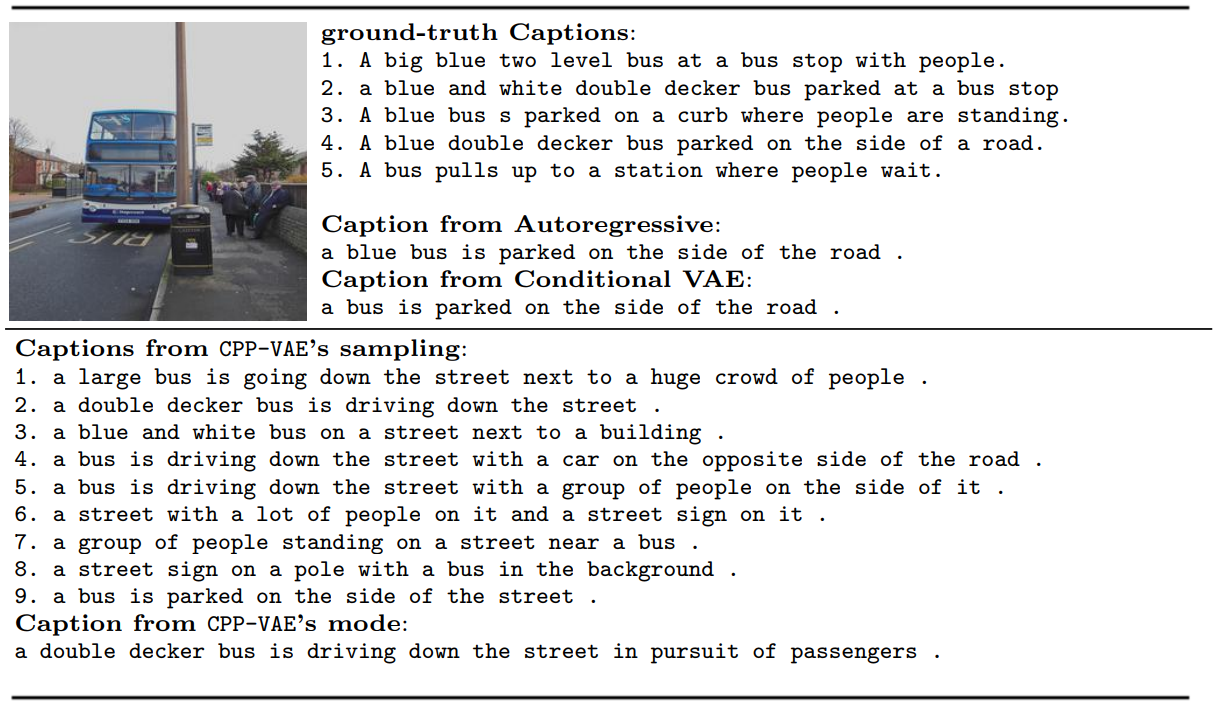}
    \caption{Qualitative evaluation of the diversity in generated captions. While captions
generated by our approach are diverse, they all describe the image properly. The caption
from mode also usually achieves a good descriptive caption.}
    \label{fig:cap1}
\end{figure}

\begin{figure}
    \centering
    \includegraphics[width=\textwidth]{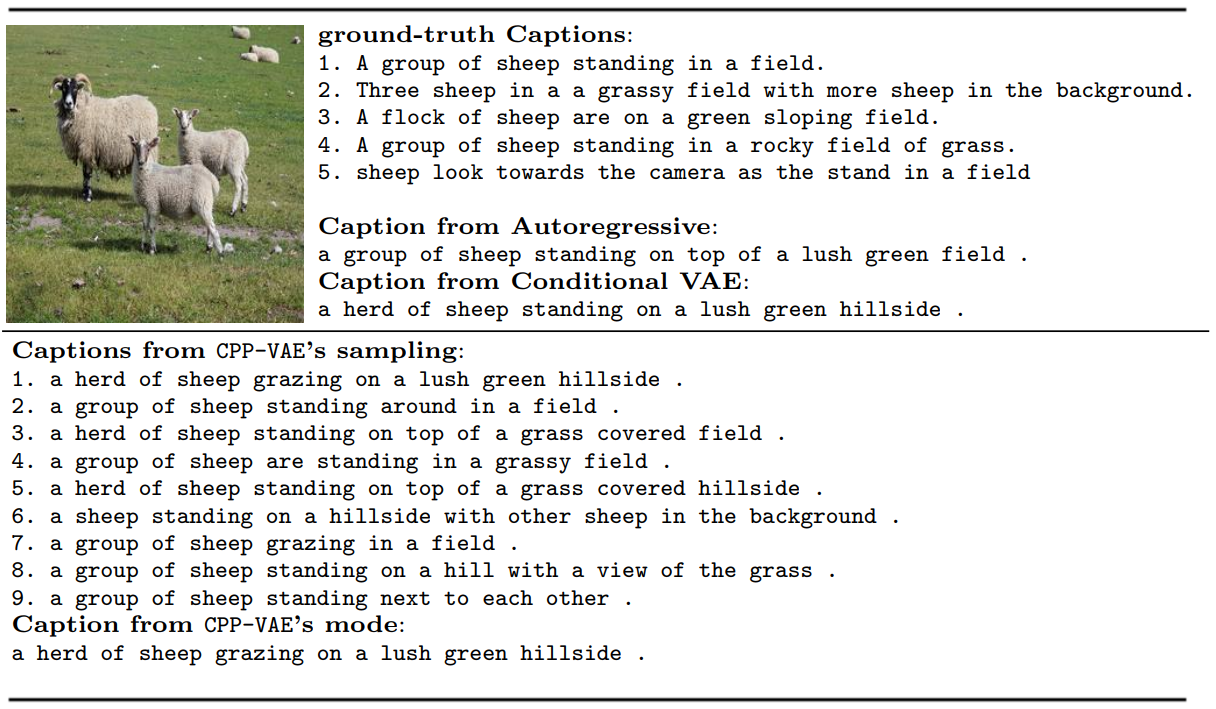}
    \caption{Additional qualitative evaluation of the diversity in generated captions.}
    \label{fig:cap2}
\end{figure}

\begin{figure}
    \centering
    \includegraphics[width=\textwidth]{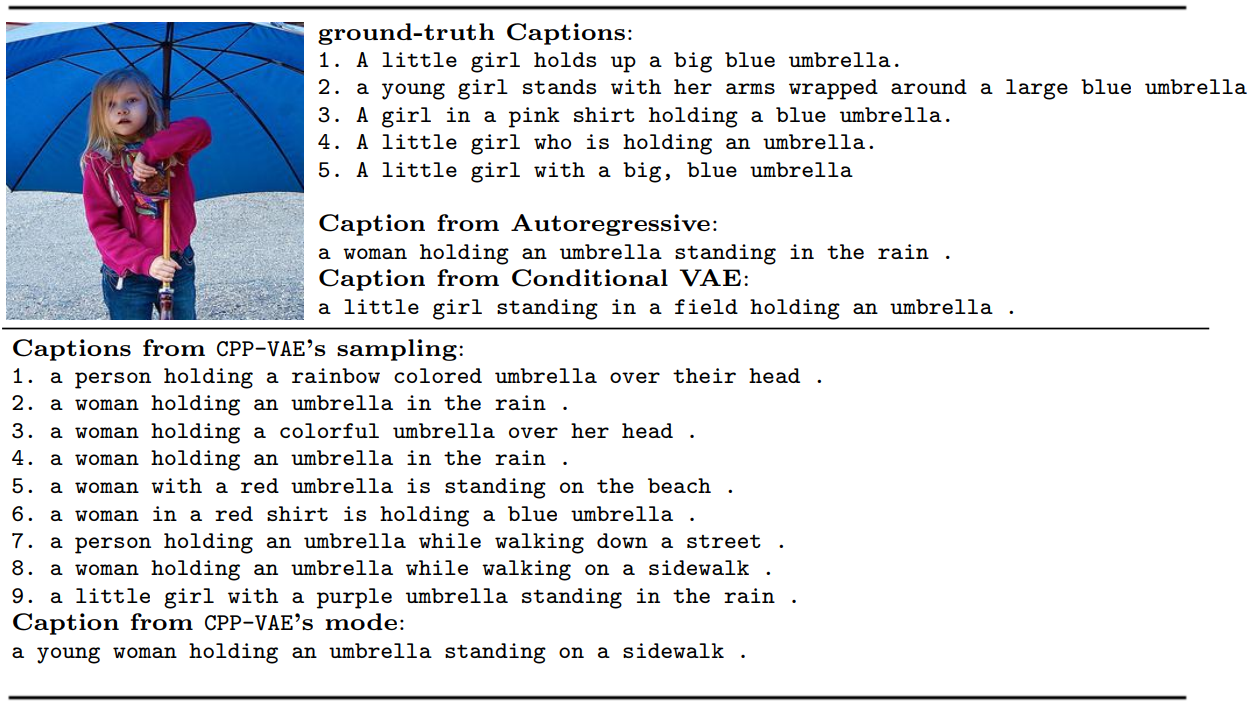}
    \caption{Additional qualitative evaluation of the diversity in generated captions.}
    \label{fig:cap3}
\end{figure}

\begin{figure}
    \centering
    \includegraphics[width=\textwidth]{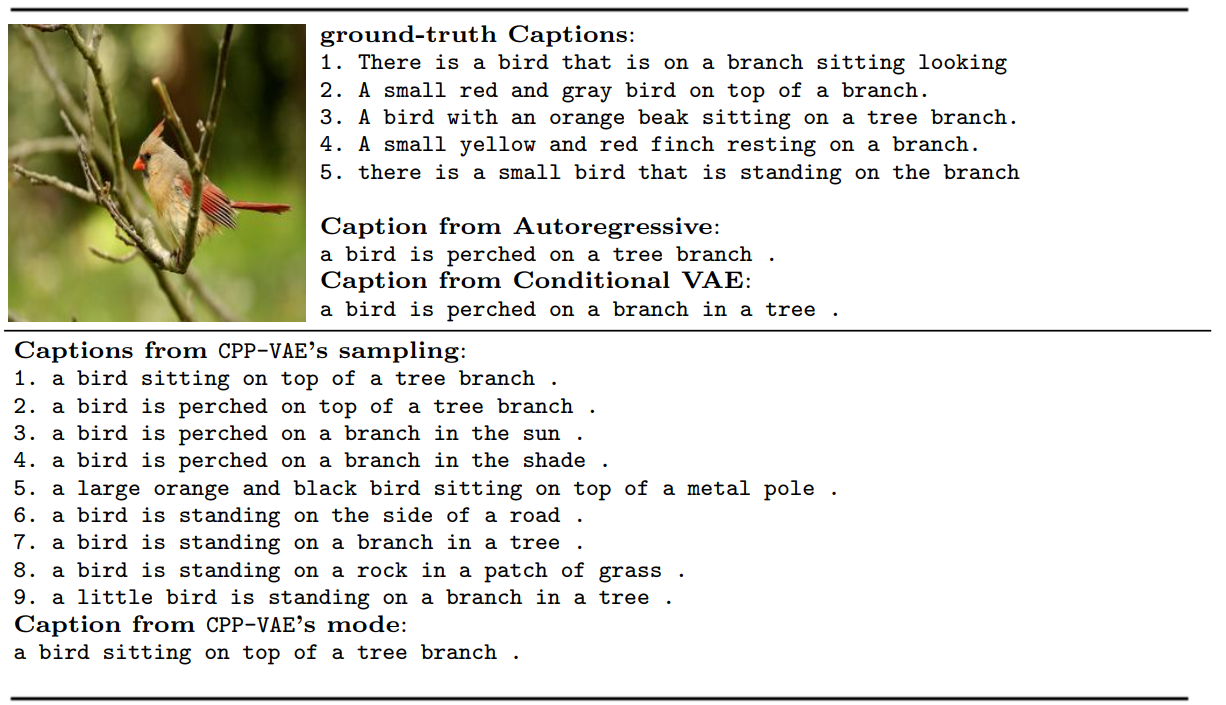}
    \caption{Additional qualitative evaluation of the diversity in generated captions.}
    \label{fig:cap4}
\end{figure}

\begin{figure}
    \centering
    \includegraphics[width=\textwidth]{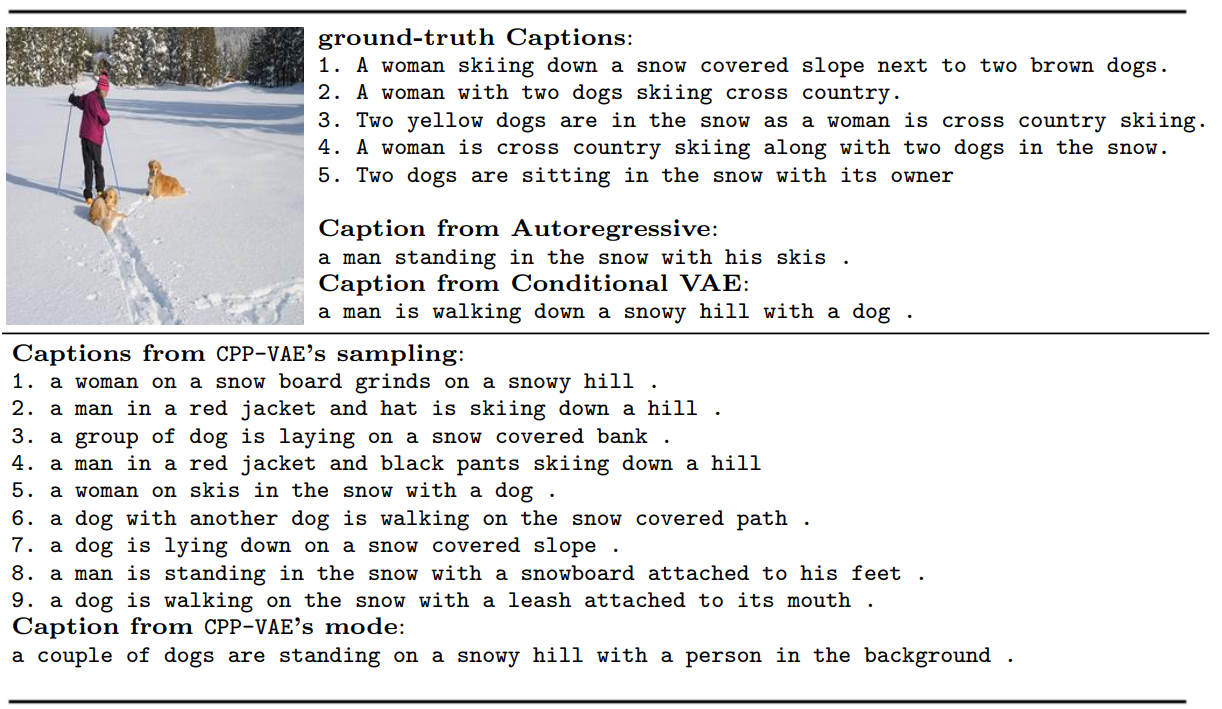}
    \caption{Additional qualitative evaluation of the diversity in generated captions.}
    \label{fig:cap5}
\end{figure}

In this section we provide additional qualitative results for our LCP-VAE approach, illustrated in Fig.~\ref{fig:cap1} to Fig.~\ref{fig:cap5}.

